\documentclass[11pt]{article}

\usepackage[final]{acl}
\usepackage[table]{xcolor}
\usepackage{booktabs}
\usepackage{times}
\usepackage{xcolor}
\usepackage{caption}
\usepackage{latexsym}
\usepackage{amsmath} 
\usepackage{multirow} 
\usepackage{multicol} 
\usepackage{enumitem}
\usepackage{graphicx}
\usepackage{xspace}
\usepackage{array}
\usepackage{subcaption}
\usepackage{svg}
\usepackage{pdflscape, amssymb, url, hyperref, enumitem, graphicx, mathtools}
\usepackage[T1]{fontenc}
\usepackage[utf8]{inputenc}

\usepackage{microtype}

\usepackage{inconsolata}
\usepackage[most]{tcolorbox}
\usepackage{xcolor}
\usepackage{booktabs}
\usepackage{tabularx}
\usepackage{array}
\newcolumntype{Y}{>{\raggedright\arraybackslash}X}

\title{
\raisebox{-0.25\height}{\includegraphics[height=2.7em]{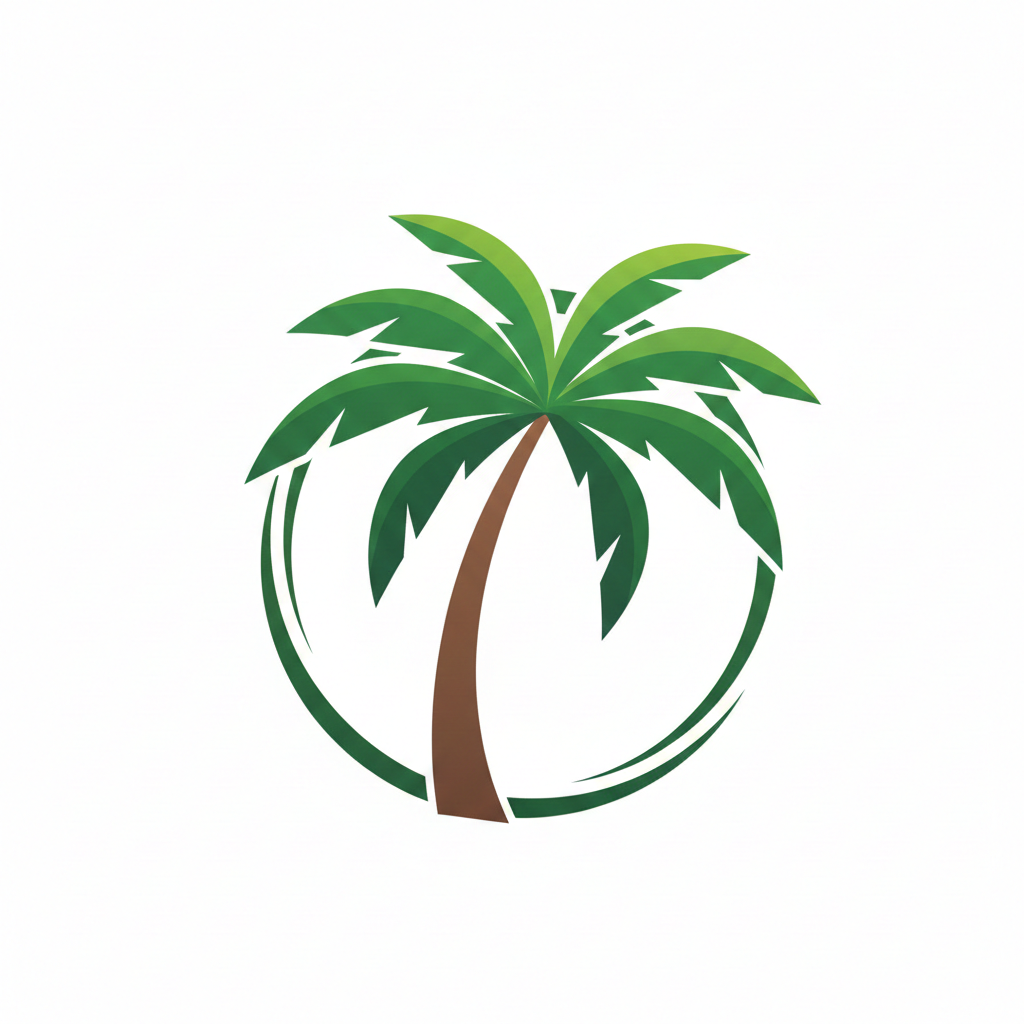}}%
PALM-Bench: A Comprehensive Benchmark for Personalized Audio-Language Models
}
\author{
\textbf{Yuwen Wang\textsuperscript{1,2}},
\textbf{Xinyuan Qian\textsuperscript{1}},
\textbf{Tian-Hao Zhang\textsuperscript{2}},
\textbf{Jiaran Gao\textsuperscript{1}},
\textbf{Yuchen Pan\textsuperscript{1}} \\
\textbf{Xin Wang\textsuperscript{2}},
\textbf{Zhou Pan\textsuperscript{2}},
\textbf{Chen Wei\textsuperscript{2}},
\textbf{Yiming Wang\textsuperscript{3}} \\
\\
\textsuperscript{1}
University of Science and Technology Beijing, China  \\
\textsuperscript{2}Li Auto, China \textsuperscript{3}Fondazione Bruno Kessler, Italy \\
\small{
\texttt{wangyuwen1114@xs.ustb.edu.cn, qianxy@ustb.edu.cn}
}
}
\begin{document}
\newcommand{\yiming}[1]{\textcolor{red}{[YM: #1]}}
\newcommand{\xinyuan}[1]{\textcolor{orange}{[XY: #1]}}
\newcommand{\yuwen}[1]{\textcolor{pink}{[YW: #1]}}
\newcommand{\tianhao}[1]{\textcolor{blue}{[TH: #1]}}

\newcommand{\task}{Audio-Language Model Personalization\xspace}
\newcommand{\benchmarkfull}{BENCHMARKFULL\xspace}
\newcommand{\benchmarkshort}{BENCHMARKSHORT\xspace}

\newcommand{\cmark}{\ding{51}}
\newcommand{\xmark}{\ding{55}}

\definecolor{lightblue}{RGB}{203, 220, 235}
\definecolor{lightgreen}{RGB}{219,234,210}
\definecolor{lightred}{RGB}{255,130,130}
\definecolor{lightyellow}{RGB}{255, 254, 200}

\newcommand{\inlineColorbox}[2]{\begingroup\setlength{\fboxsep}{1pt}\colorbox{#1}{\hspace*{2pt}\vphantom{Ay}#2\hspace*{2pt}}\endgroup}
\maketitle

\begin{abstract}
Large Audio-Language Models (LALMs) have demonstrated strong performance in audio understanding and generation. Yet, our extensive benchmarking reveals that their behavior is largely generic (e.g., summarizing spoken content) and fails to adequately support personalized question answering (e.g., summarizing what \textit{my best friend} says). In contrast, human conditions their interpretation and decision-making on each individual's personal context. To bridge this gap, we formalize the task of Personalized LALMs (PALM) for recognizing personal concepts and reasoning within personal context. Moreover, we create the first benchmark (PALM-Bench) to foster the methodological advances in PALM and enable structured evaluation on several tasks across multi-speaker scenarios. Our extensive experiments on representative open-source LALMs, show that existing training-free prompting and supervised fine-tuning strategies, while yield improvements, remains limited in modeling personalized knowledge and transferring them across tasks robustly.
\textit{\textcolor{magenta}{Data and code will be released.}}
\end{abstract}

\section{Introduction}\label{sec:intro}

\begin{figure}[]
  \centering
  \includegraphics[width=0.45\textwidth]{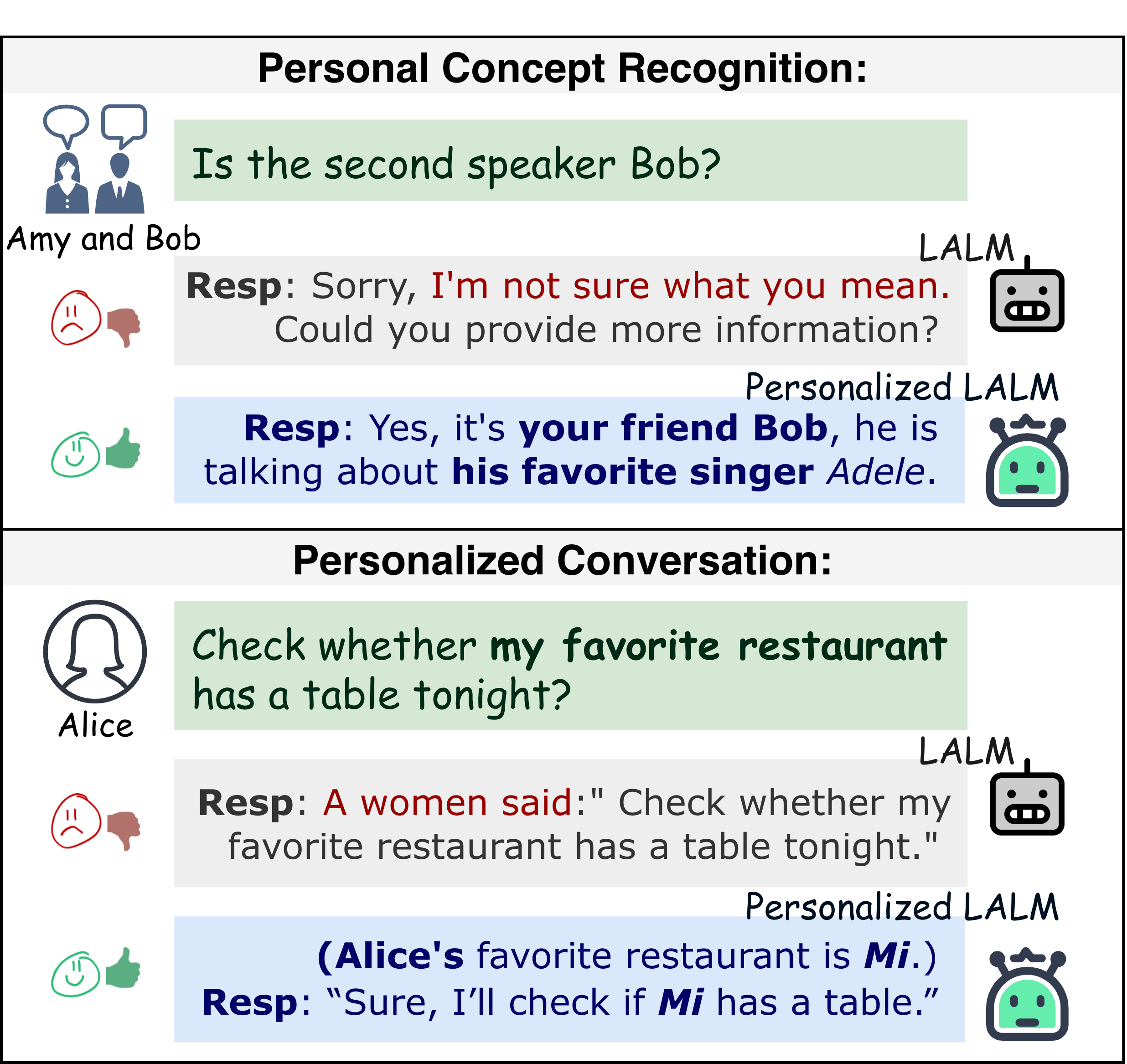}
  \caption{General \textit{v.s.} Personalized LALMs  on personal concept recognition and personalized conversation.}
  % \vspace{-15pt}
  \label{fig:task}
\end{figure}

Human auditory perception exhibits selective attention, an inherent mechanism that enables the brain to focus on a specific person's voice while neglect background noise~\cite{cherry1953some}. In practice, listeners can simultaneously comprehend spoken content and recognize the speaker, enabling personalized inference based on prior context (e.g., Figure \ref{fig:task}, inferring what ``\textit{favorite restaurant}'' refers to).

Despite advancements in Automatic Speech Recognition (ASR) and content comprehension, current LALMs such as Qwen2-Audio \cite{Qwen2-Audio} and Step-Audio 2 \cite{Step-Audio} still exhibit a fundamental difficulty in personalized understanding.
% This ability of personalized understanding remains difficult for current LALMs, such as Qwen2-Audio \cite{Qwen2-Audio} and Step-Audio 2 \cite{Step-Audio}, despite their generally improved speech understanding capability
% in terms of better Automatic Speech Recognition (ASR) and content comprehension.
Existing LALMs may understand that the second speaker mentions their favorite singer, yet lack the personalized context (e.g., the second speaker is \textit{your friend Bob}, whose favorite singer is \textit{Adele}) to make suitable responses.

% Achieving personalization in LALMs is challenging.
%We attribute this gap to two inherent structural limitations rather than engineering implementation. 

Yet, endowing LALMs with the capability of personalized understanding remains challenging. 
What might be the reason?
% Why so? 
First, pre-training biases often favor general concepts over instance-level specificity, collapsing subject-specific traits (e.g., \textit{Alice's} voice) into coarse attributes like gender or age. 
Second, an inherent tension exists between semantic abstraction and acoustic fidelity. Unlike the visual modality with explicit personalization cues (e.g., faces or clothing), speaker-specific information in speech resides in low-dimensional acoustic features (e.g., fundamental frequency) and is tightly coupled with linguistic content,  making it more likely to be attenuated when backbones like Whisper \cite{whisper} prioritize deep semantic representations.
Given these complexities, existing solutions fall short. For instance, cascading external modules for speaker recognition and language modeling causes error propagation and disconnects perception from reasoning. Joint training is computationally expensive while may suffer from catastrophic forgetting. Simple linear projection struggles to capture specific user differences, as it fails to bridge the gap between unstructured acoustic signals and operable semantic entities within LALMs. These technical challenges, together with the lack of unified task definitions and benchmarks, further limit progress in achieving personalized LALMs.

To bridge this gap, we formalize the task of personalization of LALMs, short for PALM, to reflect the capability of recognizing personal concepts and understanding personalized context. Moreover, we devise PALM-Bench, the first benchmark for the development and evaluation of PALM to facilitate further exploration in this direction. It standardizes task settings and evaluation criteria, enabling comprehensive and reproducible analysis of personalization in LALMs.
Specifically, PALM-Bench considers three progressively challenging subtasks: i) concept activation, ii) selective understanding and iii) personalized reasoning, that are reflected via audio question-answer (QA) pairs. 
On PALM-Bench, we extensively evaluate state-of-the-art open-source LALMs, adopting both supervised fine-tuning (SFT) and diverse prompting strategies ranging from standard instructions to variants augmented with Chain of Thought (CoT). Our contributions are summarized as:
\begin{itemize}[noitemsep,topsep=0pt,leftmargin=*] 
    \item We formalize the task of personalization of LALMs, enabling them to recognize and respond to user-specific concepts.
    \item We propose a flexible and controllable data construction pipeline for LALMs personalization, supporting diverse task requirements.
    % \item We create a profile-enriched dataset specifically designed to explore the task of LALMs personalization, providing a solid foundation for both training and evaluation.
    \item We release PALM-Bench, the first benchmark designed to systematically evaluate the personalization capabilities of LALMs featuring varying levels of complexity.
    \item We conduct extensive evaluation on state-of-the-art models on PALM-Bench with standard tuning techniques, revealing their strengths and limitations in the personalization task.
\end{itemize}

\section{Related Work}\label{sec:related}

\textbf{Large Audio-Language Models.} LALMs have achieved remarkable success in end-to-end speech understanding and generation through unified modality encoding. Representative works, such as Kimi-Audio \cite{Kimi-Audio} and MiDashengLM \cite{MiDashengLM}, have successfully integrated ASR with dialogue capabilities into a single framework, expanding the application of general multimodal models in speech. While these models demonstrate broad general knowledge, they were not designed for handling personalized queries (e.g., summarizing what \textit{your best friend} is saying). This is because current models follow a content-centric approach, prioritizing semantic understanding of speech content. In this way, personalization information is often embedded in acoustic features or exists as coarse labels (e.g., gender), preventing the model from abstracting it into stable semantic conditions. Thus, while current models can comprehend what is said, they struggle to consistently perceive \textit{who is speaking} or \textit{who said what} in long-context scenarios, thereby limiting their potential for personalized interaction.

\noindent\textbf{Approaches to Personalization.} 
Personalization requires models to treat personalized knowledge as a stable condition for reasoning. Visual models achieve this by embedding subject details into learnable tokens via prompt tuning \cite{MC-LLaVA,Yo'LLaVA}. In contrast, personalization in the speech domain has been explored from a different perspective. For example, \citet{persoDA} adapts to user-specific acoustic environments by modeling characteristic background noise patterns, while \citet{CachingNetworks} caches frequent commands for fast decoding. However, they treat personalization as a constraint for transcription correctness rather than a semantic context. Overall, the core challenge in achieving personalized LALMs lies in integrating personalized information into semantic reasoning and content generation.

\noindent\textbf{Benchmarking Models for Speech.} Existing LALMs benchmarks tackle the problem of logical reasoning \cite{SpeechR}, long-context understanding \cite{BLAB} and paralinguistic perception \cite{SD-Eval,VoiceBench}. However, these benchmarks treat personalization variations as disturbances to be countered, instead of a fundamental condition for reasoning. This contrasts with the real-world requirements of personalization: personalized interactions require models to treat specific subjects and their associated information as stable, referable semantic conditions, enabling consistent reasoning across tasks and contexts. 
Given the lack of a structured methodology to evaluate this ability, we propose a hierarchical task design covering subject recognition, sentence attribution, and subject-conditioned reasoning to systematically assess personalization in LALMs.

\section{Formulating LALMs Personalization}\label{sec:problem formulation}

% \textbf{Preliminaries and Input Definition.} 
We formulate  LALMs personalization  as a conditional text generation problem. 
Let $\mathcal{X}$ and $\mathcal{Y}$ denote the input and output, respectively. For each sample, the input  $\mathcal{X} = (\mathcal{A}, \mathcal{Q}, \mathcal{P})$ is a tuple, consisting of: 
% Let  $\mathcal{X} = (\mathcal{A}, \mathcal{Q}, \mathcal{P})$ denotes the input, consisting of: 
% i) \textbf{Audio ($\mathcal{A}$)}: A sequence involving $N \in \{1, 2, 3, 4\}$ distinct speakers, who constitute the ground-truth set $\mathcal{S}_{audio}$.
% ii) \textbf{Query ($\mathcal{Q}$)}: A natural language query or instruction specifying the task intent.
% iii) \textbf{Personalized Profile ($\mathcal{P}$)}: A structured text containing personalized knowledge ( $\mathcal{P} = \emptyset$ for tasks independent of external profiles).
\begin{itemize} [noitemsep]
    \item \textbf{Audio ($\mathcal{A}$)}: A sequence involving $N \in \{1, 2, 3, 4\}$ distinct speakers, who constitute the ground-truth set $\mathcal{S}_{audio}$.
    \item \textbf{Query ($\mathcal{Q}$)}: A natural language query or instruction specifying the task intent.
    \item \textbf{Personalized Profile ($\mathcal{P}$)}: A structured text containing personalized knowledge ( $\mathcal{P} = \emptyset$ for tasks independent of external profiles).
\end{itemize}

Our objective is to model $P(Y | \mathcal{A}, \mathcal{Q}, \mathcal{P})$, the probability  to generate a natural language response $Y$ that satisfies the constraints in $\mathcal{Q}$. To systematically evaluate personalized capability, we decompose the problem into three subtasks as follows.

\subsection{Concept Activation (Recognition)}
Concept activation serves as the foundation for personalization, determining the existence of target subjects within the audio. The primary challenge, particularly in multi-speaker environments, is achieving precise presence detection while maintaining robust refusal for negative samples. We reflect this capability via the recognition task. Let $\mathcal{S}_{target}$ be the set of subjects queried by the user in $\mathcal{Q}$. The objective is subject-wise verification, where the model is required to explicitly validate the presence of each queried subject $s \in \mathcal{S}_{target}$. For each  $s$, the sub-response $y_s$ is formulated as:
    $$y_s =
            \begin{cases}
            \text{Positive Confirmation}, & \text{if } s \in \mathcal{S}_{audio} \\
            \text{Refusal Response}, & \text{if } s \notin \mathcal{S}_{audio}
            \end{cases}
     $$
The final output $Y$ aggregates $\{y_s | \forall s \in \mathcal{S}_{target}\}$, grounding predictions in audio evidence. For negative samples, the expected behavior is an explicit negative response rather than an unsupported claim of presence (details in Appendix \ref{appendix-qa examples}).

\subsection{Selective Understanding (Captioning)}
This task demands selective attention: the model must isolate utterances belonging to the target subject and derive semantic understanding while filtering out non-target speaker content. We reflect this capability via the captioning task: given the audio $\mathcal{A}$ and a query $\mathcal{Q}$ focusing on the utterances of a specific target $s$, the model is required to generate a content summary or answer. Similar to Task 1, we enforce a strict refusal mechanism to prevent hallucination and ensure model output:
    $$Y =
        \begin{cases}
        \text{Content Response about } s, & \text{if } s \in \mathcal{S}_{audio} \\
        \text{Refusal Response}, & \text{if } s \notin \mathcal{S}_{audio}
        \end{cases}
    $$
This definition requires the model to implicitly distinguish between speakers. In complex environments with alternating speakers where $s \in \mathcal{S}_{audio}$, the model must ensure $Y$ is derived exclusively from the target subject, effectively disregarding the context of others (details in Appendix \ref{appendix-qa examples}).

\subsection{Personalized Reasoning}
The final stage integrates acoustic perception with external personalized knowledge. A user profile $\mathcal{P}$, representing long-term preferences, is associated with the target speaker $s$. In this task, the query $\mathcal{Q}$ is an open-ended recommendation or reasoning question. The model is required to synthesize the target's intent (extracted from $\mathcal{A}$) with static preferences in $\mathcal{P}$ to generate a response aligned with the specific subject's interests (e.g., recommending a book or selecting a gift). The generation probability is modeled as $P(Y | \mathcal{A}, \mathcal{Q}, \mathcal{P})$. This task integrates personalized priors into logical reasoning for context-aware response generation. By including recommendation-style queries with temporal variation, it supports future extension to personalized retrieval-augmented generation systems.

\section{The PALM-Bench Dataset}\label{dataset}

\begin{figure*}[t]
  \centering
  \includegraphics[width=\textwidth]{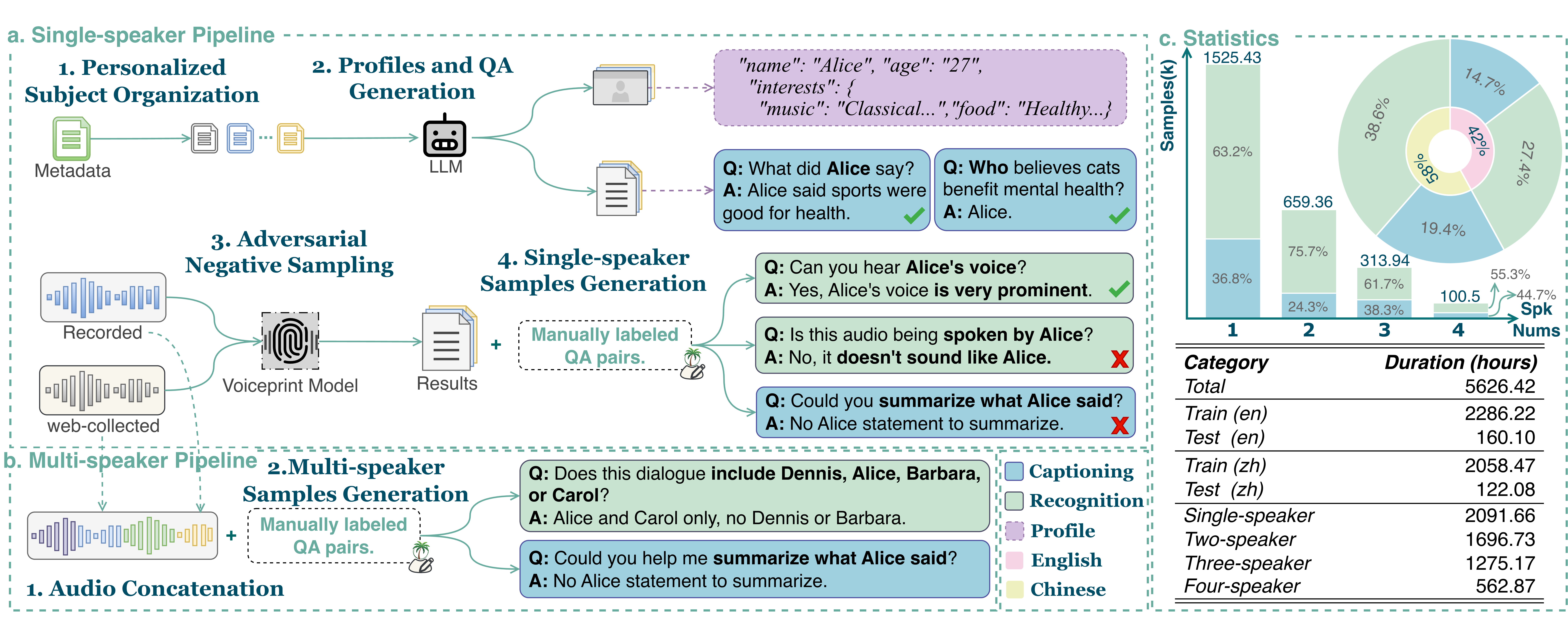}
  \caption{\textbf{Dataset creation and statistics.} Left: The pipeline constructs diverse QA pairs grounded in speaker identities, covering (a) single- and (b) multi-speaker scenarios with negative sampling. (c) Detailed statistics showing the distribution of samples across speaker numbers, languages and durations.}
  \label{fig:dataset}
\end{figure*}

\subsection{Single-speaker Pipeline}
As illustrated in Figure \ref{fig:dataset} (a), our dataset generation pipeline for the single-speaker scenario composes of four stages, details are described below.

\noindent{\textbf{Personalized Subject Organization.}} We construct the dataset using samples from the NCSSD \cite{ncssd} dataset. 
Focusing on recorded speech, we group samples by speaker ID as independent personal subjects, resulting in a core set of 27 speakers (16 Chinese and 11 English) that serve as evaluation targets. \textit{\textbf{Publicly sourced data are processed to avoid sensitive information}}.

\noindent{\textbf{Profiles and QA Generation.}} We use \textit{Qwen-Plus\footnote{\url{https://bailian.console.aliyun.com}}} to generate initial profiles and reference answers from the full ASR transcripts, and further ensure their accuracy via multi-round human verification. The final profiles include personality traits (e.g., outgoing or caring) and interests (e.g., preferred authors or movies). We generate strictly speech-grounded QA pairs relevant to the captioning task to limit hallucination (details in Appendix \ref{appendix-details-of-dataset}).

\noindent{\textbf{Adversarial Negative Sampling.}} To evaluate rejection capability when target subject is absent, we construct adversarial negative samples with graded difficulty. We utilize web-collected audio from \textit{NCSSD} as the negative pool. Using \textit{spkrec-ecapa-voxceleb\footnote{\url{https://huggingface.co/speechbrain/spkrec-ecapa-voxceleb}}}, the cosine similarity between negative candidates and the target's reference speech is computed. In this way, samples are stratified into four levels: \textit{hard, semi-hard, medium} and \textit{easy}. This design forces the model to learn speaker representations rather than relying on coarse cues (e.g., gender), ensuring robustness in open conditions.

\noindent{\textbf{Single-speaker Samples Generation.}} We generate positive and negative samples via QA templates for recognition and captioning, with configurable template counts (default: 8) and positive–negative ratios (default: 1:1). To mitigate long-tail effects due to limited positive audio, we apply data augmentation to positive samples at controllable ratios (e.g., speed perturbation, noise, and echo), and vary the amount of positive audio across task settings.

\subsection{Multi-speaker Pipeline} 
We construct controlled scenarios involving 2 to 4 speakers to simulate turn-taking while decoupling speech separation, as shown in Figure \ref{fig:dataset} (b). We employ random non-overlapping concatenation with fixed 0.5-second silence intervals, producing semantically disjoint segments that prevent the model from exploiting dialogue continuity to track personalized targets. We categorize the scene difficulty based on the speaker number. For recognition, we design permutation-based templates (e.g., A present but B absent) to evaluate personalization robustness under interference. For captioning, the challenge lies in attribution consistency, requiring the model to localize target speech and prevent content leakage from non-target subjects. By default, the dataset maintains a 4:6 ratio for single-to-multi-speaker samples and a 7:3 task ratio between recognition and captioning, both adjustable as needed.

\begin{figure*}[t]
  \centering
  \includegraphics[width=\textwidth]{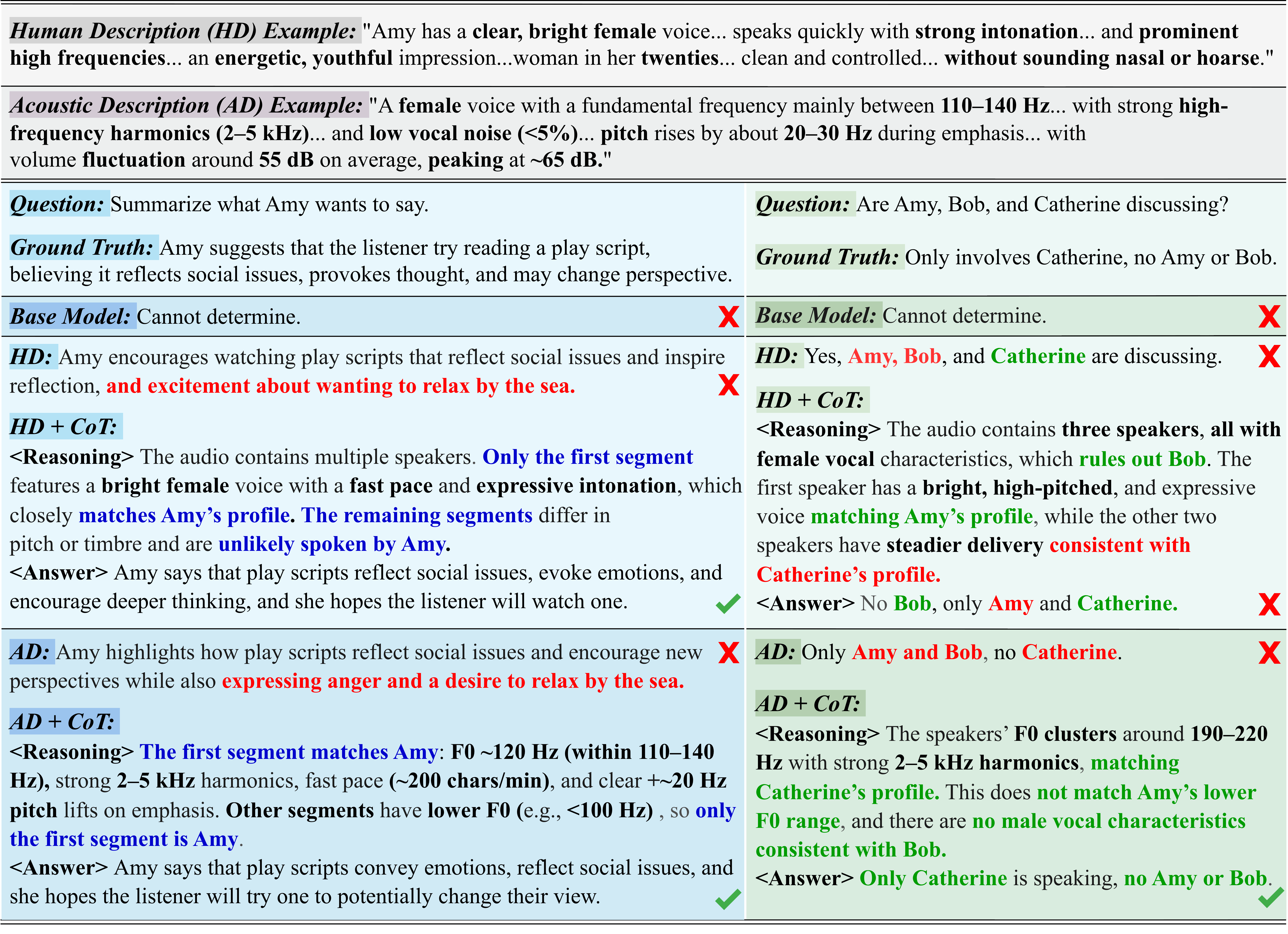}
  \caption{\textbf{Examples of different prompting strategies.} Top panels (grey) show the distinct human and acoustic descriptions. Left (blue) and right (green) regions illustrate \textit{captioning} and \textit{recognition} tasks. Within the responses, \textbf{bold black} highlights reasoning based on the provided descriptions. \textbf{\textcolor{blue}{Bold blue}} and \textbf{\textcolor{green}{bold green}} mark correct answers for their respective tasks, contrasting with \textbf{\textcolor{red}{bold red }}failures.}
  \label{fig:training-free-example}
\end{figure*}

% \vspace{-2.5pt}
\subsection{Statistics of PALM-Bench Dataset}
As illustrated in Figure \ref{fig:dataset} (c), the dataset comprises 2.6 million samples, totaling 5,626 hours, derived from 227k unique audio clips.
The distribution spans 56.5\% Chinese and 43.5\% English, providing a bilingual foundation for personalized understanding.
The training set comprises 2.46 million samples (5,345 hours), while the test set contains 139k samples spanning 281 hours. There is no overlap between the train and test splits to evaluate model generalization. On average, each Chinese speaker contributes 85.7k training samples and 3.4k test samples, while each English speaker contributes 109.1k training samples and 8.3k test samples.

\section{Evaluation on PALM-Bench}\label{sec:experiments}

We benchmark existing open-source LALMs, including Kimi-Audio \cite{Kimi-Audio}, Qwen2-Audio \cite{Qwen2-Audio}, Qwen2.5-Omni \cite{Qwen2.5-Omni}, Qwen3-Omni \cite{Qwen3-Omni} and MiDashengLM \cite{MiDashengLM}. 
We investigate both training-free prompting and training-based adaption strategies to promote personal concept recognition and personalized understanding through PALM-Bench under different task settings.
In terms of \textit{\textbf{performance metrics}}, we primarily report BLEU (BS) \cite{bleu}, F1-score (FS, in \%), BERTScore (in \%) \cite{bertscore} and LLMScore (LS), where FS and LS are reported for \textit{recognition}, and BS and LS are reported for \textit{captioning}. LS is evaluated via \textit{Qwen3-Omni-30B-Thinking}. Prompts are provided in Appendix \ref{appendix-prompt}.

\subsection{Training-free Prompting Strategies}

% Preamble:
% \usepackage{multirow}
% \usepackage[table]{xcolor}
% \usepackage{graphicx}

% Green levels (improvement only; NO delta text)
% Green levels (improvement only; NO delta text)
\definecolor{basegreen}{RGB}{180,207,176}
\definecolor{myred}{RGB}{255,100,100} 

\newcommand{\impOne}[1]{\cellcolor{basegreen!30}#1}   % 0 < Δ ≤ 15
\newcommand{\impTwo}[1]{\cellcolor{basegreen!70}#1}   % 15 < Δ ≤ 30
\newcommand{\impThree}[1]{\cellcolor{basegreen}#1}    % Δ > 30
\newcommand{\downred}[1]{\cellcolor{myred!30}#1}

\newcommand{\capImpOne}[1]{\colorbox{basegreen!30}{#1}}
\newcommand{\capImpTwo}[1]{\colorbox{basegreen!70}{#1}}
\newcommand{\capImpThree}[1]{\colorbox{basegreen}{#1}}
\newcommand{\capDownRed}[1]{\colorbox{myred!30}{#1}}

\begin{table*}[t]
  \centering
  \resizebox{\textwidth}{!}{%
  \begin{tabular}{c|cc|cc|cc|cc|cc|cc|cc|cc}
    \toprule
    \multirow{3}{*}{\textbf{Model}}           
        & \multicolumn{4}{c|}{One Speaker} & \multicolumn{4}{c|}{Two Speaker} & \multicolumn{4}{c|}{Three Speaker} & \multicolumn{4}{c}{Four Speaker} \\
        \cline{2-17}
        & \multicolumn{2}{c|}{Recognition} & \multicolumn{2}{c|}{Captioning}
        & \multicolumn{2}{c|}{Recognition} & \multicolumn{2}{c|}{Captioning}
        & \multicolumn{2}{c|}{Recognition} & \multicolumn{2}{c|}{Captioning}
        & \multicolumn{2}{c|}{Recognition} & \multicolumn{2}{c}{Captioning} \\
        \cline{2-17}
        & FS$\uparrow$ & LS$\uparrow$ & BS$\uparrow$ & LS$\uparrow$ & FS$\uparrow$ & LS$\uparrow$ & BS$\uparrow$ & LS$\uparrow$
        & FS$\uparrow$ & LS$\uparrow$ & BS$\uparrow$ & LS$\uparrow$ & FS$\uparrow$ & LS$\uparrow$ & BS$\uparrow$ & LS$\uparrow$  \\
    \hline

% ================= Base =================
\multicolumn{17}{l}{\textbf{\textit{Base Model}}} \\
\hline
Qwen2-Audio & \textbf{27.02} & \textbf{25.15} & \textbf{20.44} & \textbf{39.39} & \textbf{33.18} & \textbf{28.51} & 12.69 & \textbf{38.33} & \textbf{40.30} & \textbf{34.46} & 9.22 & \textbf{24.64} & \textbf{37.76} & \textbf{29.64} & 9.42 & \textbf{17.86} \\
Qwen3-Omni & 4.40 & 17.93 & 6.55 & 13.73 & 3.53 & 6.37 & 5.86 & 12.22 & 3.59 & 6.17 & 2.57 & 7.25 & 3.26 & 6.69 & 1.00 & 8.93 \\
Kimi-Audio & 2.89 & 8.07 & 19.49 & 25.66 & 7.81 & 20.49 & \textbf{18.50} & 27.22 & 12.48 & 29.46 & \textbf{16.79} & 20.29 & 12.37 & 19.81 & \textbf{15.55} & 16.07 \\
Qwen2.5-Omni & 1.19 & 2.68 & 7.62 & 11.20 & 3.15 & 4.70 & 8.21 & 10.56 & 4.07 & 5.68 & 4.81 & 5.80 & 3.30 & 4.78 & 10.22 & 10.71 \\
MiDashengLM & 4.64 & 6.17 & 13.63 & 17.25 & 9.23 & 7.29 & 7.24 & 8.33 & 8.85 & 7.34 & 9.55 & 10.14 & 7.48 & 6.28 & 12.34 & 12.50 \\

% ================= Human =================
\hline
\multicolumn{17}{l}{\textbf{\textit{Base Model + Human Descriptio (HD)}}} \\
\hline
Qwen2-Audio & \impTwo{50.84} & \impTwo{47.22} & \downred{20.20} & \downred{37.73} & \impOne{44.90} & \impOne{34.97} & \impOne{13.23} & 38.33 & \impOne{45.49} & \impOne{38.11} & \impOne{9.78} & \downred{22.46} & \impOne{42.97} & \impOne{30.74} & \downred{8.99} & \impOne{19.64} \\
Qwen3-Omni & \impThree{36.46} & \impTwo{45.32} & \impTwo{\textbf{29.03}} & \impThree{\textbf{52.89}} & \impThree{\textbf{57.08}} & \impThree{\textbf{56.14}} & \impTwo{\textbf{21.97}} & \impTwo{\textbf{40.00}} & \impThree{\textbf{60.57}} & \impThree{\textbf{57.52}} & \impOne{15.61} & \impTwo{\textbf{27.54}} & \impThree{\textbf{61.90}} & \impThree{\textbf{51.50}} & \impTwo{\textbf{18.37}} & \impOne{19.64} \\
Kimi-Audio & \impTwo{29.48} & \impTwo{24.50} & \impOne{26.11} & \impOne{32.14} & \impTwo{27.60} & \downred{19.58} & \downred{18.30} & \impOne{\textbf{40.00}} & \impTwo{34.73} & \impOne{31.26} & \downred{\textbf{15.64}} & \impOne{23.19} & \impTwo{40.63} & \impOne{32.10} & \impOne{15.63} & \impOne{19.64} \\
Qwen2.5-Omni & \impThree{\textbf{52.20}} & \impThree{\textbf{48.32}} & \impTwo{26.14} & \impThree{44.82} & \impThree{56.58} & \impThree{48.27} & \impOne{14.49} & \impOne{25.00} & \impThree{58.06} & \impThree{48.38} & \impOne{13.50} & \impTwo{26.81} & \impThree{56.97} & \impThree{42.90} & \impOne{14.98} & \impOne{12.50} \\
MiDashengLM & \impThree{46.42} & \impThree{38.30} & \impOne{24.41} & \impOne{36.89} & \impThree{54.43} & \impThree{48.15} & \impOne{16.69} & \impOne{21.67} & \impThree{55.27} & \impThree{49.32} & \impOne{14.21} & \impOne{20.29} & \impThree{54.72} & \impThree{47.13} & \impOne{14.08} & \impOne{\textbf{25.00}} \\

% ================= Human + CoT =================
\hline
\multicolumn{17}{l}{\textbf{\textit{Base Model + Human Description + CoT (HD+CoT)}}} \\
\hline
Qwen2-Audio & \impTwo{42.35} & \impOne{33.83} & \downred{19.69} & \downred{31.15} & \impOne{45.65} & \impOne{34.86} & \impOne{16.09} & \downred{27.22} & \impOne{42.39} & \downred{34.10} & \impOne{9.62} & \downred{18.12} & \impOne{41.92} & \impOne{30.19} & \impOne{11.14} & 17.86 \\
Qwen3-Omni & \impThree{\textbf{57.42}} & \impTwo{46.91} & \impTwo{\textbf{30.11}} & \impTwo{\textbf{37.31}} & \impThree{\textbf{60.13}} & \impThree{49.75} & \impTwo{\textbf{27.50}} & \impThree{\textbf{47.22}} & \impThree{60.76} & \impThree{50.05} & \impOne{\textbf{20.50}} & \impThree{\textbf{34.78}} & \impThree{\textbf{60.69}} & \impThree{43.72} & \impOne{\textbf{20.58}} & \impOne{\textbf{19.64}} \\
Kimi-Audio & \impThree{55.97} & \impThree{\textbf{58.70}} & \downred{16.92} & \impOne{37.27} & \impThree{57.55} & \impThree{50.59} & \downred{6.29} & \impOne{37.22} & \impThree{\textbf{61.10}} & \impTwo{54.37} & \impOne{17.32} & \impOne{\textbf{34.78}} & \impThree{60.60} & \impThree{\textbf{49.18}} & \downred{14.96} & \downred{12.50} \\
Qwen2.5-Omni & \impThree{57.41} & \impThree{47.30} & \impOne{20.66} & \impOne{26.18} & \impThree{59.00} & \impThree{\textbf{51.99}} & \impOne{14.22} & \impOne{25.56} & \impThree{\textbf{61.10}} & \impThree{\textbf{54.82}} & \impOne{12.69} & \impOne{15.94} & \impThree{59.72} & \impThree{\textbf{49.18}} & \impOne{13.82} & \downred{8.93} \\
MiDashengLM & \impThree{54.92} & \impThree{47.22} & \impOne{19.23} & \impOne{28.24} & \impThree{53.36} & \impThree{51.33} & \impOne{14.98} & \impTwo{27.78} & \impThree{54.92} & \impThree{53.65} & \impOne{12.51} & \impOne{20.29} & \impThree{55.50} & \impThree{46.31} & \impOne{12.60} & \downred{7.14} \\

% ================= Acoustic =================
\hline
\multicolumn{17}{l}{\textbf{\textit{Base Model + Acoustic Description (AD)}}} \\
\hline
Qwen2-Audio & \impTwo{\textbf{51.42}} & \impTwo{\textbf{43.57}} & \downred{20.23} & \downred{37.41} & \impOne{46.42} & \impOne{38.14} & \impOne{14.03} & \downred{35.00} & \impOne{46.17} & \impOne{40.45} & \impOne{10.49} & \impOne{31.88} & \impOne{44.39} & \impOne{33.06} & \downred{8.88} & \downred{14.29} \\
Qwen3-Omni & \impThree{39.14} & \impTwo{27.30} & \impTwo{\textbf{29.46}} & \impTwo{33.62} & \impThree{\textbf{56.30}} & \impTwo{31.31} & \impTwo{\textbf{24.30}} & \impThree{\textbf{46.67}} & \impThree{\textbf{58.84}} & \impTwo{35.05} & \impOne{\textbf{17.97}} & \impThree{\textbf{36.96}} & \impThree{\textbf{59.62}} & \impTwo{28.42} & \impOne{\textbf{19.02}} & \impOne{\textbf{28.57}} \\
Kimi-Audio & \impThree{36.52} & \impTwo{33.75} & \impOne{26.21} & \impOne{32.72} & \impOne{34.84} & \impOne{28.09} & \downred{17.86} & \impOne{40.00} & \impOne{40.51} & \impOne{35.11} & \downred{13.72} & \downred{17.39} & \impTwo{43.21} & \impTwo{35.11} & \downred{14.78} & \impOne{17.86} \\
Qwen2.5-Omni & \impTwo{36.07} & \impThree{38.34} & \impTwo{23.56} & \impTwo{32.37} & \impThree{48.30} & \impThree{42.14} & \impOne{18.27} & \impTwo{33.89} & \impThree{53.82} & \impThree{45.72} & \impOne{14.44} & \impTwo{23.91} & \impThree{56.65} & \impThree{\textbf{45.49}} & \impOne{18.13} & \impOne{21.43} \\
MiDashengLM & \impThree{33.93} & \impTwo{30.75} & \impOne{22.82} & \impTwo{\textbf{33.87}} & \impThree{51.04} & \impThree{\textbf{45.00}} & \impOne{16.56} & \impTwo{31.11} & \impThree{50.50} & \impThree{\textbf{46.80}} & \impOne{13.65} & \impOne{13.77} & \impThree{45.66} & \impThree{37.98} & \impOne{13.57} & \impOne{16.07} \\

% ================= Acoustic + CoT =================
\hline
\multicolumn{17}{l}{\textbf{\textit{Base Model + Acoustic Description + CoT (AD+CoT)}}} \\
\hline
Qwen2-Audio & \impTwo{42.31} & \impOne{31.72} & \downred{17.59} & \downred{30.60} & \impOne{45.15} & \impOne{32.59} & \impOne{13.82} & \downred{28.89} & \impOne{43.26} & \downred{29.64} & \impOne{10.46} & \downred{21.74} & \impOne{42.37} & 29.64 & \impOne{10.63} & \downred{14.29} \\
Qwen3-Omni & \impThree{\textbf{57.85}} & \impTwo{47.99} & \impOne{16.37} & \impTwo{\textbf{37.93}} & \impThree{\textbf{59.37}} & \impThree{47.78} & \impTwo{\textbf{24.81}} & \impTwo{\textbf{38.33}} & \impThree{\textbf{61.84}} & \impThree{49.46} & \impOne{\textbf{17.72}} & \impTwo{\textbf{25.36}} & \impThree{\textbf{60.87}} & \impThree{40.44} & \impOne{\textbf{19.79}} & \impTwo{\textbf{26.79}} \\
Kimi-Audio & \impThree{55.61} & \impThree{48.02} & \impOne{\textbf{27.98}} & \impOne{32.52} & \impThree{56.81} & \impThree{51.67} & \impOne{20.48} & \impOne{33.89} & \impThree{60.13} & \impThree{\textbf{56.09}} & \downred{15.74} & \impOne{23.19} & \impThree{60.78} & \impThree{\textbf{50.19}} & \downred{12.71} & \downred{8.93} \\
Qwen2.5-Omni & \impThree{54.11} & \impThree{\textbf{49.05}} & \impOne{21.00} & \impOne{20.20} & \impThree{56.78} & \impThree{\textbf{52.06}} & \impOne{13.86} & \impOne{24.44} & \impThree{57.29} & \impThree{51.49} & \impOne{13.79} & \impOne{16.67} & \impThree{56.26} & \impThree{44.13} & \impOne{13.70} & \impOne{12.50} \\
MiDashengLM & \impThree{51.32} & \impThree{47.75} & \impOne{17.87} & \impOne{30.60} & \impThree{49.96} & \impThree{48.54} & \impOne{13.72} & \impTwo{27.78} & \impThree{51.13} & \impThree{47.79} & \impOne{11.66} & \impOne{21.74} & \impThree{52.10} & \impThree{45.40} & \downred{11.88} & \impOne{17.86} \\

\bottomrule
\end{tabular}
}
\caption{\textbf{Training-free comparison on recognition and captioning across 1–4 speakers (spk 1–4).} \textbf{Bold} denotes the best per strategy. Colors visualize the gap to the Base Model: \capDownRed{red} indicates drops, while green indicates gains scaled by intensity (\capImpOne{<15}, \capImpTwo{15\text{-}30}, \capImpThree{>30}).
$\uparrow$: higher is better.
}
\label{tab:table1}
\end{table*}

To evaluate personalization without parameter updates, we study different prompting strategies considering also explicit CoT reasoning. As shown in Figure \ref{fig:training-free-example} and Table \ref{tab:table1}, we evaluate: \textit{Base Model}, \textit{Human Description} (HD), \textit{Human Description with CoT} (HD+CoT), \textit{Acoustic Description} (AD) and \textit{Acoustic Description with CoT} (AD+CoT). 
In particular, HD provides high-level semantic cues (e.g., tone and speaking rate), while AD encodes physical speech attributes (e.g., pitch and spectral energy) generated by \textit{Qwen3-Omni-Thinking}. 
This comparison isolates semantic and acoustic cues and examines whether CoT reasoning improves their use in conditional generation.

\noindent \textbf{Base Model Performance.} We first evaluate base models on this dataset. The results show limited performance on both tasks, with clear degradation as the number of speakers increases. Although Qwen2-Audio achieves high scores on most metrics, it often generates specific responses when information is insufficient, rather than following abstention or uncertainty behavior, indicating weaker instruction following in this setting.

\noindent \textbf{Effects of HD and CoT Prompting.} We investigate the effects of introducing HD and combining it with CoT prompting. HD improves model performance in multi-speaker scenarios without parameter updates. Most models achieve improvements on the recognition task, with gains under multi-speaker conditions (Qwen3-Omni: FS improves from 3.26\% to 61.90\% for 4-speaker). This suggests that natural language descriptions constrain the target subject, reducing uncertainty in target localization. In contrast, gains on the captioning task are less stable and vary across models (in 2-speaker setting, BS decreases for Kimi-Audio from 18.50 to 18.30, but increases for Qwen3-Omni from 5.86 to 21.97). This suggests that while HD supports target presence detection, it is insufficient for fine-grained semantic attribution in complex dialogues, especially with speaker turn switching.

Adding CoT leads to different model behaviors. Models with stronger cross-modal reasoning and instruction-following ability show improvements in multi-speaker settings (Qwen3-Omni: LS improves from 7.25 to 34.78 for 3-speaker captioning). However, this effect doesn't hold for all models, and performance degrades in some cases (Qwen2-Audio: LS decreases from 39.39 to 31.15 for 1-speaker captioning), indicating that CoT may introduce interference when such abilities are limited.

\noindent \textbf{Efficacy of Acoustic Description.} After adding AD, most models show substantial improvements on the recognition task, with larger gains in multi-speaker scenarios (Qwen3-Omni: LS increases from 6.69 to 28.42 for 4-speaker recognition). These show that providing explicit acoustic constraints also leads to more robust recognition performance. Compared with HD, AD offers stronger physical-level constraints and supports more precise localization in crowded acoustic conditions, as Kimi-Audio achieves higher recognition FS with AD than with HD in the 3-speaker setting (40.51\% vs. 34.73\%). When the task focus shifts toward physical localization, explicit acoustic attributes are more reliable than descriptive language cues.

After adding CoT, model behavior under AD differs across tasks. In most cases, AD+CoT further improves recognition performance (Kimi-Audio: LS from 35.11 to 56.09 for 3-speaker). In contrast, captioning performance often becomes unstable (Qwen3-Omni: LS drops from 46.67 to 38.33 for 2-speaker). Overall, these results indicate that relying on a single conditioning source is insufficient, and effective personalization requires coordination between semantic guidance and acoustic constraints.

\begin{table}[]
  \centering
  \resizebox{0.42\textwidth}{!}{%
  \begin{tabular}{c|cc|cc}
    \toprule
    \multirow{2}{*}{\textbf{Method}} &
    \multicolumn{2}{c|}{Recognition} &
    \multicolumn{2}{c}{Captioning} \\
    \cline{2-5}
    & FS$\uparrow$ & LS$\uparrow$ & BS$\uparrow$ & LS$\uparrow$ \\
    \hline
    \multicolumn{5}{l}{\textbf{\textit{Full-Parameter}}}\\
    \hline
        1e-6-epoch3& 57.52 & 44.87 & 31.56 & 44.87 \\
        1e-5-epoch5& \textbf{95.01} & 92.95 & 69.37 & 92.95 \\
        3e-6-epoch3& 92.96 & 94.23 & 70.52 & 94.23 \\
    \hline
    \multicolumn{5}{l}{\textbf{\textit{LoRA}}}\\
    \hline
       FFN & 94.95 & 93.59 & 70.39 & 95.51 \\
       Attn & 94.78 & 92.31 & \textbf{72.56} & 94.23 \\
       Audio-tower & 94.59 & 93.59 & 69.71 & \textbf{94.87} \\
       Attn+FFN & 94.85 & \textbf{96.15} & 70.49 & 92.95 \\
       All-Linear & 94.85 & 94.87 & 71.62 & 94.23 \\
    \hline
    \multicolumn{5}{l}{\textbf{\textit{Prompt Tuning}}}\\
    \hline
        token-2  & 73.53 & 44.87 & 48.15 & 67.95 \\
        token-4  & 90.62 & 87.18 & 62.99 & 82.69 \\
        token-8  & 89.17 & 83.33 & 67.74 & 87.18 \\
        token-16  & 94.80 & 93.59 & 65.05 & 85.9 \\
        token-32  & 94.66 & 95.39 & 69.80 & \textbf{94.87} \\
        token-64  & 92.24 & 89.74 & 67.45 & 91.67 \\
    \bottomrule
  \end{tabular}
  }
  \caption{\textbf{Comparison of SFT strategies.} We benchmark three SFT strategies with varying configurations. \textbf{Bold} indicates the best performance.}
  \label{tab:table2}
\end{table}

While HD and AD are all generated by Qwen3-Omni itself, the model does not fully benefit from its own descriptions to achieve effective personalization. This observation suggests that a gap still exists between description generation and personalization performance.

\definecolor{lightgray}{gray}{0.9}
\definecolor{midgray}{gray}{0.8}

\begin{table*}[t]
  \centering
  \resizebox{\textwidth}{!}{%
  \begin{tabular}{c|cc|cc|cc|cc|cc|cc|cc|cc}
    \toprule
    \multirow{3}{*}{\textbf{Model}}           
        & \multicolumn{4}{c|}{One Speaker} & \multicolumn{4}{c|}{Two Speaker} & \multicolumn{4}{c|}{Three Speaker} & \multicolumn{4}{c}{Four Speaker} \\
        \cline{2-17}
        & \multicolumn{2}{c|}{Recgnition} & \multicolumn{2}{c|}{Captioning} &  \multicolumn{2}{c|}{Recgnition} & \multicolumn{2}{c|}{Captioning} & \multicolumn{2}{c|}{Recgnition} & \multicolumn{2}{c|}{Captioning} & \multicolumn{2}{c|}{Recgnition} & \multicolumn{2}{c}{Captioning} \\
        \cline{2-17}
        & FS$\uparrow$ & LS$\uparrow$ & BS$\uparrow$ & LS$\uparrow$ & FS$\uparrow$ & LS$\uparrow$ & BS$\uparrow$ & LS$\uparrow$ & FS$\uparrow$ & LS$\uparrow$ & BS$\uparrow$ &  LS$\uparrow$ & FS$\uparrow$ & LS$\uparrow$ & BS$\uparrow$ & LS$\uparrow$  \\
    
    \hline
    \rowcolor{lightgray}
    \multicolumn{17}{c}{\textbf{\textit{(a) After Single-speaker Adaptation}}}\\
    \hline
    
    \multicolumn{17}{l}{\textbf{\textit{Full-Parameter}}}\\
    \hline
        1e-6-epoch3 &86.28 &76.95 &83.68 &82.48 &56.60 &53.85 &32.66 &\textbf{45.45} &55.74 &49.49 &\textbf{16.56} &11.76 &54.49 &56.00 &\textbf{19.77} &\textbf{14.29} \\
        1e-5-epoch5 &60.16 &50.16 &33.12 &47.57 &60.11 &42.55 &13.52 &22.73 &59.20 &37.37 &6.28 &\textbf{14.71} &\textbf{67.11} &48.00 &3.16 &0.00 \\
        3e-6-epoch3 &84.82 &73.54 &66.81 &77.36 &57.58 &48.56 &21.60 &34.09 &58.01 &47.47 &10.76 &\textbf{14.71} &60.69 &44.00 &13.85 &7.14 \\
    \hline
    
    \multicolumn{17}{l}{\textbf{\textit{LoRA}}}\\
    \hline
       FFN &85.08 &74.84 &81.18 &79.78 &54.84 &52.64 &28.83 &27.27 &58.93 &55.56 &15.98 &\textbf{14.71} &56.52 &60.00 &14.68 &\textbf{14.29} \\
       Attn &85.77 &\textbf{83.60} &83.08 &83.02 &60.66 &\textbf{66.11} &28.49 &31.82 &60.99 &\textbf{59.6} &12.52 &5.88 &63.64 &64.00 &14.75 &\textbf{14.29} \\
       Audio-tower &\textbf{88.23} &81.33 &\textbf{90.01} &\textbf{89.49} &\textbf{61.44} &57.45 &\textbf{33.41} &36.36 &63.11 &56.57 &11.81 &5.88 &61.57 &48.00 &13.72 &7.14 \\
       Attn+FFN &88.09 &81.98 &88.22 &87.60 &60.18 &59.86 &30.41 &34.09 &61.87 &54.55 &\textbf{16.56} &\textbf{14.71} &61.85 &\textbf{72.00} &19.28 &\textbf{14.29} \\
       All-Linear &85.42 &74.19 &80.49 &78.98 &57.16 &54.57 &29.72 &36.36 &55.51 &53.54 &12.21 &8.82 &61.36 &48.00 &11.54 &7.14 \\
    \hline
    
    \multicolumn{17}{l}{\textbf{\textit{Prompt Tuning}}}\\
    \hline
        Token-2 &71.24 &43.02 &47.82 &59.16 &55.67 &31.73 &22.53 &22.73 &59.31 &38.38 &13.53 &5.88 &49.20 &28.00 &15.44 &0.00 \\
        Token-4  &82.21 &67.05 &54.42 &67.65 &52.35 &37.98 &25.11 &22.73 &56.03 &48.48 &16.40 &5.88 &47.74 &12.00 &12.40 &0.00 \\
        Token-8  &83.38 &70.13 &58.45 &71.02 &57.33 &41.83 &24.99 &27.27 &62.65 &40.4 &12.10 &0.00 &59.44 &60.00 &11.84 &0.00 \\
        Token-16  &81.99 &69.64 &57.68 &66.04 &54.44 &35.58 &19.77 &15.91 &55.35 &43.43 &13.72 &0.00 &66.66 &48.00 &14.61 &7.14 \\
        Token-32  &84.82 &69.97 &61.83 &73.05 &55.77 &34.38 &26.82 &27.27 &58.30 &51.52 &12.29 &0.00 &61.04 &56.00 &15.16 &0.00 \\
        Token-64 &84.15 &72.56 &64.48 &74.26 &57.78 &48.56 &23.77 &27.27 &\textbf{63.35} &42.42 &14.10 &5.88 &63.46 &40.00 &13.29 &7.14 \\
    \hline
    
    \rowcolor{midgray}
    \multicolumn{17}{c}{\textbf{\textit{(b) After Multi-Speaker Training}}}\\
    \hline

    \multicolumn{17}{l}{\textbf{\textit{Full-Parameter}}}\\
    \hline
        1e-5-epoch5 & 95.28 & \textbf{99.39} & 99.90 & \textbf{99.93}
        & 94.57 & 99.31 & 81.16 & \textbf{83.89}
        & 92.42 & 99.91 & \textbf{78.03} & \textbf{74.64}
        & \textbf{89.96} & 99.73 & 72.70 & \textbf{71.43} \\
    \hline
    
    \multicolumn{17}{l}{\textbf{\textit{LoRA}}}\\
    \hline
        FFN & 94.98 & 84.42 & 99.95 & 86.09 & 94.51 & 85.09 & 81.06 & 71.11 & 92.39 & 86.13 & 73.63 & 66.67 & \textbf{89.96} & 83.88 & 69.45 & 57.14 \\
        Attn & 95.29 & 86.65 & 99.96 & 85.81 & \textbf{94.58} & 85.70 & 82.69 & 76.11 & \textbf{92.42} & 85.86 & 77.40 & 59.42 & \textbf{89.96} & 88.80 & 76.07 & 60.71 \\
        Audio-Tower & 95.26 & 83.98 & 97.20 & 86.40 & \textbf{94.58} & 84.53 & 68.75 & 68.89 & 92.41 & 84.68 & 60.83 & 60.14 & 89.78 & 86.75 & 54.91 & 60.71  \\
        Attn+FFN & \textbf{95.36} & 87.10 & \textbf{100.00} & 84.21 & \textbf{94.58} & 86.21 & \textbf{83.73} & 78.33 & 92.41 & 87.75 & 76.01 & 62.32 & \textbf{89.96} & 87.98 & \textbf{84.56} & \textbf{71.43} \\
        All-Linear &  95.15 & 86.13 & 99.90 & 85.81 & \textbf{94.58} & 85.72 & 82.97 & 76.67 & \textbf{92.42} & 86.31 & 76.33 & 54.92 & \textbf{89.96}  & 83.33 & 74.19 & 58.93 \\
    \hline
    
    \multicolumn{17}{l}{\textbf{\textit{Prompt Tuning}}}\\
    \hline
        Token-16 & 80.67 & 70.29 & 51.31 & 57.68
        & 91.79 & \textbf{99.76} & 33.25 & 31.82
        & \textbf{93.60} & \textbf{100.00} & 34.81 & 35.29
        & 86.07 & \textbf{100.00} & 25.80 & 28.57 \\
    \bottomrule
  \end{tabular}
  }
  \caption{\textbf{Cross-scenario transfer evaluation.} We compare performance after (a) single-speaker adaptation  vs. (b)  explicit multi-speaker training. \textbf{Bold} denotes the best result within each block.}
  \label{tab:table3}
\end{table*}

\subsection{Training-based Adaptation Strategies}
We evaluate three representative training-based strategies: Full-parameter fine-tuning (Full-FT), LoRA \cite{lora} and Prompt Tuning \cite{PromptTuning}. These strategies differ in parameter update scope and adaptation behavior.
Specifically, we apply different training strategies on Qwen2-Audio-7B-Instruct using a unified training strategy with different speaker settings. 
Moreover, we examine several LoRA placement strategies, applying LoRA to FFN layers, attention layers (Attn), their combination (Attn+FFN), the audio tower, and all linear layers (ALL-Linear).
Regarding prompt tuning, we also vary the number of tuned tokens (2$-$64).
Note that hyperparameters are tuned on validation sets for all models, and we report results from the best-performing configurations (more details in Appendix \ref{appendix-experiment}).

\noindent \textbf{Single-Speaker Training Effects.} Compared with training-free prompting, all SFT strategies improve performance after single-speaker training (Table \ref{tab:table2}). Overall, LoRA methods show more stable performance and match or exceed Full-FT on both recognition and captioning tasks, indicating that parameter updates are sufficient for adaptation in single-speaker settings. On the captioning task, LoRA-attn performs best (BS=72.56), which suggests that adapting attention layers better supports acoustic-to-semantic mapping. In contrast, prompt tuning is highly sensitive to token capacity and remains inferior to LoRA even with more tokens (token-2: BS=48.15), shows that conditional tokens alone are insufficient for generative alignment.

Table \ref{tab:table3} (a) evaluates direct transfer to multi-speaker scenarios after single-speaker adaptation. Performance degrades across all strategies as the speaker number increases. In comparison, captioning is more sensitive to scene complexity than recognition, because it requires accurate attribution of spoken content to the target personal subject under multi-speaker conditions. Methods that maintain reasonable recognition performance still show sharp degradation in captioning quality (token-8: FS=59.44 and BS=11.84 for 4-speaker). This indicates limited robustness of single-speaker adaptation for multi-speaker semantic generation.

\begin{table}[]
  \centering
  \small
  \resizebox{0.46\textwidth}{!}{%
  \begin{tabular}{cccc}
    \toprule
    \multirow{2}{*}{\textbf{Method}} & \textbf{AISHELL} & \textbf{LibriSpeech} & \textbf{MMAU} \\
    & CER $\downarrow$ & WER $\downarrow$ & ACC $\uparrow$ \\
    \midrule
    Base Model   & 0.03 & 0.15 & 0.64 \\
    \midrule
    Full-Parameter     & 0.62 & 0.25 & 0.51 \\
    \midrule
    \multicolumn{4}{l}{\textbf{\textit{LoRA}}}\\
    FFN           & 0.36 & 0.18 & \textbf{0.64} \\
    Attn          & 0.41 & 0.18 & 0.57   \\
    Audio-Tower   & 4.40 & 1.16 & 0.30 \\
    Attn+FFN      & 0.13 & 0.16 & 0.62   \\
    All-Linear    & 0.07 & 0.16 & 0.59 \\
    \midrule
    \multicolumn{4}{l}{\textbf{\textit{Prompt Tuning}}}\\
    Token-16 & \textbf{0.03} & \textbf{0.15} & 0.63 \\
    \bottomrule
  \end{tabular}
  }
  \caption{\textbf{Evaluation on other public datasets} to assess catastrophic forgetting and the stability of SFT. }
  \label{tab:table4}
\end{table}

\noindent  \textbf{Multi-Speaker Training Effects.} Explicit multi-speaker training improves performance in cross-speaker scenarios, which indicates strong latent acoustic adaptation capacity in LALMs, see Table \ref{tab:table3} (b). As the speaker number increases and scene complexity grows, the performance of all training strategies degrades, with the captioning task showing greater sensitivity and more pronounced degradation. Across different SFT strategies, LoRA and Full-FT exhibit clear differences in task emphasis. 
Specifically, LoRA methods achieve better overall performance on captioning and the Attn+FFN strategy is most stable under complex multi-speaker conditions, achieving good results (BS=76.33/84.56 for 3/4 speakers). This suggests a better balance between recognition capability and generative personalization robustness. In contrast, Full-FT remains relatively stable in recognition (LS=74.64/71.43 for 3/4 speakers), while prompt tuning performs poorly on captioning (LS=31.82 for 2-speaker), indicating that a fixed number of learnable tokens is insufficient for complex multi-speaker semantic generation.

\noindent  \textbf{General capabilities after training.}
% We also evaluate fine-tuned models in terms of ASR and audio understanding after multi-speaker training. 
Specifically, we use general datasets and benchmarks including AISHELL-1 \cite{aishell}, LibriSpeech \cite{librispeech}, MMAU \cite{mmau}, and we report CER, WER and Accuracy metrics, to assess potential catastrophic forgetting in fine-tuned models. 
From Table \ref{tab:table4}, we observe substantial degradation relative to the base model, with a increased CER on AISHELL-1, which reflects severe catastrophic forgetting. Among LoRA-based methods, Attn+FFN and All-Linear remain close to the base model while maintaining strong performance in multi-speaker scenarios, which demonstrates good stability and generalization. In contrast, applying LoRA to the audio tower causes degradation in general capabilities, which indicates that updating the acoustic frontend disrupts pre-trained acoustic and semantic representations and becomes a major source of catastrophic forgetting.

\section{Conclusion}\label{sec:conclusion}
In this work, we provide a comprehensive definition of personalized LALMs and introduce PALM-Bench, the first benchmark specifically designed to evaluate this capability. To this end, we conduct an extensive evaluation including multiple state-of-the-art LALMs, covering both training-free strategies and multiple SFT approaches. The results systematically reveal the strengths and limitations of existing models in personalization, demonstrate substantial differences across models and adaptation strategies in their ability to exploit personal knowledge, transfer across tasks and maintain stable generalization. By enabling controlled evaluation across speaker-level scenarios and task types, PALM-Bench establishes a foundation for more reliable personalization modeling of LALMs in real-world applications.

\section*{Limitations}
Although PALM-Bench supports the evaluation of personalized LALMs, this study has several limitations. For example, user profiles are derived from simulated summaries of dialogue history and do not include memory mechanisms for real time updates. Moreover, personalization modeling is restricted to the audio modality and does not incorporate visual information or long term interaction history.

% Bibliography entries for the entire Anthology, followed by custom entries
%\bibliography{anthology,custom}
% Custom bibliography entries only
\nocite{*}
\bibliography{custom}

% \section{Example Appendix}
% \label{sec:appendix}

% This is an appendix.
\appendix
\clearpage
\section{Experiment} \label{appendix-experiment}

\subsection{Experimental Setup} \label{appendix-experiment-setup}
All experiments are conducted using Qwen2-Audio-7B-Instruct and implemented with the \textit{ms-swift} framework. Full-parameter fine-tuning and LoRA experiments are run on four NVIDIA A100 GPUs, while prompt tuning experiments are conducted on two NVIDIA RTX 3090 GPUs.
All models are optimized using AdamW, and key hyperparameters are tuned separately for each setting.

\subsection{Training Details for the Single-Speaker Setting}
\label{appendix-experiment-single}

\textbf{LoRA.} We sweep the learning rate from $1\times10^{-3}$ to $1\times10^{-5}$ and evaluate 3, 5 and 10 training epochs. The best performance is obtained with a learning rate of $1\times10^{-5}$ and 5 epochs. Under this configuration, training takes about 9 hours on average.

\noindent \textbf{Full-parameter fine-tuning.} We explore several combinations of learning rates and training epochs, all of which are reported in the main results. The best configuration reaches best performance at epoch 3, with a total training time of about 2 hours.

\noindent \textbf{Prompt Tuning.} The final setting uses a learning rate of $1\times10^{-5}$ and 5 epochs, which yields stable results. Smaller learning rates, including $1\times10^{-6}$ and $1\times10^{-7}$, are also tested but do not consistently improve performance. Each speaker is assigned an independent set of soft prompt tokens, which are randomly initialized and stored separately. Prompt lengths ranging from 2 to 64 tokens are evaluated.

\subsection{Training Details for the Multi-Speaker Setting}
\label{appendix-experiment-multi}

\textbf{LoRA.} In the multi-speaker setting, LoRA fine-tuning uses larger batch sizes. To improve stability, the learning rate is reduced from $1\times10^{-3}$ to $1\times10^{-4}$, and the number of epochs is reduced from 10 to 5. This configuration results in a total training time of about 3.5 days.

\noindent \textbf{Full-parameter fine-tuning.} The final configuration uses a learning rate of $1\times10^{-5}$ and 5 epochs, with a total training time of about 12 hours.

\noindent \textbf{Prompt Tuning.} Prompt tuning in the multi-speaker setting uses a learning rate of $1\times10^{-3}$ and is trained for 15 epochs. Although only soft prompt tokens are updated, longer training is required to reach stable convergence, resulting in a total training time of about 7 days. For multi-speaker experiments, 16 tokens are selected as a balanced choice between performance and efficiency.

\subsection{Full Experimental Results}
\label{appendix-full experiment results}
\begin{table*}[t]
  \centering
  \resizebox{\textwidth}{!}{%
  \begin{tabular}{c|cccc|cccc|cccc|cccc}
    \hline
    \multirow{3}{*}{\textbf{Model}}
        & \multicolumn{8}{c|}{One Speaker} & \multicolumn{8}{c}{Two Speaker} \\
        \cline{2-17}
        & \multicolumn{4}{c|}{Recognition} & \multicolumn{4}{c|}{Captioning}
        & \multicolumn{4}{c|}{Recognition} & \multicolumn{4}{c}{Captioning} \\
        \cline{2-17}
        & BLEU$\uparrow$ & FS$\uparrow$ & BS$\uparrow$ & LS$\uparrow$
        & BLEU$\uparrow$ & FS$\uparrow$ & BS$\uparrow$ & LS$\uparrow$
        & BLEU$\uparrow$ & FS$\uparrow$ & BS$\uparrow$ & LS$\uparrow$
        & BLEU$\uparrow$ & FS$\uparrow$ & BS$\uparrow$ & LS$\uparrow$ \\
    \hline
    \multicolumn{17}{l}{%
    \raisebox{0pt}[2.5ex][1.3ex]{\textbf{\textit{Base Model}}}%
    }\\
    \hline

Qwen2-Audio & \textbf{13.81} & \textbf{27.02} & \textbf{64.27} & \textbf{25.15} & \textbf{20.44} & \textbf{41.24} & \textbf{69.11} & \textbf{39.39} & \textbf{22.02} & \textbf{33.18} & \textbf{65.99} & \textbf{28.51} & 12.69 & \textbf{32.47} & \textbf{67.17} & \textbf{38.33} \\
Qwen3-Omni & 0.05 & 4.40 & 54.31 & 17.93 & 6.55 & 14.49 & 58.99 & 13.73 & 0.19 & 3.53 & 55.63 & 6.37 & 5.86 & 13.75 & 58.08 & 12.22 \\
Kimi-Audio & 0.17 & 2.89 & 56.48 & 8.07 & 19.49 & 28.39 & 65.31 & 25.66 & 0.62 & 7.81 & 56.05 & 20.49 & \textbf{18.50} & 31.72 & 66.99 & 27.22 \\
Qwen2.5-Omni & 0.03 & 1.19 & 57.27 & 2.68 & 7.62 & 11.74 & 59.42 & 11.20 & 0.57 & 3.15 & 58.48 & 4.70 & 8.21 & 9.82 & 56.52 & 10.56 \\
midasheng & 0.24 & 4.64 & 57.27 & 6.17 & 13.63 & 20.15 & 61.99 & 17.25 & 2.33 & 9.23 & 59.42 & 7.29 & 7.24 & 12.23 & 56.90 & 8.33 \\

    \hline
    \multicolumn{17}{l}{%
    \raisebox{0pt}[2.5ex][1.3ex]{\textbf{\textit{Base Model + Human Description (HD)}}}%
    }\\
    \hline

Qwen2-Audio & \impOne{26.38} & \impTwo{50.84} & \impOne{\textbf{71.39}} & \impTwo{47.22} & \downred{20.20} & \downred{38.58} & \downred{68.11} & \downred{37.73} & \impOne{22.51} & \impOne{44.90} & \impOne{70.18} & \impOne{34.97} & \impOne{13.23} & \downred{31.97} & \impOne{67.57} & 38.33 \\
Qwen3-Omni & \impTwo{18.92} & \impTwo{36.46} & \impOne{65.46} & \impOne{45.32} & \impTwo{\textbf{29.03}} & \impTwo{\textbf{47.24}} & \impOne{\textbf{71.90}} & \impThree{\textbf{52.89}} & \impThree{41.73} & \impThree{\textbf{57.08}} & \impOne{73.32} & \impThree{\textbf{56.14}} & \impTwo{\textbf{21.97}} & \impTwo{\textbf{41.53}} & \impOne{\textbf{70.30}} & \impOne{\textbf{40.00}} \\
Kimi-Audio & \impOne{12.35} & \impOne{29.48} & \impOne{65.15} & \impOne{24.50} & \impOne{26.11} & \impOne{34.25} & \impOne{67.67} & \impOne{32.14} & \impOne{15.42} & \impOne{27.60} & \impOne{64.76} & \downred{19.58} & \downred{18.30} & \impOne{33.71} & \impOne{67.06} & \impOne{\textbf{40.00}} \\
Qwen2.5-Omni & \impThree{\textbf{34.87}} & \impThree{\textbf{52.20}} & \impOne{70.95} & \impThree{\textbf{48.32}} & \impTwo{26.14} & \impTwo{42.51} & \impOne{70.06} & \impThree{44.82} & \impThree{\textbf{42.66}} & \impThree{56.58} & \impOne{\textbf{74.25}} & \impThree{48.27} & \impOne{14.49} & \impOne{31.89} & \impOne{65.53} & \impOne{25.00} \\
midasheng & \impThree{31.84} & \impThree{46.42} & \impOne{69.42} & \impThree{38.30} & \impOne{24.41} & \impOne{38.72} & \impOne{68.30} & \impOne{36.89} & \impThree{37.79} & \impThree{54.43} & \impOne{73.42} & \impThree{48.15} & \impOne{16.69} & \impOne{30.77} & \impOne{63.27} & \impOne{21.67} \\

    \hline
    \multicolumn{17}{l}{%
    \raisebox{0pt}[2.5ex][1.3ex]{\textbf{\textit{Base Model + Human Description + CoT (HD+CoT)}}}%
    }\\
    \hline

Qwen2-Audio & \impOne{25.52} & \impOne{42.35} & \impOne{68.37} & \impOne{33.83} & \downred{19.69} & \downred{33.19} & \downred{67.55} & \downred{31.15} & \impOne{25.06} & \impOne{45.65} & \impOne{70.24} & \impOne{34.86} & \impOne{16.09} & \downred{31.31} & \impOne{67.51} & \downred{27.22} \\
Qwen3-Omni & \impThree{36.93} & \impThree{\textbf{57.42}} & \impOne{72.75} & \impOne{46.91} & \impTwo{\textbf{30.11}} & \impOne{44.40} & \impOne{70.49} & \impOne{\textbf{37.31}} & \impThree{42.14} & \impThree{\textbf{60.13}} & \impOne{75.34} & \impThree{49.75} & \impTwo{\textbf{27.50}} & \impTwo{\textbf{43.79}} & \impOne{\textbf{70.19}} & \impTwo{\textbf{47.22}} \\
Kimi-Audio & \impThree{38.43} & \impThree{55.97} & \impOne{72.85} & \impThree{\textbf{58.70}} & \downred{16.92} & \impOne{\textbf{47.41}} & \impOne{\textbf{71.28}} & \impOne{37.27} & \impThree{40.49} & \impThree{57.55} & \impOne{74.87} & \impThree{50.59} & \downred{6.29} & \impOne{38.61} & \impOne{67.70} & \impOne{37.22} \\
Qwen2.5-Omni & \impThree{\textbf{38.63}} & \impThree{57.41} & \impOne{\textbf{73.34}} & \impThree{47.30} & \impOne{20.66} & \impOne{43.83} & \impOne{69.43} & \impOne{26.18} & \impThree{\textbf{42.19}} & \impThree{59.00} & \impOne{\textbf{75.91}} & \impThree{\textbf{51.99}} & \impOne{14.22} & \impOne{31.39} & \impOne{64.51} & \impOne{25.56} \\
midasheng & \impThree{33.84} & \impThree{54.92} & \impOne{72.19} & \impThree{47.22} & \impOne{19.23} & \impOne{41.03} & \impOne{68.96} & \impOne{28.24} & \impThree{31.18} & \impThree{53.36} & \impOne{73.03} & \impThree{51.33} & \impOne{14.98} & \impOne{34.64} & \impOne{65.34} & \impOne{27.78} \\

    \hline
    \multicolumn{17}{l}{%
    \raisebox{0pt}[2.5ex][1.3ex]{\textbf{\textit{Base Model + Acoustic Description (AD)}}}%
    }\\
    \hline

Qwen2-Audio & \impOne{\textbf{26.31}} & \impTwo{\textbf{51.42}} & \impOne{\textbf{71.33}} & \impOne{\textbf{43.57}} & \downred{20.23} & \downred{38.54} & \downred{68.21} & \downred{\textbf{37.41}} & \impOne{23.23} & \impOne{46.42} & \impOne{70.82} & \impOne{38.14} & \impOne{14.03} & \impOne{32.50} & \impOne{67.82} & \downred{35.00} \\
Qwen3-Omni & \impTwo{21.66} & \impThree{39.14} & \impOne{66.25} & \impOne{27.30} & \impTwo{\textbf{29.46}} & \impThree{\textbf{48.41}} & \impOne{\textbf{72.18}} & \impThree{33.62} & \impThree{\textbf{40.78}} & \impThree{\textbf{56.30}} & \impOne{\textbf{72.92}} & \impOne{31.31} & \impTwo{\textbf{24.30}} & \impThree{\textbf{45.40}} & \impOne{\textbf{72.09}} & \impThree{\textbf{46.67}} \\
Kimi-Audio & \impTwo{21.53} & \impTwo{36.52} & \impOne{67.39} & \impOne{33.75} & \impOne{26.21} & \impOne{36.20} & \impOne{67.85} & \impOne{32.72} & \impTwo{24.78} & \impOne{34.84} & \impOne{66.95} & \impOne{28.09} & \downred{17.86} & \impOne{34.52} & \downred{66.63} & \impOne{40.00} \\
Qwen2.5-Omni & \impTwo{19.66} & \impTwo{36.07} & \impOne{66.99} & \impTwo{38.34} & \impOne{23.56} & \impOne{32.94} & \impOne{65.68} & \impOne{32.37} & \impThree{37.31} & \impThree{48.30} & \impOne{71.94} & \impThree{42.14} & \impOne{18.27} & \impOne{34.34} & \impOne{66.83} & \impOne{33.89} \\
midasheng & \impTwo{20.16} & \impTwo{33.93} & \impOne{65.15} & \impOne{30.75} & \impOne{22.82} & \impOne{35.50} & \impOne{66.87} & \impOne{33.87} & \impThree{38.47} & \impThree{51.04} & \impOne{72.24} & \impThree{\textbf{45.00}} & \impOne{16.56} & \impOne{31.72} & \impOne{63.76} & \impOne{31.11} \\

    \hline
    \multicolumn{17}{l}{%
    \raisebox{0pt}[2.5ex][1.3ex]{\textbf{\textit{Base Model + Acoustic Description + CoT (AD+CoT)}}}%
    }\\
    \hline

Qwen2-Audio & \impOne{22.54} & \impOne{42.31} & \impOne{68.44} & \impOne{31.72} & \downred{17.59} & \downred{33.14} & \downred{67.65} & \downred{30.60} & \impOne{22.73} & \impOne{45.15} & \impOne{70.20} & \impOne{32.59} & \impOne{13.82} & \downred{31.97} & \downred{66.58} & \downred{28.89} \\
Qwen3-Omni & \impThree{\textbf{37.96}} & \impThree{\textbf{57.85}} & \impOne{\textbf{72.81}} & \impTwo{47.99} & \impOne{16.37} & \impTwo{\textbf{47.01}} & \impOne{\textbf{71.39}} & \impOne{\textbf{37.93}} & \impThree{\textbf{41.01}} & \impThree{\textbf{59.37}} & \impOne{\textbf{74.98}} & \impThree{47.78} & \impTwo{\textbf{24.81}} & \impTwo{\textbf{43.78}} & \impOne{\textbf{70.17}} & \impOne{\textbf{38.33}} \\
Kimi-Audio & \impThree{37.52} & \impThree{55.61} & \impOne{72.33} & \impTwo{48.02} & \impOne{\textbf{27.98}} & \impOne{46.34} & \impOne{70.32} & \impOne{32.52} & \impThree{39.00} & \impThree{56.81} & \impOne{73.96} & \impThree{51.67} & \impOne{20.48} & \impOne{37.62} & \downred{66.54} & \impOne{33.89} \\
Qwen2.5-Omni & \impThree{35.97} & \impThree{54.11} & \impOne{72.04} & \impThree{\textbf{49.05}} & \impOne{21.00} & \impOne{41.58} & \impOne{68.32} & \impOne{20.20} & \impThree{37.44} & \impThree{56.78} & \impOne{74.66} & \impThree{\textbf{52.06}} & \impOne{13.86} & \impOne{30.94} & \impOne{64.39} & \impOne{24.44} \\
midasheng & \impTwo{28.14} & \impThree{51.32} & \impOne{70.89} & \impThree{47.75} & \impOne{17.87} & \impOne{40.16} & \impOne{68.96} & \impOne{30.60} & \impTwo{26.92} & \impThree{49.96} & \impOne{71.85} & \impThree{48.54} & \impOne{13.72} & \impOne{33.13} & \impOne{64.48} & \impOne{27.78} \\

    \hline
  \end{tabular}%
  }
  \caption{Full results under 1-2 speaker settings. \textbf{Bold} marks the best per strategy. Colors indicate the gap to the Base Model: \capDownRed{red} for drops, \capImpOne{<15}, \capImpTwo{15\text{-}30}, and \capImpThree{>30} for gains. $\uparrow$ denotes higher is better.}
  \label{tab:appendix_training-free_1+2}
\end{table*}
\begin{table*}[t]
  \centering
  \resizebox{\textwidth}{!}{%
  \begin{tabular}{c|cccc|cccc|cccc|cccc}
    \hline
    \multirow{3}{*}{\textbf{Model}}
        & \multicolumn{8}{c|}{Three Speaker} & \multicolumn{8}{c}{Four Speaker} \\
        \cline{2-17}
        & \multicolumn{4}{c|}{Recognition} & \multicolumn{4}{c|}{Captioning}
        & \multicolumn{4}{c|}{Recognition} & \multicolumn{4}{c}{Captioning} \\
        \cline{2-17}
        & BLEU$\uparrow$ & FS$\uparrow$ & BS$\uparrow$ & LS$\uparrow$
        & BLEU$\uparrow$ & FS$\uparrow$ & BS$\uparrow$ & LS$\uparrow$
        & BLEU$\uparrow$ & FS$\uparrow$ & BS$\uparrow$ & LS$\uparrow$
        & BLEU$\uparrow$ & FS$\uparrow$ & BS$\uparrow$ & LS$\uparrow$ \\
    \hline

    \multicolumn{17}{l}{\raisebox{0pt}[2.5ex][1.3ex]{\textbf{\textit{Base Model}}}}\\
    \hline
    Qwen2-Audio& \textbf{31.53} & \textbf{40.30} & \textbf{69.15} & \textbf{34.46} & 9.22 & \textbf{25.94} & \textbf{64.71} & \textbf{24.64} & \textbf{27.93} & \textbf{37.76} & \textbf{68.16} & \textbf{29.64} & 9.42 & \textbf{26.58} & \textbf{63.89} & \textbf{17.86} \\
    Qwen3-Omni & 0.14 & 3.59 & 54.22 & 6.17 & 2.57 & 7.87 & 54.10 & 7.25 & 0.16 & 3.26 & 54.34 & 6.69 & 1.00 & 7.58 & 54.81 & 8.93 \\
    Kimi-Audio & 1.24 & 12.48 & 55.27 & 29.46 & \textbf{16.79} & 25.49 & 62.90 & 20.29 & 2.38 & 12.37 & 57.19 & 19.81 & \textbf{15.55} & 24.92 & 61.43 & 16.07 \\
    Qwen2.5-Omni & 0.43 & 4.07 & 57.44 & 5.68 & 4.81 & 6.46 & 54.45 & 5.80 & 0.23 & 3.30 & 57.25 & 4.78 & 10.22 & 9.85 & 56.80 & 10.71 \\
    midasheng & 1.52 & 8.85 & 57.81 & 7.34 & 9.55 & 12.04 & 56.84 & 10.14 & 2.31 & 7.48 & 57.46 & 6.28 & 12.34 & 16.18 & 58.60 & 12.50 \\
    \hline

    \multicolumn{17}{l}{\raisebox{0pt}[2.5ex][1.3ex]{\textbf{\textit{Base Model + Human Description (HD)}}}}\\
    \hline
    Qwen2-Audio & \downred{26.31} & \impOne{45.49} & \impOne{71.14} & \impOne{38.11} & \impOne{9.78} & \impOne{26.93} & \downred{64.63} & \downred{22.46} & \downred{25.30} & \impOne{42.97} & \impOne{70.59} & \impOne{30.74} & \downred{8.99} & \downred{26.19} & \downred{63.70} & \impOne{19.64} \\
    Qwen3-Omni & \impThree{\textbf{45.92}} & \impThree{\textbf{60.57}} & \impTwo{75.15} & \impThree{\textbf{57.52}} & \impOne{15.61} & \impOne{\textbf{33.48}} & \impOne{\textbf{65.35}} & \impOne{27.54} & \impThree{\textbf{52.52}} & \impThree{\textbf{61.90}} & \impTwo{\textbf{76.57}} & \impThree{\textbf{51.50}} & \impOne{\textbf{18.37}} & \impOne{\textbf{34.77}} & \impOne{\textbf{65.58}} & \impOne{19.64} \\
    Kimi-Audio & \impTwo{20.79} & \impTwo{34.73} & \impOne{66.82} & \impOne{31.26} & \downred{\textbf{15.64}} & \impOne{28.54} & \impOne{64.06} & \impOne{23.19} & \impThree{33.13} & \impTwo{40.63} & \impOne{69.79} & \impOne{32.10} & \impOne{15.63} & \impOne{26.75} & \impOne{63.60} & \impOne{19.64} \\
    Qwen2.5-Omni & \impThree{44.52} & \impThree{58.06} & \impTwo{\textbf{75.32}} & \impThree{48.38} & \impOne{13.50} & \impOne{31.02} & \impOne{64.37} & \impOne{26.81} & \impThree{50.11} & \impThree{56.97} & \impTwo{75.33} & \impThree{42.90} & \impOne{14.98} & \impOne{29.34} & \impOne{63.09} & \impOne{12.50} \\
    midasheng & \impThree{44.09} & \impThree{55.27} & \impTwo{73.64} & \impThree{49.32} & \impOne{14.21} & \impOne{29.47} & \impOne{62.80} & \impOne{20.29} & \impThree{47.76} & \impThree{54.72} & \impTwo{74.07} & \impThree{47.13} & \impOne{14.08} & \impOne{31.50} & \impOne{63.99} & \impOne{\textbf{25.00}} \\
    \hline

    \multicolumn{17}{l}{\raisebox{0pt}[2.5ex][1.3ex]{\textbf{\textit{Base Model + Human Description + CoT (HD+CoT)}}}}\\
    \hline
    Qwen2-Audio & \downred{24.70} & \impOne{42.39} & \impOne{69.27} & \downred{34.10} & \impOne{9.62} & \impOne{27.09} & \downred{63.92} & \downred{18.12} & \downred{20.97} & \impOne{41.92} & \impOne{69.58} & \impOne{30.19} & \impOne{11.14} & \impOne{27.03} & \impOne{64.14} & \impOne{17.86} \\
    Qwen3-Omni & \impThree{47.52} & \impThree{60.76} & \impTwo{76.68} & \impThree{50.05} & \impTwo{\textbf{20.50}} & \impTwo{\textbf{37.53}} & \impOne{\textbf{66.82}} & \impTwo{\textbf{34.78}} & \impThree{49.08} & \impThree{\textbf{60.69}} & \impTwo{77.34} & \impThree{43.72} & \impTwo{\textbf{20.58}} & \impOne{\textbf{35.38}} & \impOne{\textbf{65.64}} & \impOne{\textbf{19.64}} \\
    Kimi-Audio & \impOne{27.43} & \impThree{\textbf{61.10}} & \impThree{76.96} & \impTwo{54.37} & \impOne{17.32} & \impOne{36.07} & \impOne{66.27} & \impOne{\textbf{34.78}} & \impThree{\textbf{51.21}} & \impThree{60.60} & \impThree{77.79} & \impThree{\textbf{49.18}} & \downred{14.96} & \impOne{29.81} & \impOne{63.40} & \downred{12.50} \\
    Qwen2.5-Omni & \impThree{\textbf{48.24}} & \impThree{\textbf{61.10}} & \impThree{\textbf{77.54}} & \impThree{\textbf{54.82}} & \impOne{12.69} & \impOne{29.14} & \impOne{63.13} & \impOne{15.94} & \impThree{49.55} & \impThree{59.72} & \impThree{\textbf{77.82}} & \impThree{\textbf{49.18}} & \impOne{13.82} & \impOne{30.92} & \impOne{63.83} & \downred{8.93} \\
    midasheng & \impThree{38.51} & \impThree{54.92} & \impTwo{74.44} & \impThree{53.65} & \impOne{12.51} & \impOne{28.50} & \impOne{62.40} & \impOne{20.29} & \impThree{40.01} & \impThree{55.50} & \impTwo{74.93} & \impThree{46.31} & \impOne{12.60} & \impOne{27.96} & \impOne{61.71} & \downred{7.14} \\
    \hline

    \multicolumn{17}{l}{\raisebox{0pt}[2.5ex][1.3ex]{\textbf{\textit{Base Model + Acoustic Description (AD)}}}}\\
    \hline
    Qwen2-Audio & \downred{28.13} & \impOne{46.17} & \impOne{71.29} & \impOne{40.45} & \impOne{10.49} & \impOne{29.18} & \impOne{65.80} & \impOne{31.88} & \downred{14.03} & \impOne{44.39} & \impOne{70.76} & \impOne{33.06} & \downred{8.88} & \downred{24.70} & \downred{63.52} & \downred{14.29} \\
    Qwen3-Omni & \impThree{\textbf{44.67}} & \impThree{\textbf{58.84}} & \impTwo{\textbf{74.49}} & \impTwo{35.05} & \impTwo{\textbf{17.97}} & \impTwo{\textbf{37.22}} & \impOne{\textbf{67.63}} & \impTwo{\textbf{36.96}} & \impThree{48.35} & \impThree{\textbf{59.62}} & \impTwo{75.70} & \impOne{28.42} & \impTwo{\textbf{19.02}} & \impThree{\textbf{39.14}} & \impTwo{\textbf{68.37}} & \impTwo{\textbf{28.57}} \\
    Kimi-Audio & \impOne{28.09} & \impOne{40.51} & \impOne{68.98} & \impOne{35.11} & \downred{13.72} & \impOne{25.82} & \downred{62.85} & \downred{17.39} & \impThree{38.04} & \impTwo{43.21} & \impOne{71.13} & \impOne{35.11} & \downred{14.78} & \impOne{26.70} & \impOne{63.06} & \impOne{17.86} \\
    Qwen2.5-Omni & \impThree{41.57} & \impThree{53.82} & \impTwo{74.07} & \impThree{45.72} & \impOne{14.44} & \impTwo{30.77} & \impOne{63.86} & \impOne{23.91} & \impThree{\textbf{49.87}} & \impThree{56.65} & \impTwo{\textbf{75.93}} & \impThree{\textbf{45.49}} & \impOne{18.13} & \impTwo{30.42} & \impOne{64.72} & \impOne{21.43} \\
    midasheng & \impThree{39.05} & \impThree{50.50} & \impOne{72.25} & \impThree{\textbf{46.80}} & \impOne{13.65} & \impTwo{27.25} & \impOne{61.28} & \impOne{13.77} & \impThree{40.35} & \impThree{45.66} & \impOne{70.93} & \impThree{37.98} & \impOne{13.57} & \impOne{26.81} & \impOne{62.31} & \impOne{16.07} \\
    \hline

    \multicolumn{17}{l}{\raisebox{0pt}[2.5ex][1.3ex]{\textbf{\textit{Base Model + Acoustic Description + CoT (AD+CoT)}}}}\\
    \hline
    Qwen2-Audio & \downred{24.26} & \impOne{43.26} & \impOne{69.64} & \downred{29.64} & \impOne{10.46} & \impOne{26.35} & \downred{64.53} & \downred{21.74} & \downred{25.62} & \impOne{42.37} & \impOne{69.57} & \impOne{29.64} & \impOne{10.63} & \impOne{27.25} & \impOne{64.49} & \downred{14.29} \\
    Qwen3-Omni & \impThree{\textbf{47.77}} & \impThree{\textbf{61.84}} & \impTwo{\textbf{76.87}} & \impThree{49.46} & \impTwo{\textbf{17.72}} & \impTwo{\textbf{32.99}} & \impOne{\textbf{64.72}} & \impOne{\textbf{25.36}} & \impThree{46.87} & \impThree{\textbf{60.87}} & \impTwo{76.78} & \impThree{40.44} & \impTwo{\textbf{19.79}} & \impTwo{\textbf{34.79}} & \impOne{\textbf{65.64}} & \impTwo{\textbf{26.79}} \\
    Kimi-Audio & \impThree{46.87} & \impThree{60.13} & \impTwo{76.05} & \impThree{\textbf{56.09}} & \downred{15.74} & \impOne{31.92} & \impOne{63.88} & \impOne{23.19} & \impThree{\textbf{48.61}} & \impThree{60.78} & \impThree{\textbf{76.82}} & \impThree{\textbf{50.19}} & \downred{12.71} & \impOne{27.85} & \impOne{62.69} & \downred{8.93} \\
    Qwen2.5-Omni & \impThree{40.71} & \impThree{57.29} & \impTwo{75.58} & \impThree{51.49} & \impOne{13.79} & \impTwo{30.41} & \impOne{63.32} & \impOne{16.67} & \impThree{40.43} & \impThree{56.26} & \impTwo{75.35} & \impThree{44.13} & \impOne{13.70} & \impOne{27.37} & \impOne{62.27} & \impOne{12.50} \\
    midasheng & \impThree{31.91} & \impThree{51.13} & \impOne{72.82} & \impThree{47.79} & \impOne{11.66} & \impTwo{29.03} & \impOne{62.82} & \impOne{21.74} & \impThree{35.26} & \impThree{52.10} & \impOne{73.83} & \impThree{45.40} & \downred{11.88} & \impOne{28.21} & \impOne{62.62} & \impOne{17.86} \\
    \hline

  \end{tabular}%
  }
  \caption{Full results under 3-4 speaker settings. Bold and color coding follow the same conventions as Table \ref{tab:appendix_training-free_1+2}.}
  \label{tab:appendix_training-free_3+4}
\end{table*}

We 
% first 
present training-free comparisons of different prompting strategies. Results are grouped by the number of speakers, with 1- and 2-speaker (Table \ref{tab:appendix_training-free_1+2}) and 3- and 4-speaker (Table \ref{tab:appendix_training-free_3+4}) settings reported separately to reflect the clear increase in task difficulty from smaller to larger speaker pools.

\section{Additional Details of PALM-Bench Dataset}
\label{appendix-details-of-dataset}

\subsection{Additional Qualitative Results}
\label{appendix-qualitative-results}

\begin{table*}[t]
\centering
\small
\setlength{\tabcolsep}{6pt}
\renewcommand{\arraystretch}{1.15}
\begin{tabular}{p{0.20\textwidth} p{0.72\textwidth}}
\toprule
\textbf{Model} & \textbf{Results} \\
\toprule
\multicolumn{2}{p{0.92\textwidth}}{\textbf{Question:} Is the speaker Amy?} \\
\multicolumn{2}{p{0.92\textwidth}}{\textbf{Ground truth:} No. The intonation and speaking style do not match Amy’s characteristics.} \\
\hline
\multicolumn{2}{l}{\textbf{\textit{Base Model}}} \\
\hline
Kimi-Audio & Unable to identify the speaker. \\
MiDashengLM & Unable to make a determination. \\
Qwen2-Audio & Cannot determine the speaker. \\
Qwen2.5-Omni & Unable to identify the speaker. \\
Qwen3-Omni & Unable to make a determination. \\
\hline

\multicolumn{2}{l}{\textbf{\textit{Base Model + Human Description (HD)}}} \\
\hline
Kimi-Audio & No. \\
MiDashengLM & The speaker is Amy. \\
Qwen2-Audio & Yes, the speaker is Amy. \\
Qwen2.5-Omni & Yes, the speaker is Amy. \\
Qwen3-Omni & The speaker is not Amy. \\
\hline

\multicolumn{2}{l}{\textbf{\textit{Base Model + Human Description + CoT (HD+CoT)}}} \\
\hline
\multirow{2}{*}{Kimi-Audio} &
\textless Reasoning\textgreater\ The audio contains a single speaker. The voice sounds energetic, with fast speaking rate, strong pitch variation, and prominent high-frequency components, which align with Amy’s profile. The absence of a softer and more stable female voice rules out Judy. \\
& \textless Answer\textgreater\ I believe the speaker is Amy. \\
\multirow{2}{*}{MiDashengLM} &
\textless Reasoning\textgreater\ The speaker’s clear female voice, fast pace, strong intonation variation, and prominent high-frequency features closely match Amy’s profile. \\
& \textless Answer\textgreater\ Yes, the speaker is Amy. \\
\multirow{2}{*}{Qwen2-Audio} &
\textless Reasoning\textgreater\ The speaker is a young adult female with clear articulation, fast speech, and expressive intonation, which are consistent with Amy’s profile. \\
& \textless Answer\textgreater\ Yes, the speaker is Amy. \\
\multirow{2}{*}{Qwen2.5-Omni} &
\textless Reasoning\textgreater\ The audio contains a single speaker with bright timbre, fast pace, and noticeable pitch variation, especially during emotional emphasis, matching Amy’s profile. \\
& \textless Answer\textgreater\ Yes, the speaker is Amy. \\
\multirow{2}{*}{Qwen3-Omni} &
\textless Reasoning\textgreater\ The speaker’s clear female voice, fast rhythm, strong emotional expression, and lack of hoarseness closely align with Amy’s described characteristics. \\
& \textless Answer\textgreater\ The speaker is Amy. \\
\hline

\multicolumn{2}{l}{\textbf{\textit{Base Model + Acoustic Description (AD)}}} \\
\hline
Kimi-Audio & Uncertain. \\
MiDashengLM & No. \\
Qwen2-Audio & Yes, the speaker is Amy. \\
Qwen2.5-Omni & The speaker is Amy. \\
Qwen3-Omni & The speaker is not Amy. \\
\hline

\multicolumn{2}{l}{\textbf{\textit{Base Model + Acoustic Description + CoT (AD+CoT)}}} \\
\hline
\multirow{2}{*}{Kimi-Audio} &
\textless Reasoning\textgreater\ The speaker’s fundamental frequency falls within 110--140 Hz, which is typical for adult females. The voice is bright with strong high-frequency harmonics and clear pitch variation, matching Amy’s acoustic profile. \\
& \textless Answer\textgreater\ The speaker is Amy. \\

\multirow{2}{*}{MiDashengLM} &
\textless Reasoning\textgreater\ The speaker’s acoustic features, including pitch range, harmonic structure, speaking rate, and emotional variation, closely match Amy’s profile. \\
& \textless Answer\textgreater\ Yes, the speaker is Amy. \\

\multirow{2}{*}{Qwen2-Audio} &
\textless Reasoning\textgreater\ The speaker exhibits a female pitch range of 110--140 Hz, bright timbre, fast speech, and clear articulation, all consistent with Amy’s acoustic description. \\
& \textless Answer\textgreater\ Yes, the speaker is Amy. \\

\multirow{2}{*}{Qwen2.5-Omni} &
\textless Reasoning\textgreater\ The speaker shows a stable female pitch distribution, strong high-frequency harmonics, minimal noise, and expressive intonation, which align with Amy’s acoustic profile. \\
& \textless Answer\textgreater\ Yes, the speaker is Amy. \\

\multirow{2}{*}{Qwen3-Omni} &
\textless Reasoning\textgreater\ The speaker’s pitch range, harmonic energy, fast speaking rate, and low noise level fully match Amy’s acoustic characteristics. \\
& \textless Answer\textgreater\ The speaker is Amy. \\

\bottomrule
\end{tabular}
\caption{Qualitative case study of concept activation under different prompting strategies.}
\label{tab:appendix_training-free_examples_rec_1spk}
\end{table*}
\begin{table*}[t]
\centering
\small
\setlength{\tabcolsep}{6pt}
\renewcommand{\arraystretch}{1.15}
\begin{tabular}{p{0.20\textwidth} p{0.72\textwidth}}
\toprule
\textbf{Model} & \textbf{Results} \\
\toprule
\multicolumn{2}{p{0.92\textwidth}}{\textbf{Question:} Did Amy, Bob, Catherine and Daisy all participate in this conversation?} \\
\multicolumn{2}{p{0.92\textwidth}}{\textbf{Ground truth:} No. Only Amy, Catherine and Daisy participated in the conversation.} \\
\hline

\multicolumn{2}{l}{\textbf{\textit{Base Model}}} \\
\hline
Kimi-Audio & Yes. \\
MiDashengLM & Unable to determine. \\
Qwen2.5-Omni & Unable to determine. \\
Qwen2-Audio & Based on the information provided, only two participants can be identified: one female and one male, namely Amy and Bob. \\
Qwen3-Omni & Unable to determine. \\
\hline

\multicolumn{2}{l}{\textbf{\textit{Base Model + Human Description (HD)}}} \\
\hline
Kimi-Audio & Uncertain. \\
MiDashengLM & Uncertain. \\
Qwen2.5-Omni & Yes, Amy, Bob, Catherine, and Daisy all participated in the conversation. \\
Qwen2-Audio & No. Based on the provided information, only two distinct speakers participated in the conversation. \\
Qwen3-Omni & I believe Amy, Catherine, and Daisy participated, while Bob did not appear. \\
\hline

\multicolumn{2}{l}{\textbf{\textit{Base Model + Human Description + CoT (HD+CoT)}}} \\
\hline
\multirow{2}{*}{Kimi-Audio} &
\textless Reasoning\textgreater\ The conversation includes multiple speakers whose voice characteristics align with Amy, Bob, Catherine, and Daisy. \\
& \textless Answer\textgreater\ Yes, Amy, Bob, Catherine, and Daisy all participated. \\

\multirow{2}{*}{MiDashengLM} &
\textless Reasoning\textgreater\ The audio contains two female speakers. One matches Amy’s profile, while the other aligns with Daisy’s characteristics. No male voice is present, which rules out Bob and Catherine. \\
& \textless Answer\textgreater\ Yes, Amy and Daisy participated in the conversation. \\

\multirow{2}{*}{Qwen2.5-Omni} &
\textless Reasoning\textgreater\ The audio includes two female voices and one male voice. Amy and Daisy are identified based on voice traits, while Bob is also present. No evidence supports Catherine’s participation. \\
& \textless Answer\textgreater\ Amy and Daisy participated, Bob was present, and Catherine did not participate. \\

\multirow{2}{*}{Qwen2-Audio} &
\textless Reasoning\textgreater\ Based on the dialogue content, it is not possible to determine whether all four individuals participated. \\
& \textless Answer\textgreater\ Unable to determine. \\

\multirow{2}{*}{Qwen3-Omni} &
\textless Reasoning\textgreater\ All speakers are female. Amy, Catherine, and Daisy can be identified based on voice characteristics, while no male voice is detected, excluding Bob. \\
& \textless Answer\textgreater\ No. Bob did not participate in the conversation. \\
\hline

\multicolumn{2}{l}{\textbf{\textit{Base Model + Acoustic Description (AD)}}} \\
\hline
Kimi-Audio & Uncertain. \\
MiDashengLM & Yes, Amy, Bob, Catherine, and Daisy all participated. \\
Qwen2.5-Omni & Yes, Amy, Bob, Catherine, and Daisy all participated. \\
Qwen2-Audio & No. Based on the information provided, only two speakers participated. \\
Qwen3-Omni & Bob is not included among the speakers. \\
\hline

\multicolumn{2}{l}{\textbf{\textit{Base Model + Acoustic Description + CoT (AD+CoT)}}} \\
\hline
\multirow{2}{*}{Kimi-Audio} &
\textless Reasoning\textgreater\ The detected speaker characteristics align with all four profiles based on acoustic cues. \\
& \textless Answer\textgreater\ Yes, Amy, Bob, Catherine, and Daisy all participated. \\

\multirow{2}{*}{MiDashengLM} &
\textless Reasoning\textgreater\ Two speakers are identified acoustically. One matches Amy’s profile and the other aligns with Daisy. No acoustic evidence supports Catherine or Bob. \\
& \textless Answer\textgreater\ Only Amy and Daisy participated in the conversation. \\

\multirow{2}{*}{Qwen2.5-Omni} &
\textless Reasoning\textgreater\ The dominant acoustic features closely match Daisy’s profile and do not align with Amy, Bob, or Catherine. \\
& \textless Answer\textgreater\ Only Daisy participated in the conversation. \\

\multirow{2}{*}{Qwen2-Audio} &
\textless Reasoning\textgreater\ The available information indicates the presence of a single female speaker identified as Daisy. \\
& \textless Answer\textgreater\ No. Only Daisy participated. \\

\multirow{2}{*}{Qwen3-Omni} &
\textless Reasoning\textgreater\ Four distinct acoustic patterns corresponding to Amy, Bob, Catherine, and Daisy can be identified without conflict. \\
& \textless Answer\textgreater\ Yes, Amy, Bob, Catherine, and Daisy all participated. \\

\bottomrule
\end{tabular}
\caption{Qualitative case study of concept activation under different prompting strategies.}
\label{tab:appendix_training-free_examples_rec_4spk}
\end{table*}
\begin{table*}[t]
\centering
\small
\setlength{\tabcolsep}{6pt}
\renewcommand{\arraystretch}{1.15}
\begin{tabular}{p{0.20\textwidth} p{0.72\textwidth}}
\toprule
\textbf{Model} & \textbf{Results} \\
\toprule
\multicolumn{2}{p{0.92\textwidth}}{\textbf{Question:} What is Daisy’s main point in this utterance?} \\
\multicolumn{2}{p{0.92\textwidth}}{\textbf{Ground truth:} Daisy says that she will arrive shortly and asks the other person to wait for a moment.} \\
\hline
\multicolumn{2}{l}{\textbf{\textit{Base Model}}} \\
\hline
Kimi-Audio & Daisy argues that parent--child education has a positive impact on child development and family well-being. \\
MiDashengLM & Unable to determine. \\
Qwen2.5-Omni & Unable to identify the main claim. \\
Qwen2-Audio & Daisy emphasizes the importance of parent--child education for children’s growth and future development, as well as its benefits for family harmony and happiness. \\
Qwen3-Omni & Unable to determine. \\
\hline

\multicolumn{2}{l}{\textbf{\textit{Base Model + Human Description (HD)}}} \\
\hline
Kimi-Audio & Daisy’s main claim concerns the importance of parent--child education for children’s growth and family harmony. \\
MiDashengLM & Daisy does not appear in the audio, so her claim cannot be summarized. \\
Qwen2.5-Omni & Daisy argues that child education benefits both children and families and deserves greater attention. \\
Qwen2-Audio & Daisy believes that child education positively affects children’s development and contributes to family harmony and happiness. \\
Qwen3-Omni & Daisy claims that child education fosters responsibility and confidence in children and promotes family harmony. \\
\hline

\multicolumn{2}{l}{\textbf{\textit{Base Model + Human Description + CoT (HD+CoT)}}} \\
\hline
\multirow{2}{*}{Kimi-Audio} &
\textless Reasoning\textgreater\ Daisy focuses on the importance of parent--child education for children’s development and family harmony, emphasizing the need for greater attention and effort. \\
& \textless Answer\textgreater\ Daisy claims that child education is crucial for children and family well-being. \\

\multirow{2}{*}{MiDashengLM} &
\textless Reasoning\textgreater\ The audio contains only Daisy’s voice, and the content centers on the role of parent--child education in children’s growth. \\
& \textless Answer\textgreater\ Daisy’s main claim is that parent--child education is important for children’s development and family harmony. \\

\multirow{2}{*}{Qwen2.5-Omni} &
\textless Reasoning\textgreater\ Daisy’s voice characteristics match her profile, and her speech repeatedly highlights the value of parent--child education. \\
& \textless Answer\textgreater\ Daisy argues that child education benefits children and promotes family harmony. \\

\multirow{2}{*}{Qwen2-Audio} &
\textless Reasoning\textgreater\ Daisy argues that parent--child education positively affects children’s growth and future development and supports family harmony. \\
& \textless Answer\textgreater\ Daisy advocates investing more time and effort in parent--child education. \\

\multirow{2}{*}{Qwen3-Omni} &
\textless Reasoning\textgreater\ The main content focuses on Daisy’s repeated emphasis on the benefits of parent--child education for responsibility, confidence, and family harmony. \\
& \textless Answer\textgreater\ Daisy’s core claim is that parent--child education benefits children and families and should be prioritized. \\
\hline

\multicolumn{2}{l}{\textbf{\textit{Base Model + Acoustic Description (AD)}}} \\
\hline
Kimi-Audio & Daisy argues that child education is important for children’s growth and family harmony. \\
MiDashengLM & Unable to determine. \\
Qwen2.5-Omni & Daisy claims child education benefits both children and families. \\
Qwen2-Audio & Daisy believes that child education supports children’s development and family harmony. \\
Qwen3-Omni & Daisy argues that child education promotes responsibility and family well-being. \\
\hline

\multicolumn{2}{l}{\textbf{\textit{Base Model + Acoustic Description + CoT (AD+CoT)}}} \\
\hline
\multirow{2}{*}{Kimi-Audio} &
\textless Reasoning\textgreater\ Daisy’s speech content centers on the importance of parent--child education for children’s development and family harmony. \\
& \textless Answer\textgreater\ Daisy’s core claim is the value of parent--child education for children and families. \\

\multirow{2}{*}{MiDashengLM} &
\textless Reasoning\textgreater\ The acoustic characteristics indicate Daisy as the speaker, and the content emphasizes the benefits of parent--child education. \\
& \textless Answer\textgreater\ Daisy argues that education benefits children and families. \\

\multirow{2}{*}{Qwen2.5-Omni} &
\textless Reasoning\textgreater\ The speaker’s acoustic profile matches Daisy, and her speech focuses on the role of parent--child education in child development and family harmony. \\
& \textless Answer\textgreater\ Daisy’s main claim concerns the importance of parent--child education. \\

\multirow{2}{*}{Qwen2-Audio} &
\textless Reasoning\textgreater\ Daisy highlights the positive effects of parent--child education on children’s growth and family well-being. \\
& \textless Answer\textgreater\ Parent--child education should be valued and receive greater investment. \\

\multirow{2}{*}{Qwen3-Omni} &
\textless Reasoning\textgreater\ Daisy’s speech consistently emphasizes the developmental and emotional benefits of parent--child education. \\
& \textless Answer\textgreater\ Daisy’s core claim is that parent--child education positively influences children and family harmony. \\

\bottomrule
\end{tabular}
\caption{Qualitative case study of selective understanding under different prompting strategies.}
\label{tab:appendix_training-free_examples_cap_3spk}
\end{table*}
\begin{table*}[t]
\centering
\small
\setlength{\tabcolsep}{6pt}
\renewcommand{\arraystretch}{1.15}
\begin{tabular}{p{0.13\textwidth} p{0.85\textwidth}}
\toprule
\textbf{Model} & \textbf{Results} \\
\toprule
\multicolumn{2}{p{0.92\textwidth}}{\textbf{Question:} What did Bob say?} \\
\multicolumn{2}{p{0.92\textwidth}}{\textbf{Ground truth:} Bob is asking whether the other person truly understands a certain situation and requests further explanation.} \\
\hline

\multicolumn{2}{l}{\textbf{\textit{Base Model}}} \\
\hline
Kimi-Audio & ``Bob hopes the relevant departments will solve the problem soon and reopen the platform.'' \\
MiDashengLM & ``Bob asks for clarification, expresses disappointment about a TV drama, and hopes the platform will be reopened.'' \\
Qwen2.5-Omni & Unable to recognize. \\
Qwen2-Audio & ``Bob asks for clarification, expresses disappointment about a TV drama, and hopes the platform will be reopened.'' \\
Qwen3-Omni & Unable to recognize. \\
\hline

\multicolumn{2}{l}{\textbf{\textit{Base Model + Human Description (HD)}}} \\
\hline
Kimi-Audio & ``Bob hopes the platform can be reopened for continued sharing of learning resources.'' \\
MiDashengLM & No Bob voice detected. \\
Qwen2.5-Omni & No Bob voice detected. \\
Qwen2-Audio & ``Bob asks for clarification, expresses disappointment, and hopes the platform will be reopened.'' \\
Qwen3-Omni & ``Bob hopes the platform can be reopened for continued sharing of learning resources.'' \\
\hline

\multicolumn{2}{l}{\textbf{\textit{Base Model + Human Description + CoT (HD+CoT)}}} \\
\hline
\multirow{2}{*}{Kimi-Audio} &
\textless Reasoning\textgreater\ Bob expresses disappointment about a TV drama... and calls for the platform to be reopened... \\
& \textless Answer\textgreater\ Bob hopes the platform will be reopened soon. \\
\multirow{2}{*}{MiDashengLM} &
\textless Reasoning\textgreater\ Single male speaker; steady pace and intonation... matches Bob's profile... expresses disappointment and calls for reopening... \\
& \textless Answer\textgreater\ Bob asks for clarification and expresses disappointment about a TV drama. \\
\multirow{2}{*}{Qwen2.5-Omni} &
\textless Reasoning\textgreater\ Only one steady male speaker... consistent with Bob's profile... \\
& \textless Answer\textgreater\ Bob asks for clarification and comments on a disappointing TV drama. \\
\multirow{2}{*}{Qwen2-Audio} &
\textless Reasoning\textgreater\ N/A. \\
& \textless Answer\textgreater\ Bob asks for clarification and expresses disappointment. \\
\multirow{2}{*}{Qwen3-Omni} &
\textless Reasoning\textgreater\ Single male speaker; stable, clear voice... matches Bob's persona... \\
& \textless Answer\textgreater\ Bob hopes the platform will be reopened soon. \\
\hline

\multicolumn{2}{l}{\textbf{\textit{Base Model + Acoustic Description (AD)}}} \\
\hline
Kimi-Audio & Unable to answer. \\
MiDashengLM & No Bob voice detected. \\
Qwen2.5-Omni & ``Bob asks for clarification, expresses disappointment, and hopes the platform will be reopened.'' \\
Qwen2-Audio & ``Bob asks for clarification, expresses disappointment, and hopes the platform will be reopened.'' \\
Qwen3-Omni & ``Bob hopes the platform will be reopened soon.'' \\
\hline

\multicolumn{2}{l}{\textbf{\textit{Base Model + Acoustic Description + CoT (AD+CoT)}}} \\
\hline
\multirow{2}{*}{Kimi-Audio} &
\textless Reasoning\textgreater\ N/A. \\
& \textless Answer\textgreater\ Bob hopes the platform will be reopened soon. \\
\multirow{2}{*}{MiDashengLM} &
\textless Reasoning\textgreater\ Adult male voice; F0 around 90--130 Hz... matches Bob's acoustic profile... \\
& \textless Answer\textgreater\ The speaker is Bob. \\
\multirow{2}{*}{Qwen2.5-Omni} &
\textless Reasoning\textgreater\ F0 around 100--120 Hz... bright, clean timbre... consistent with Bob... \\
& \textless Answer\textgreater\ Bob discusses reopening the platform. \\
\multirow{2}{*}{Qwen2-Audio} &
\textless Reasoning\textgreater\ Bob comments on a TV drama... and platform issues... \\
& \textless Answer\textgreater\ Bob finds the TV drama disappointing. \\
\multirow{2}{*}{Qwen3-Omni} &
\textless Reasoning\textgreater\ Third speaker matches Bob's acoustic profile... \\
& \textless Answer\textgreater\ Bob hopes the platform will be reopened soon. \\
\bottomrule
\end{tabular}
\caption{Qualitative case study of selective understanding under different prompting strategies.}
\label{tab:appendix_training-free_examples_cap_4spk}
\end{table*}
We present qualitative examples illustrating how different prompting strategies affect model behavior. Table \ref{tab:appendix_training-free_examples_rec_1spk} and Table \ref{tab:appendix_training-free_examples_rec_4spk} show two examples of concept activation (recognition) under different prompts, while Table \ref{tab:appendix_training-free_examples_cap_3spk} and Table \ref{tab:appendix_training-free_examples_cap_4spk} further present two examples of selective understanding (captioning) across the same prompting settings.

\subsection{Personalized Profiles}  
\label{appendix-profile}
To support subject-centered personalization, PALM-Bench assigns each speaker a structured profile describing stable personal characteristics beyond acoustic cues. The profiles cover basic background information, personality traits, everyday interests (e.g., media preferences and hobbies), and characteristic speaking patterns, allowing models to produce speaker-consistent responses. All profiles are derived from publicly available sources and are used only to simulate personalization scenarios without revealing sensitive personal information (Table~\ref{tab:appendix_profiles}).

\begin{table*}[t]
\centering
\small
\setlength{\tabcolsep}{8pt}
\renewcommand{\arraystretch}{1.12}
\begin{tabularx}{\textwidth}{@{} l X @{}}
\toprule
\toprule

\textbf{Speaker ID} & 0 \\
\textbf{Name} & Alice \\
\textbf{Age} & 27 \\
\textbf{Occupation} & Brand marketing coordinator \\

\textbf{Personality} & Highly empathetic and responsive; tends to validate others’ feelings early and offer reassurance. Curious and question-driven; often uses gentle probing to clarify context and support reflection. Reflective and improvement-oriented; tends to frame setbacks as learning opportunities and encourages a \textbf{\textit{growth mindset}}. Emotionally expressive at times; may use strong interjections and vivid reactions when surprised or concerned. \\

\textbf{Interests} &
\textbf{movie:} Prefers \textbf{\textit{emotional dramas}} and \textbf{\textit{human-interest stories}}; often reacts strongly to major plot turns and character loss. Uses \textbf{\textit{streaming series}} as a way to decompress after stressful days. \par

\textbf{music:} Enjoys \textbf{\textit{piano}} and \textbf{\textit{classical/pop music}}; sometimes connects music practice with \textbf{\textit{confidence-building}} and \textbf{\textit{perseverance}}. \par

\textbf{hobby:} Enjoys \textbf{\textit{trip planning}} and \textbf{\textit{local exploration}}, including itineraries and spots to visit. Maintains personal routines focused on \textbf{\textit{time management}}, \textbf{\textit{habit building}}, and \textbf{\textit{self-care}}. \par

\textbf{food:} Prefers \textbf{\textit{comfort food}} and \textbf{\textit{casual dining}}; enjoys \textbf{\textit{trying new places}} with friends. \par

\textbf{books:} Reads \textbf{\textit{light self-help}} and \textbf{\textit{practical guides}} focused on \textbf{\textit{mindset}} and \textbf{\textit{productivity}}. \par

\textbf{pet:} Generally \textbf{\textit{caring toward animals}}; tends to mention pet-related concerns in everyday contexts. \par

\textbf{sports:} Interested in \textbf{\textit{fitness}} and \textbf{\textit{functional training}}; emphasizes \textbf{\textit{sustainable}} and \textbf{\textit{consistent exercise habits}}. \par

\textbf{academic:} Interested in \textbf{\textit{everyday psychology}} (motivation, confidence, coping) and shows occasional interest in \textbf{\textit{sustainability}} and \textbf{\textit{greener lifestyles}}. \par

\textbf{tech:} Uses \textbf{\textit{productivity-oriented apps and tools}}; open to trying \textbf{\textit{new technology for practical use}}. \\

\textbf{Speaking Style} & Frequently starts with open questions and follow-up prompts to clarify details. Uses supportive phrases such as ``You’re right,'' ``I understand,'' and ``Don’t worry,'' and may add expressive interjections. Often reframes problems into \textbf{\textit{actionable next steps}}, combining empathy with gentle coaching. \\

\textbf{Intro} & Alice is a warm, inquisitive conversationalist who often acts as a \textbf{\textit{supportive facilitator}}. She listens closely, asks clarifying questions, and helps others move from emotion to action by reframing setbacks into \textbf{\textit{manageable steps}} and \textbf{\textit{learning moments}}. \\

\midrule

\textbf{Speaker ID} & 9 \\
\textbf{Name} & Jason \\
\textbf{Age} & 33 \\
\textbf{Occupation} & Sports marketing analyst \\

\textbf{Personality} & Balanced and level-headed; tends to acknowledge multiple sides before offering an opinion. Supportive and reassuring; encourages persistence and a focus on \textbf{\textit{practical next steps}}. Socially curious; enjoys discussing \textbf{\textit{communities}} and \textbf{\textit{collaboration}}, and how preferences evolve over time. \textbf{\textit{Sports-leaning}} and upbeat; may reference teams, matches, and related everyday talk. \\

\textbf{Interests} &
\textbf{movie:} Prefers \textbf{\textit{light dramas}} and \textbf{\textit{uplifting stories}} that help reset mood after a rough day. Also watches \textbf{\textit{documentaries}} connecting \textbf{\textit{culture}}, \textbf{\textit{society}}, and \textbf{\textit{real-world change}}. \par

\textbf{music:} Listens to \textbf{\textit{easy-listening music}} during commutes; often uses music for \textbf{\textit{focus}} and \textbf{\textit{mood management}}. \par

\textbf{hobby:} Uses \textbf{\textit{photography}} and \textbf{\textit{photo editing}} as creative outlets. Enjoys exploring \textbf{\textit{new cuisines}}, especially \textbf{\textit{bold and spicy flavors}}, alongside \textbf{\textit{casual city walks}}. \par

\textbf{food:} Prefers \textbf{\textit{spicy regional food}} and \textbf{\textit{comfort dishes}}; enjoys \textbf{\textit{trying new restaurants}} with friends. \par

\textbf{books:} Reads \textbf{\textit{practical non-fiction}}, including topics on \textbf{\textit{consumer behavior}}, \textbf{\textit{communication}}, and \textbf{\textit{self-improvement}}. \par

\textbf{pet:} Generally \textbf{\textit{fond of animals}}; tends to discuss pets in everyday, caring terms. \par

\textbf{sports:} Enjoys \textbf{\textit{soccer fandom}} and \textbf{\textit{match discussions}}; follows \textbf{\textit{major clubs}} and \textbf{\textit{key fixtures}}. \par

\textbf{academic:} Interested in \textbf{\textit{behavioral insights}} and \textbf{\textit{social trends}}, focusing on what drives people’s choices. \par

\textbf{tech:} Uses \textbf{\textit{social media platforms}} and \textbf{\textit{analytics tools}}; curious about how \textbf{\textit{online communities}} form and evolve. \\

\textbf{Speaking Style} & Measured and supportive. Often begins with calm confirmations such as ``I agree'' or ``That makes sense,'' then adds a \textbf{\textit{practical suggestion}} or reframing. Comfortable switching between everyday topics and broader discussions on \textbf{\textit{society}} and \textbf{\textit{consumer trends}}. \\

\textbf{Intro} & Jason is a steady, supportive conversationalist with a \textbf{\textit{practical orientation}}. He balances empathy with action, encourages persistence, and frequently connects discussions to \textbf{\textit{sports}}, \textbf{\textit{photography}}, and \textbf{\textit{community dynamics}}. \\

\bottomrule
\bottomrule

\end{tabularx}
\caption{Example profiles used in PALM-Bench. \textbf{\textit{Bold italic text}} highlights attributes related to personalized information that support personalized reasoning.}
\label{tab:appendix_profiles}
\end{table*}

\subsection{QA Template}
\label{appendix-qa template}

\begin{table*}[t]
\centering
\small
\setlength{\tabcolsep}{6pt}
\renewcommand{\arraystretch}{1.05}
\begin{tabularx}{\textwidth}{c >{\raggedright\arraybackslash}X >{\raggedright\arraybackslash}X}
\toprule

\multicolumn{3}{c}{\textbf{Recognition}} \\
\midrule

\multicolumn{3}{c}{\textbf{\textit{Positive Samples}}} \\
\midrule
\textbf{Index} & \textbf{Question} & \textbf{Answer} \\
\midrule

1  & Can you help me confirm whether \textbf{\{spk\}} spoke?
   & Confirmed—\textcolor{green}{\textbf{\{spk\}}} did speak. \\

2  & Is \textbf{\{spk\}} in this audio clip?
   & Yes. I can clearly tell it’s \textcolor{green}{\textbf{\{spk\}}}’s voice. \\

3  & Does \textbf{\{spk\}} appear in the audio?
   & Yes—\textcolor{green}{\textbf{\{spk\}}}’s voice is very noticeable in this clip. \\

4  & Is \textbf{\{spk\}} in this audio recording?
   & Yes. That familiar voice is unmistakably \textcolor{green}{\textbf{\{spk\}}}. \\

5  & Is \textbf{\{spk\}}’s voice present in this recording?
   & It is. I’m very familiar with \textcolor{green}{\textbf{\{spk\}}}’s voice, so I can confirm. \\

6  & Does this audio file include any speech from \textbf{\{spk\}}?
   & Yes—and the tone and speaking pace are very much \textcolor{green}{\textbf{\{spk\}}}’s style. \\

7  & Does the audio content contain \textbf{\{spk\}}’s voice?
   & It does. \textcolor{green}{\textbf{\{spk\}}} speaks quite a bit, and the voice is highly recognizable. \\

8  & Does \textbf{\{spk\}} show up in this audio clip?
   & Yes. \textcolor{green}{\textbf{\{spk\}}} took part in the conversation—you can hear it clearly. \\

9  & Is this clip \textbf{\{spk\}} speaking?
   & Yes, it’s \textcolor{green}{\textbf{\{spk\}}} speaking. It’s very clear. \\

10 & Is the speaker \textbf{\{spk\}}?
   & Exactly—it’s \textcolor{green}{\textbf{\{spk\}}}. Their voice is very distinctive. \\

11 & Is the person speaking in the audio \textbf{\{spk\}}?
   & Yes, the speaker is \textcolor{green}{\textbf{\{spk\}}}. \\

12 & Is the voice in this recording from \textbf{\{spk\}}?
   & It’s \textcolor{green}{\textbf{\{spk\}}} in person—I can confirm that. \\

13 & Is the person speaking \textbf{\{spk\}}?
   & Yes—it’s \textcolor{green}{\textbf{\{spk\}}}. The voice is very recognizable. \\

14 & Is the person making the sound \textbf{\{spk\}}?
   & Yes, the person speaking is \textcolor{green}{\textbf{\{spk\}}}. \\

15 & Is this voice \textbf{\{spk\}}’s?
   & Yes—this is \textcolor{green}{\textbf{\{spk\}}}’s voice. \\

16 & Can you hear \textbf{\{spk\}}’s voice in this audio?
   & Absolutely. \textcolor{green}{\textbf{\{spk\}}}’s voice is very distinctive. \\

17 & Can you identify \textbf{\{spk\}} from this audio?
   & Yes—and I can tell \textcolor{green}{\textbf{\{spk\}}} said quite a lot. \\

18 & Can you hear \textbf{\{spk\}} speaking in this recording?
   & Yes. \textcolor{green}{\textbf{\{spk\}}}’s tone and speaking pace are very familiar. \\

19 & Can you tell from this audio that \textbf{\{spk\}} is speaking?
   & Yes. It’s \textcolor{green}{\textbf{\{spk\}}} sharing their views. \\

20 & Can you confirm that \textbf{\{spk\}} is in this audio?
   & I’m completely sure—it’s definitely \textcolor{green}{\textbf{\{spk\}}}’s voice. \\

\addlinespace[0.6em]
\midrule
\multicolumn{3}{c}{\textbf{Negative Samples}} \\
\midrule
\textbf{Index} & \textbf{Question} & \textbf{Answer} \\
\midrule

1  & Is \textbf{\{spk\}} in this audio clip?
   & I didn’t hear \textcolor{red}{\textbf{\{spk\}}}’s voice. \\

2  & Does \textbf{\{spk\}} appear in the audio?
   & \textcolor{red}{\textbf{\{spk\}}}’s voice wasn’t detected in the audio. \\

3  & Is \textbf{\{spk\}} in this audio clip?
   & It doesn’t sound like \textcolor{red}{\textbf{\{spk\}}}. \\

4  & Is \textbf{\{spk\}}’s voice present in this recording?
   & I didn’t clearly hear \textcolor{red}{\textbf{\{spk\}}}’s voice. \\

5  & Does this audio file include \textbf{\{spk\}} speaking?
   & It doesn’t seem like \textcolor{red}{\textbf{\{spk\}}} speaks in the audio. \\

6  & Does the audio content include \textbf{\{spk\}}’s voice?
   & No—it doesn’t include \textcolor{red}{\textbf{\{spk\}}}. It sounds like someone else. \\

7  & Did \textbf{\{spk\}} appear in this audio clip?
   & \textcolor{red}{\textbf{\{spk\}}} did not appear in the audio. \\

8  & Is \textbf{\{spk\}} the one speaking in this audio clip?
   & No, the voice doesn’t sound like \textcolor{red}{\textbf{\{spk\}}}. \\

9  & Is the speaker \textbf{\{spk\}}?
   & No, the tone and style don’t really match \textcolor{red}{\textbf{\{spk\}}}. \\

10 & Is the person speaking in the audio \textbf{\{spk\}}?
   & The speaker is likely not \textcolor{red}{\textbf{\{spk\}}}. \\

11 & Is the voice in this recording from \textbf{\{spk\}}?
   & I don’t think it’s \textcolor{red}{\textbf{\{spk\}}}. \\

12 & Is \textbf{\{spk\}} the one speaking?
   & It doesn’t sound like \textcolor{red}{\textbf{\{spk\}}}. \\

13 & Is the person making the sound \textbf{\{spk\}}?
   & It doesn’t really sound like \textcolor{red}{\textbf{\{spk\}}}. \\

14 & Can you help me confirm whether \textbf{\{spk\}} spoke?
   & Confirmed—\textcolor{red}{\textbf{\{spk\}}} didn’t speak. \\

15 & Is this voice \textbf{\{spk\}}’s?
   & This isn’t \textcolor{red}{\textbf{\{spk\}}}’s voice. \\

16 & Can you hear \textbf{\{spk\}}’s voice in this audio clip?
   & No, \textcolor{red}{\textbf{\{spk\}}}’s voice doesn’t appear. \\

17 & Can you identify \textbf{\{spk\}} from this audio?
   & No, I can’t make out \textcolor{red}{\textbf{\{spk\}}}’s voice. \\

18 & Can you hear \textbf{\{spk\}} speaking in this recording?
   & I can’t hear \textcolor{red}{\textbf{\{spk\}}} speaking. \\

19 & Can you tell if \textbf{\{spk\}} is the one speaking in this audio?
   & I can’t tell—it’s not \textcolor{red}{\textbf{\{spk\}}}. \\

20 & Can you be sure \textbf{\{spk\}} is in this audio clip?
   & I’m very sure this isn’t \textcolor{red}{\textbf{\{spk\}}}’s voice. \\

\bottomrule
\end{tabularx}
\caption{Recognition QA templates. \textbf{Bold text} highlights the speaker placeholder. \textbf{\textcolor{green}{Green}} indicates present speakers (positive samples), while \textbf{\textcolor{red}{red}} indicates absent speakers (negative samples).}
\label{tab:qa_templates_single_recognition}
\end{table*}
\begin{table*}[t]
\centering
\small
\setlength{\tabcolsep}{6pt}
\renewcommand{\arraystretch}{1.05}
\begin{tabularx}{\textwidth}{c >{\raggedright\arraybackslash}X >{\raggedright\arraybackslash}X}
\toprule

\multicolumn{3}{c}{\textbf{\textit{Captioning}}} \\
\midrule

\multicolumn{3}{c}{\textbf{Positive Samples}} \\
\midrule
\textbf{Index} & \textbf{Question} &  \\
\midrule

1  & Could you help me summarize what \textbf{\{spk\}} said?
   & \\

2  & What did \textbf{\{spk\}} say in this audio clip?
   & \\

3  & What are the main points of \textbf{\{spk\}}’s remarks in this segment?
   & \\

4  & Summarize what \textbf{\{spk\}} said in the audio.
   & \\

5  & What was \textbf{\{spk\}} trying to express in the audio?
   & \\

6  & What is \textbf{\{spk\}} talking about in the audio?
   & \\

7  & Summarize what \textbf{\{spk\}} was trying to say.
   & \\

8  & What key information did \textbf{\{spk\}} mention in this audio?
   & \\

9  & What did \textbf{\{spk\}} say?
   & \\

10 & What did \textbf{\{spk\}} talk about in this conversation?
   & \\

11 & What exactly did \textbf{\{spk\}} say in the audio?
   & \\

\addlinespace[0.6em]
\midrule
\multicolumn{3}{c}{\textbf{Negative Samples}} \\
\midrule
\textbf{Index} & \textbf{Question} & \textbf{Answer} \\
\midrule

1  & Can you help me summarize what \textbf{\{spk\}} said?
   & \textcolor{red}{\textbf{\{spk\}}} doesn’t speak in this audio, so there’s nothing to summarize. \\

2  & What did \textbf{\{spk\}} say in this audio?
   & \textcolor{red}{\textbf{\{spk\}}} isn’t in this audio, so there’s nothing he said here. \\

3  & What is the main point of \textbf{\{spk\}}’s statement?
   & The speaker isn’t \textcolor{red}{\textbf{\{spk\}}}, so there’s no relevant content. \\

4  & Summarize what \textbf{\{spk\}} said in the audio.
   & \textcolor{red}{\textbf{\{spk\}}}’s voice isn’t in the audio, so I can’t summarize it. \\

5  & What did \textbf{\{spk\}} express in the audio?
   & This audio doesn’t involve \textcolor{red}{\textbf{\{spk\}}}, so there’s nothing from him to describe. \\

6  & What is \textbf{\{spk\}} talking about in the audio?
   & \textcolor{red}{\textbf{\{spk\}}} doesn’t take part in the discussion in this audio, so there’s no related content. \\

7  & Summarize what \textbf{\{spk\}} was trying to say.
   & The person speaking isn’t \textcolor{red}{\textbf{\{spk\}}}, so I can’t summarize what he meant to say. \\

8  & What key information did \textbf{\{spk\}} mention in this audio?
   & \textcolor{red}{\textbf{\{spk\}}} doesn’t speak in the audio, so there’s no key information from him. \\

9  & What did \textbf{\{spk\}} say?
   & \textcolor{red}{\textbf{\{spk\}}} isn’t in this audio, so he didn’t say anything. \\

10 & What did \textbf{\{spk\}} say in this conversation?
   & \textcolor{red}{\textbf{\{spk\}}} doesn’t speak in this conversation, so there’s no related content. \\

11 & What exactly did \textbf{\{spk\}} say in the audio?
   & \textcolor{red}{\textbf{\{spk\}}}’s voice isn’t in the audio, so he didn’t say anything. \\

\bottomrule
\end{tabularx}
\caption{Captioning QA templates. Positive samples provide question templates only. \textbf{Bold text} highlights the speaker placeholder, and \textbf{\textcolor{red}{red}} indicates an absent speaker.}
\label{tab:qa_templates_single_caption}
\end{table*}
\begin{table*}[t]
\centering
\small
\setlength{\tabcolsep}{6pt}
\renewcommand{\arraystretch}{1.05}
\begin{tabularx}{\textwidth}{c >{\raggedright\arraybackslash}X >{\raggedright\arraybackslash}X}
\toprule
\multicolumn{3}{c}{\textbf{\textit{Recognition (Two Speakers: A+B)}}} \\
\midrule

\multicolumn{3}{c}{\textbf{Case 1: Both Present (A and B)}} \\
\midrule
\textbf{Index} & \textbf{Question} & \textbf{Answer} \\
\midrule

1  & Did \textbf{\{A\}} and \textbf{\{B\}} both appear in this conversation?
   & Yes, both \textcolor{green}{\textbf{\{A\}}} and \textcolor{green}{\textbf{\{B\}}} are present. \\
2  & Are \textbf{\{A\}} and \textbf{\{B\}} in this conversation?
   & Yes, it includes both \textcolor{green}{\textbf{\{A\}}} and \textcolor{green}{\textbf{\{B\}}}. \\
3  & Can you hear both \textbf{\{A\}} and \textbf{\{B\}} in this dialogue?
   & Yes. Both \textcolor{green}{\textbf{\{A\}}} and \textcolor{green}{\textbf{\{B\}}} are speaking in this clip. \\
4  & Does this dialogue include \textbf{\{A\}} and \textbf{\{B\}}?
   & Yes, you can hear the voices of \textcolor{green}{\textbf{\{A\}}} and \textcolor{green}{\textbf{\{B\}}}. \\
5  & It sounds like both \textbf{\{A\}} and \textbf{\{B\}} spoke, right?
   & Correct. Both \textcolor{green}{\textbf{\{A\}}} and \textcolor{green}{\textbf{\{B\}}} spoke. \\
6  & Are \textbf{\{A\}} and \textbf{\{B\}} talking to each other?
   & Yes, it is \textcolor{green}{\textbf{\{A\}}} and \textcolor{green}{\textbf{\{B\}}} in conversation. \\
7  & This recording sounds like two people—one is \textbf{\{A\}} and the other is \textbf{\{B\}}, right?
   & Exactly. You can hear both \textcolor{green}{\textbf{\{A\}}} and \textcolor{green}{\textbf{\{B\}}}. \\
8  & Please confirm: are the speakers \textbf{\{A\}} and \textbf{\{B\}}?
   & Confirmed—the two speakers are \textcolor{green}{\textbf{\{A\}}} and \textcolor{green}{\textbf{\{B\}}}. \\
9  & I hear two different voices—are they \textbf{\{A\}} and \textbf{\{B\}}?
   & Yes, they are \textcolor{green}{\textbf{\{A\}}} and \textcolor{green}{\textbf{\{B\}}}. \\
10 & Is \textbf{\{A\}} speaking in this audio, and do you also hear \textbf{\{B\}}?
   & Yes, \textcolor{green}{\textbf{\{B\}}} is also present. \\
11 & Is \textbf{\{B\}} speaking in this audio, and do you also hear \textbf{\{A\}}?
   & Yes, \textcolor{green}{\textbf{\{A\}}} is also present. \\
12 & Is \textbf{\{A\}} speaking in this audio?
   & Yes, \textcolor{green}{\textbf{\{A\}}} is speaking—very clearly. \\
13 & Can you identify \textbf{\{A\}}’s voice?
   & Yes, \textcolor{green}{\textbf{\{A\}}} definitely spoke. \\
14 & Did \textbf{\{A\}} participate in this conversation?
   & Yes, \textcolor{green}{\textbf{\{A\}}} is one of the participants. \\
15 & Is \textbf{\{A\}}’s voice in the audio?
   & Yes, you can hear \textcolor{green}{\textbf{\{A\}}} speaking. \\
16 & I think I heard \textbf{\{A\}}—is that right?
   & Yes, \textcolor{green}{\textbf{\{A\}}} participated in this conversation. \\
17 & Did \textbf{\{A\}} speak during this conversation?
   & Yes, \textcolor{green}{\textbf{\{A\}}} spoke in the conversation. \\
18 & Is \textbf{\{A\}} among the speakers?
   & Yes, \textcolor{green}{\textbf{\{A\}}} is included—the voice is very distinctive. \\
19 & Can you help confirm whether \textbf{\{A\}} spoke?
   & Confirmed—\textcolor{green}{\textbf{\{A\}}} did speak. \\
20 & Is \textbf{\{B\}} speaking in this audio?
   & Yes, \textcolor{green}{\textbf{\{B\}}} is speaking—very clearly. \\
21 & Can you identify \textbf{\{B\}}’s voice?
   & Yes, \textcolor{green}{\textbf{\{B\}}} definitely spoke. \\
22 & Did \textbf{\{B\}} participate in this conversation?
   & Yes, \textcolor{green}{\textbf{\{B\}}} is one of the participants. \\
23 & Is \textbf{\{B\}}’s voice in the audio?
   & Yes, you can hear \textcolor{green}{\textbf{\{B\}}} speaking. \\
24 & I think I heard \textbf{\{B\}}—is that right?
   & Yes, \textcolor{green}{\textbf{\{B\}}} participated in this conversation. \\
25 & Did \textbf{\{B\}} speak during this conversation?
   & Yes, \textcolor{green}{\textbf{\{B\}}} spoke in the conversation. \\
26 & Is \textbf{\{B\}} among the speakers?
   & Yes, \textcolor{green}{\textbf{\{B\}}} is included—the voice is very distinctive. \\
27 & Can you help confirm whether \textbf{\{B\}} spoke?
   & Confirmed—\textcolor{green}{\textbf{\{B\}}} did speak. \\

\addlinespace[0.6em]
\midrule
\multicolumn{3}{c}{\textbf{Case 2: Only A Present (A present, B absent)}} \\
\midrule
\textbf{Index} & \textbf{Question} & \textbf{Answer} \\
\midrule

1  & Did \textbf{\{A\}} and \textbf{\{B\}} both appear in this conversation?
   & Only \textcolor{green}{\textbf{\{A\}}} is present; \textcolor{red}{\textbf{\{B\}}} is not. \\
2  & Are \textbf{\{A\}} and \textbf{\{B\}} in this conversation?
   & It only includes \textcolor{green}{\textbf{\{A\}}}’s voice. \\
3  & Can you hear both \textbf{\{A\}} and \textbf{\{B\}} in this dialogue?
   & You can only hear \textcolor{green}{\textbf{\{A\}}}; \textcolor{red}{\textbf{\{B\}}} is not present. \\
4  & Does this dialogue include \textbf{\{A\}} and \textbf{\{B\}}?
   & It only contains \textcolor{green}{\textbf{\{A\}}}’s voice. \\
5  & It sounds like both \textbf{\{A\}} and \textbf{\{B\}} spoke, right?
   & No—only \textcolor{green}{\textbf{\{A\}}} spoke. \\
6  & Are \textbf{\{A\}} and \textbf{\{B\}} talking to each other?
   & It is \textcolor{green}{\textbf{\{A\}}} talking with someone else, not \textcolor{red}{\textbf{\{B\}}}. \\
7  & Are both \textbf{\{A\}} and \textbf{\{B\}} speaking in this audio?
   & No. Only \textcolor{green}{\textbf{\{A\}}} participated; \textcolor{red}{\textbf{\{B\}}} did not. \\
8  & This recording sounds like two people—one is \textbf{\{A\}} and the other is \textbf{\{B\}}, right?
   & Not quite—it’s \textcolor{green}{\textbf{\{A\}}} and another person, not \textcolor{red}{\textbf{\{B\}}}. \\
9  & Please confirm: are the speakers \textbf{\{A\}} and \textbf{\{B\}}?
   & I think it’s \textcolor{green}{\textbf{\{A\}}} and someone else; there is no \textcolor{red}{\textbf{\{B\}}} voice. \\
10 & I hear two different voices—are they \textbf{\{A\}} and \textbf{\{B\}}?
   & It’s \textcolor{green}{\textbf{\{A\}}} and someone else, not \textcolor{red}{\textbf{\{B\}}}. \\
11 & Is \textbf{\{A\}} speaking in this audio?
   & Yes, \textcolor{green}{\textbf{\{A\}}} is speaking—very clearly. \\
12 & Can you identify \textbf{\{A\}}’s voice?
   & Yes, \textcolor{green}{\textbf{\{A\}}} definitely spoke. \\
13 & Did \textbf{\{A\}} participate in this conversation?
   & Yes, \textcolor{green}{\textbf{\{A\}}} is one of the participants. \\
14 & Is \textbf{\{A\}}’s voice in the audio?
   & Yes, you can hear \textcolor{green}{\textbf{\{A\}}} speaking. \\
15 & I think I heard \textbf{\{A\}}—is that right?
   & Yes, \textcolor{green}{\textbf{\{A\}}} also participated in this conversation. \\
16 & Did \textbf{\{A\}} speak during this conversation?
   & Yes, \textcolor{green}{\textbf{\{A\}}} also spoke in the conversation. \\
17 & Is \textbf{\{A\}} among the speakers?
   & Yes, \textcolor{green}{\textbf{\{A\}}} is included—the voice is very distinctive. \\
18 & Can you help confirm whether \textbf{\{A\}} spoke?
   & Confirmed—\textcolor{green}{\textbf{\{A\}}} did speak. \\
19 & Is \textbf{\{A\}} speaking in this audio, and do you also hear \textbf{\{B\}}?
   & No, \textcolor{red}{\textbf{\{B\}}}’s voice is not present. \\
20 & Is \textbf{\{B\}} speaking in this audio?
   & No, \textcolor{red}{\textbf{\{B\}}}’s voice is not present. \\
21 & Can you identify \textbf{\{B\}}’s voice?
   & I can’t identify \textcolor{red}{\textbf{\{B\}}}’s voice here. \\
22 & Did \textbf{\{B\}} participate in this conversation?
   & No, \textcolor{red}{\textbf{\{B\}}} did not participate. \\
23 & Is \textbf{\{B\}}’s voice in the audio?
   & No, \textcolor{red}{\textbf{\{B\}}}’s voice is not present. \\
24 & I think I heard \textbf{\{B\}}—is that right?
   & No, \textcolor{red}{\textbf{\{B\}}}’s voice did not appear. \\
25 & Did \textbf{\{B\}} speak during this conversation?
   & No, \textcolor{red}{\textbf{\{B\}}} did not speak. \\
26 & Is \textbf{\{B\}} among the speakers?
   & No, \textcolor{red}{\textbf{\{B\}}} is not included. \\
27 & Can you help confirm whether \textbf{\{B\}} spoke?
   & \textcolor{red}{\textbf{\{B\}}} did not speak. \\

\bottomrule
\end{tabularx}
\caption{Multi-speaker recognition QA templates for a two-speaker candidate set \{\{A\},\{B\}\}: Case 1 and Case 2. \textbf{\textcolor{green}{Green}} indicates present speakers and \textbf{\textcolor{red}{red}} indicates absent speakers.}
\label{tab:ms_recognition_2spk_case12}
\end{table*}

\begin{table*}[t]
\centering
\small
\setlength{\tabcolsep}{6pt}
\renewcommand{\arraystretch}{1.05}
\begin{tabularx}{\textwidth}{c >{\raggedright\arraybackslash}X >{\raggedright\arraybackslash}X}
\toprule
\multicolumn{3}{c}{\textbf{\textit{Recognition (Two Speakers: A+B)}}} \\
\midrule

\multicolumn{3}{c}{\textbf{Case 3: Only B Present (B present, A absent)}} \\
\midrule
\textbf{Index} & \textbf{Question} & \textbf{Answer} \\
\midrule

1  & Did \textbf{\{A\}} and \textbf{\{B\}} both appear in this conversation?
   & Only \textcolor{green}{\textbf{\{B\}}} is present; \textcolor{red}{\textbf{\{A\}}} is not. \\
2  & Are \textbf{\{A\}} and \textbf{\{B\}} in this conversation?
   & It only includes \textcolor{green}{\textbf{\{B\}}}’s voice. \\
3  & Can you hear both \textbf{\{A\}} and \textbf{\{B\}} in this dialogue?
   & You can only hear \textcolor{green}{\textbf{\{B\}}}; \textcolor{red}{\textbf{\{A\}}} is not present. \\
4  & Does this dialogue include \textbf{\{A\}} and \textbf{\{B\}}?
   & It only contains \textcolor{green}{\textbf{\{B\}}}’s voice. \\
5  & It sounds like both \textbf{\{A\}} and \textbf{\{B\}} spoke, right?
   & No—only \textcolor{green}{\textbf{\{B\}}} spoke. \\
6  & Are \textbf{\{A\}} and \textbf{\{B\}} talking to each other?
   & It is \textcolor{green}{\textbf{\{B\}}} talking with someone else, not \textcolor{red}{\textbf{\{A\}}}. \\
7  & Are both \textbf{\{A\}} and \textbf{\{B\}} speaking in this audio?
   & No. Only \textcolor{green}{\textbf{\{B\}}} participated; \textcolor{red}{\textbf{\{A\}}} did not. \\
8  & This recording sounds like two people—one is \textbf{\{A\}} and the other is \textbf{\{B\}}, right?
   & Not quite—it’s \textcolor{green}{\textbf{\{B\}}} and another person, not \textcolor{red}{\textbf{\{A\}}}. \\
9  & Please confirm: are the speakers \textbf{\{A\}} and \textbf{\{B\}}?
   & I think it’s \textcolor{green}{\textbf{\{B\}}} and someone else; there is no \textcolor{red}{\textbf{\{A\}}} voice. \\
10 & I hear two different voices—are they \textbf{\{A\}} and \textbf{\{B\}}?
   & It’s \textcolor{green}{\textbf{\{B\}}} and someone else, not \textcolor{red}{\textbf{\{A\}}}. \\
11 & Is \textbf{\{B\}} speaking in this audio?
   & Yes, \textcolor{green}{\textbf{\{B\}}} is speaking—very clearly. \\
12 & Can you identify \textbf{\{B\}}’s voice?
   & Yes, \textcolor{green}{\textbf{\{B\}}} definitely spoke. \\
13 & Did \textbf{\{B\}} participate in this conversation?
   & Yes, \textcolor{green}{\textbf{\{B\}}} is one of the participants. \\
14 & Is \textbf{\{B\}}’s voice in the audio?
   & Yes, you can hear \textcolor{green}{\textbf{\{B\}}} speaking. \\
15 & I think I heard \textbf{\{B\}}—is that right?
   & Yes, \textcolor{green}{\textbf{\{B\}}} also participated in this conversation. \\
16 & Did \textbf{\{B\}} speak during this conversation?
   & Yes, \textcolor{green}{\textbf{\{B\}}} also spoke in the conversation. \\
17 & Is \textbf{\{B\}} among the speakers?
   & Yes, \textcolor{green}{\textbf{\{B\}}} is included—the voice is very distinctive. \\
18 & Can you help confirm whether \textbf{\{B\}} spoke?
   & Confirmed—\textcolor{green}{\textbf{\{B\}}} did speak. \\
19 & Is \textbf{\{B\}} speaking in this audio, and do you also hear \textbf{\{A\}}?
   & No, \textcolor{red}{\textbf{\{A\}}}’s voice is not present. \\
20 & Is \textbf{\{A\}} speaking in this audio?
   & No, \textcolor{red}{\textbf{\{A\}}}’s voice is not present. \\
21 & Can you identify \textbf{\{A\}}’s voice?
   & I can’t identify \textcolor{red}{\textbf{\{A\}}}’s voice here. \\
22 & Did \textbf{\{A\}} participate in this conversation?
   & No, \textcolor{red}{\textbf{\{A\}}} did not participate. \\
23 & Is \textbf{\{A\}}’s voice in the audio?
   & No, \textcolor{red}{\textbf{\{A\}}}’s voice is not present. \\
24 & I think I heard \textbf{\{A\}}—is that right?
   & No, \textcolor{red}{\textbf{\{A\}}}’s voice did not appear. \\
25 & Did \textbf{\{A\}} speak during this conversation?
   & No, \textcolor{red}{\textbf{\{A\}}} did not speak. \\
26 & Is \textbf{\{A\}} among the speakers?
   & No, \textcolor{red}{\textbf{\{A\}}} is not included. \\
27 & Can you help confirm whether \textbf{\{A\}} spoke?
   & \textcolor{red}{\textbf{\{A\}}} did not speak. \\

\addlinespace[0.6em]
\midrule
\multicolumn{3}{c}{\textbf{Case 4: None Present (A absent, B absent)}} \\
\midrule
\textbf{Index} & \textbf{Question} & \textbf{Answer} \\
\midrule

1  & Did \textbf{\{A\}} and \textbf{\{B\}} both appear in this conversation?
   & Neither \textcolor{red}{\textbf{\{A\}}} nor \textcolor{red}{\textbf{\{B\}}} is present. \\
2  & Are \textbf{\{A\}} and \textbf{\{B\}} in this conversation?
   & No—neither \textcolor{red}{\textbf{\{A\}}} nor \textcolor{red}{\textbf{\{B\}}} is detected. \\
3  & Can you hear both \textbf{\{A\}} and \textbf{\{B\}} in this dialogue?
   & No. You can’t hear \textcolor{red}{\textbf{\{A\}}} or \textcolor{red}{\textbf{\{B\}}} in this clip. \\
4  & Does this dialogue include \textbf{\{A\}} and \textbf{\{B\}}?
   & No, it includes neither \textcolor{red}{\textbf{\{A\}}} nor \textcolor{red}{\textbf{\{B\}}}. \\
5  & It sounds like both \textbf{\{A\}} and \textbf{\{B\}} spoke, right?
   & No—neither \textcolor{red}{\textbf{\{A\}}} nor \textcolor{red}{\textbf{\{B\}}} spoke. \\
6  & Are \textbf{\{A\}} and \textbf{\{B\}} talking to each other?
   & No. It’s not \textcolor{red}{\textbf{\{A\}}} and \textcolor{red}{\textbf{\{B\}}}; it’s two different speakers. \\
7  & Are both \textbf{\{A\}} and \textbf{\{B\}} speaking in this audio?
   & No—it’s not \textcolor{red}{\textbf{\{A\}}} or \textcolor{red}{\textbf{\{B\}}} speaking. \\
8  & This recording sounds like two people—one is \textbf{\{A\}} and the other is \textbf{\{B\}}, right?
   & No, those two speakers are not \textcolor{red}{\textbf{\{A\}}} and \textcolor{red}{\textbf{\{B\}}}. \\
9  & Please confirm: are the speakers \textbf{\{A\}} and \textbf{\{B\}}?
   & I don’t think it’s \textcolor{red}{\textbf{\{A\}}} or \textcolor{red}{\textbf{\{B\}}}. \\
10 & I hear two different voices—are they \textbf{\{A\}} and \textbf{\{B\}}?
   & No, they are not \textcolor{red}{\textbf{\{A\}}} and \textcolor{red}{\textbf{\{B\}}}. \\
11 & Is \textbf{\{A\}} speaking in this audio?
   & No, \textcolor{red}{\textbf{\{A\}}}’s voice is not present. \\
12 & Can you identify \textbf{\{A\}}’s voice?
   & I can’t identify \textcolor{red}{\textbf{\{A\}}}’s voice here. \\
13 & Did \textbf{\{A\}} participate in this conversation?
   & No, \textcolor{red}{\textbf{\{A\}}} did not participate. \\
14 & Is \textbf{\{A\}}’s voice in the audio?
   & No, \textcolor{red}{\textbf{\{A\}}}’s voice is not present. \\
15 & I think I heard \textbf{\{A\}}—is that right?
   & No, \textcolor{red}{\textbf{\{A\}}}’s voice did not appear. \\
16 & Did \textbf{\{A\}} speak during this conversation?
   & No, \textcolor{red}{\textbf{\{A\}}} did not speak. \\
17 & Is \textbf{\{A\}} among the speakers?
   & No, \textcolor{red}{\textbf{\{A\}}} is not included. \\
18 & Can you help confirm whether \textbf{\{A\}} spoke?
   & \textcolor{red}{\textbf{\{A\}}} did not speak. \\
19 & Is \textbf{\{B\}} speaking in this audio?
   & No, \textcolor{red}{\textbf{\{B\}}}’s voice is not present. \\
20 & Can you identify \textbf{\{B\}}’s voice?
   & I can’t identify \textcolor{red}{\textbf{\{B\}}}’s voice here. \\
21 & Did \textbf{\{B\}} participate in this conversation?
   & No, \textcolor{red}{\textbf{\{B\}}} did not participate. \\
22 & Is \textbf{\{B\}}’s voice in the audio?
   & No, \textcolor{red}{\textbf{\{B\}}}’s voice is not present. \\
23 & I think I heard \textbf{\{B\}}—is that right?
   & No, \textcolor{red}{\textbf{\{B\}}}’s voice did not appear. \\
24 & Did \textbf{\{B\}} speak during this conversation?
   & No, \textcolor{red}{\textbf{\{B\}}} did not speak. \\
25 & Is \textbf{\{B\}} among the speakers?
   & No, \textcolor{red}{\textbf{\{B\}}} is not included. \\
26 & Can you help confirm whether \textbf{\{B\}} spoke?
   & \textcolor{red}{\textbf{\{B\}}} did not speak. \\

\bottomrule
\end{tabularx}
\caption{Multi-speaker recognition QA templates for a two-speaker candidate set \{\{A\},\{B\}\}: Case 3 and Case 4. \textbf{\textcolor{green}{Green}} indicates present speakers and \textbf{\textcolor{red}{red}} indicates absent speakers.}
\label{tab:ms_recognition_2spk_case34}
\end{table*}

\begin{table*}[t]
\centering
\small
\setlength{\tabcolsep}{6pt}
\renewcommand{\arraystretch}{1.05}
\begin{tabularx}{\textwidth}{c >{\raggedright\arraybackslash}X >{\raggedright\arraybackslash}X}
\toprule
\multicolumn{3}{c}{\textbf{\textit{Recognition (Three Speakers: A+B+C)}}} \\
\midrule

\multicolumn{3}{c}{\textbf{Case 1: All Present (A, B, and C)}} \\
\midrule
\textbf{Index} & \textbf{Question} & \textbf{Answer} \\
\midrule

1  & Among \textbf{\{A\}}, \textbf{\{B\}}, and \textbf{\{C\}}, who appears in this conversation and who does not?
   & \textcolor{green}{\textbf{\{A\}}}, \textcolor{green}{\textbf{\{B\}}}, and \textcolor{green}{\textbf{\{C\}}} are all present. \\

2  & Which voices can you hear among \textbf{\{A\}}, \textbf{\{B\}}, and \textbf{\{C\}} in this dialogue?
   & You can hear all three voices: \textcolor{green}{\textbf{\{A\}}}, \textcolor{green}{\textbf{\{B\}}}, and \textcolor{green}{\textbf{\{C\}}}. \\

3  & Are \textbf{\{A\}}, \textbf{\{B\}}, and \textbf{\{C\}} talking to each other?
   & Yes—\textcolor{green}{\textbf{\{A\}}}, \textcolor{green}{\textbf{\{B\}}}, and \textcolor{green}{\textbf{\{C\}}} are having the conversation. \\

4  & I hear \textbf{\{A\}}, \textbf{\{B\}}, and \textbf{\{C\}} in the conversation, right?
   & Correct. \textcolor{green}{\textbf{\{A\}}}, \textcolor{green}{\textbf{\{B\}}}, and \textcolor{green}{\textbf{\{C\}}} all spoke. \\

5  & Did \textbf{\{A\}}, \textbf{\{B\}}, and \textbf{\{C\}} all participate in this conversation?
   & Exactly—\textcolor{green}{\textbf{\{A\}}}, \textcolor{green}{\textbf{\{B\}}}, and \textcolor{green}{\textbf{\{C\}}} all spoke in this clip. \\

6  & Can you identify \textbf{\{A\}}, \textbf{\{B\}}, and \textbf{\{C\}}’s voices in this conversation?
   & Yes, you can hear \textcolor{green}{\textbf{\{A\}}}, \textcolor{green}{\textbf{\{B\}}}, and \textcolor{green}{\textbf{\{C\}}}. \\

7  & Does this conversation include \textbf{\{A\}}, \textbf{\{B\}}, and \textbf{\{C\}}?
   & Yes. The voices of \textcolor{green}{\textbf{\{A\}}}, \textcolor{green}{\textbf{\{B\}}}, and \textcolor{green}{\textbf{\{C\}}} are all present. \\

8  & Are \textbf{\{A\}}, \textbf{\{B\}}, and \textbf{\{C\}} having a discussion?
   & Yes, I can hear all three: \textcolor{green}{\textbf{\{A\}}}, \textcolor{green}{\textbf{\{B\}}}, and \textcolor{green}{\textbf{\{C\}}}. \\

9  & Did \textbf{\{A\}} and \textbf{\{B\}} appear in this conversation?
   & Yes—both \textcolor{green}{\textbf{\{A\}}} and \textcolor{green}{\textbf{\{B\}}} are present. \\

10 & Does this conversation include the voices of \textbf{\{A\}} and \textbf{\{B\}}?
   & Yes, it includes both \textcolor{green}{\textbf{\{A\}}} and \textcolor{green}{\textbf{\{B\}}}. \\

11 & Can you hear both \textbf{\{C\}} and \textbf{\{B\}} in this conversation?
   & Yes—both \textcolor{green}{\textbf{\{C\}}} and \textcolor{green}{\textbf{\{B\}}} are speaking in this clip. \\

12 & It sounds like both \textbf{\{C\}} and \textbf{\{B\}} spoke, right?
   & Correct. Both \textcolor{green}{\textbf{\{C\}}} and \textcolor{green}{\textbf{\{B\}}} spoke. \\

13 & Are both \textbf{\{A\}} and \textbf{\{C\}} speaking in this audio?
   & Yes, both \textcolor{green}{\textbf{\{A\}}} and \textcolor{green}{\textbf{\{C\}}} participated in the conversation. \\

14 & Please confirm: do the speakers include \textbf{\{A\}} and \textbf{\{C\}}?
   & Confirmed—the speakers include \textcolor{green}{\textbf{\{A\}}} and \textcolor{green}{\textbf{\{C\}}}. \\

15 & Is \textbf{\{A\}} speaking in this audio?
   & Yes, \textcolor{green}{\textbf{\{A\}}} is speaking—very clearly. \\

16 & Can you identify \textbf{\{A\}}’s voice?
   & Yes, \textcolor{green}{\textbf{\{A\}}} definitely spoke. \\

17 & Is \textbf{\{A\}}’s voice in the audio?
   & Yes, you can hear \textcolor{green}{\textbf{\{A\}}} speaking. \\

18 & Is \textbf{\{B\}} speaking in this audio?
   & Yes, \textcolor{green}{\textbf{\{B\}}} is speaking—very clearly. \\

19 & Can you identify \textbf{\{B\}}’s voice?
   & Yes, \textcolor{green}{\textbf{\{B\}}} definitely spoke. \\

20 & Did \textbf{\{B\}} participate in this conversation?
   & Yes, \textcolor{green}{\textbf{\{B\}}} is one of the participants. \\

21 & Is \textbf{\{C\}} speaking in this audio?
   & Yes, \textcolor{green}{\textbf{\{C\}}} is speaking—very clearly. \\

22 & Did \textbf{\{C\}} participate in this conversation?
   & Yes, \textcolor{green}{\textbf{\{C\}}} is one of the participants. \\

23 & Is \textbf{\{C\}}’s voice in the audio?
   & Yes, you can hear \textcolor{green}{\textbf{\{C\}}} speaking. \\

\addlinespace[0.6em]
\midrule
\multicolumn{3}{c}{\textbf{Case 2: Two Present (P1 and P2 present, UP1 absent)}} \\
\midrule
\textbf{Index} & \textbf{Question} & \textbf{Answer} \\
\midrule

1  & Among \textbf{\{A\}}, \textbf{\{B\}}, and \textbf{\{C\}}, who appears in this conversation and who does not?
   & \textcolor{green}{\textbf{\{P1\}}} and \textcolor{green}{\textbf{\{P2\}}} spoke in the conversation, while \textcolor{red}{\textbf{\{UP1\}}} did not. \\

2  & Which voices can you hear among \textbf{\{A\}}, \textbf{\{B\}}, and \textbf{\{C\}} in this dialogue?
   & You can hear \textcolor{green}{\textbf{\{P1\}}} and \textcolor{green}{\textbf{\{P2\}}}, but not \textcolor{red}{\textbf{\{UP1\}}}. \\

3  & Are \textbf{\{A\}}, \textbf{\{B\}}, and \textbf{\{C\}} talking to each other?
   & I can hear \textcolor{green}{\textbf{\{P1\}}} and \textcolor{green}{\textbf{\{P2\}}}, but I don’t hear \textcolor{red}{\textbf{\{UP1\}}}. \\

4  & I hear \textbf{\{A\}}, \textbf{\{B\}}, and \textbf{\{C\}} in the conversation, right?
   & No. Only \textcolor{green}{\textbf{\{P1\}}} and \textcolor{green}{\textbf{\{P2\}}} are present; \textcolor{red}{\textbf{\{UP1\}}} is not. \\

5  & Did \textbf{\{A\}}, \textbf{\{B\}}, and \textbf{\{C\}} all participate in this conversation?
   & No—this clip only contains \textcolor{green}{\textbf{\{P1\}}} and \textcolor{green}{\textbf{\{P2\}}}. \\

6  & Can you identify \textbf{\{A\}}, \textbf{\{B\}}, and \textbf{\{C\}}’s voices in this conversation?
   & I can hear \textcolor{green}{\textbf{\{P1\}}} and \textcolor{green}{\textbf{\{P2\}}}, but not \textcolor{red}{\textbf{\{UP1\}}}. \\

7  & Can you hear both \textbf{\{P1\}} and \textbf{\{P2\}} in the conversation?
   & Yes—both \textcolor{green}{\textbf{\{P1\}}} and \textcolor{green}{\textbf{\{P2\}}} are speaking in this clip. \\

8  & It sounds like both \textbf{\{P1\}} and \textbf{\{P2\}} spoke, right?
   & Correct. Both \textcolor{green}{\textbf{\{P1\}}} and \textcolor{green}{\textbf{\{P2\}}} spoke. \\

9  & Are both \textbf{\{P2\}} and \textbf{\{P1\}} speaking in this audio?
   & Yes, both \textcolor{green}{\textbf{\{P2\}}} and \textcolor{green}{\textbf{\{P1\}}} participated in the conversation. \\

10 & Please confirm: do the speakers include \textbf{\{P2\}} and \textbf{\{P1\}}?
   & Confirmed—the speakers include \textcolor{green}{\textbf{\{P2\}}} and \textcolor{green}{\textbf{\{P1\}}}. \\

11 & Can you hear both \textbf{\{P1\}} and \textbf{\{UP1\}} in the conversation?
   & You can only hear \textcolor{green}{\textbf{\{P1\}}}’s voice. \\

12 & It sounds like both \textbf{\{UP1\}} and \textbf{\{P1\}} spoke, right?
   & No—only \textcolor{green}{\textbf{\{P1\}}} spoke; \textcolor{red}{\textbf{\{UP1\}}} did not. \\

13 & Are both \textbf{\{P2\}} and \textbf{\{UP1\}} speaking in this audio?
   & No—only \textcolor{green}{\textbf{\{P2\}}} spoke. \\

14 & Please confirm: do the speakers include \textbf{\{UP1\}} and \textbf{\{P2\}}?
   & The speakers include only \textcolor{green}{\textbf{\{P2\}}}. \\

15 & I think I heard \textbf{\{P1\}}—is that right?
   & Yes, \textcolor{green}{\textbf{\{P1\}}} also participated in this conversation. \\

16 & Did \textbf{\{P1\}} speak during this conversation?
   & Yes, \textcolor{green}{\textbf{\{P1\}}} spoke in the conversation. \\

17 & Is \textbf{\{P2\}} among the speakers?
   & Yes, \textcolor{green}{\textbf{\{P2\}}} is included—the voice is very distinctive. \\

18 & Can you help confirm whether \textbf{\{P2\}} spoke?
   & Confirmed—\textcolor{green}{\textbf{\{P2\}}} did speak. \\

19 & I think I heard \textbf{\{UP1\}}—is that right?
   & No, \textcolor{red}{\textbf{\{UP1\}}}’s voice did not appear. \\

20 & Did \textbf{\{UP1\}} speak during this conversation?
   & No, \textcolor{red}{\textbf{\{UP1\}}} did not speak. \\

\bottomrule
\end{tabularx}
\caption{Multi-speaker recognition QA templates for a three-speaker candidate set \{\{A\},\{B\},\{C\}\}: Case 1 (all present) and Case 2 (two present). \textbf{\textcolor{green}{Green}} indicates present speakers and \textbf{\textcolor{red}{red}} indicates absent speakers.}
\label{tab:ms_recognition_3spk_case12}
\end{table*}

\begin{table*}[t]
\centering
\small
\setlength{\tabcolsep}{6pt}
\renewcommand{\arraystretch}{1.05}
\begin{tabularx}{\textwidth}{c >{\raggedright\arraybackslash}X >{\raggedright\arraybackslash}X}
\toprule
\multicolumn{3}{c}{\textbf{\textit{Recognition (Three Speakers: A+B+C)}}} \\
\midrule

\multicolumn{3}{c}{\textbf{Case 3: One Present (P1 present, UP1 and UP2 absent)}} \\
\midrule
\textbf{Index} & \textbf{Question} & \textbf{Answer} \\
\midrule

1  & Among \textbf{\{A\}}, \textbf{\{B\}}, and \textbf{\{C\}}, who appears in this conversation and who does not?
   & Only \textcolor{green}{\textbf{\{P1\}}} spoke in the conversation; \textcolor{red}{\textbf{\{UP1\}}} and \textcolor{red}{\textbf{\{UP2\}}} did not. \\

2  & Which voices can you hear among \textbf{\{A\}}, \textbf{\{B\}}, and \textbf{\{C\}} in this dialogue?
   & You can only hear \textcolor{green}{\textbf{\{P1\}}}; you can’t hear \textcolor{red}{\textbf{\{UP1\}}} or \textcolor{red}{\textbf{\{UP2\}}}. \\

3  & Are \textbf{\{A\}}, \textbf{\{B\}}, and \textbf{\{C\}} talking to each other?
   & It’s \textcolor{green}{\textbf{\{P1\}}} talking with two other people; \textcolor{red}{\textbf{\{UP1\}}} and \textcolor{red}{\textbf{\{UP2\}}} do not appear. \\

4  & I hear \textbf{\{A\}}, \textbf{\{B\}}, and \textbf{\{C\}} in the conversation, right?
   & No. Only \textcolor{green}{\textbf{\{P1\}}} is present; \textcolor{red}{\textbf{\{UP1\}}} and \textcolor{red}{\textbf{\{UP2\}}} are not. \\

5  & Did \textbf{\{A\}}, \textbf{\{B\}}, and \textbf{\{C\}} all participate in this conversation?
   & No—only \textcolor{green}{\textbf{\{P1\}}} participated in this clip. \\

6  & Can you identify \textbf{\{A\}}, \textbf{\{B\}}, and \textbf{\{C\}}’s voices in this conversation?
   & I can only hear \textcolor{green}{\textbf{\{P1\}}}; I don’t hear \textcolor{red}{\textbf{\{UP1\}}} or \textcolor{red}{\textbf{\{UP2\}}}. \\

7  & Does this conversation include \textbf{\{A\}}, \textbf{\{B\}}, and \textbf{\{C\}}?
   & It only includes \textcolor{green}{\textbf{\{P1\}}}. \\

8  & Are \textbf{\{A\}}, \textbf{\{B\}}, and \textbf{\{C\}} having a discussion?
   & It’s \textcolor{green}{\textbf{\{P1\}}} discussing with others; it does not include \textcolor{red}{\textbf{\{UP1\}}} or \textcolor{red}{\textbf{\{UP2\}}}. \\

9  & Can you hear both \textbf{\{P1\}} and \textbf{\{UP1\}} in the conversation?
   & You can only hear \textcolor{green}{\textbf{\{P1\}}}’s voice. \\

10 & It sounds like both \textbf{\{UP1\}} and \textbf{\{P1\}} spoke, right?
   & No—only \textcolor{green}{\textbf{\{P1\}}} spoke; \textcolor{red}{\textbf{\{UP1\}}} did not. \\

11 & Are both \textbf{\{P1\}} and \textbf{\{UP2\}} speaking in this audio?
   & No—only \textcolor{green}{\textbf{\{P1\}}} spoke in this clip. \\

12 & Please confirm: do the speakers include \textbf{\{UP2\}} and \textbf{\{P1\}}?
   & The speakers include only \textcolor{green}{\textbf{\{P1\}}}. \\

13 & I think I heard \textbf{\{P1\}}—is that right?
   & Yes, \textcolor{green}{\textbf{\{P1\}}} also participated in this conversation. \\

14 & Did \textbf{\{P1\}} speak during this conversation?
   & Yes, \textcolor{green}{\textbf{\{P1\}}} spoke in the conversation. \\

15 & I think I heard \textbf{\{UP1\}}—is that right?
   & No, \textcolor{red}{\textbf{\{UP1\}}}’s voice did not appear. \\

16 & Did \textbf{\{UP1\}} speak during this conversation?
   & No, \textcolor{red}{\textbf{\{UP1\}}} did not speak. \\

17 & Are \textbf{\{UP2\}} among the speakers?
   & No, \textcolor{red}{\textbf{\{UP2\}}} is not included. \\

18 & Can you help confirm whether \textbf{\{UP2\}} spoke?
   & \textcolor{red}{\textbf{\{UP2\}}} did not speak. \\

\addlinespace[0.6em]
\midrule
\multicolumn{3}{c}{\textbf{Case 4: None Present (A, B, and C absent)}} \\
\midrule
\textbf{Index} & \textbf{Question} & \textbf{Answer} \\
\midrule

1  & Among \textbf{\{A\}}, \textbf{\{B\}}, and \textbf{\{C\}}, who appears in this conversation and who does not?
   & None of \textcolor{red}{\textbf{\{A\}}}, \textcolor{red}{\textbf{\{B\}}}, or \textcolor{red}{\textbf{\{C\}}} is present. \\

2  & Which voices can you hear among \textbf{\{A\}}, \textbf{\{B\}}, and \textbf{\{C\}} in this dialogue?
   & You can’t hear any of \textcolor{red}{\textbf{\{A\}}}, \textcolor{red}{\textbf{\{B\}}}, or \textcolor{red}{\textbf{\{C\}}}. \\

3  & Are \textbf{\{A\}}, \textbf{\{B\}}, and \textbf{\{C\}} talking to each other?
   & No. The conversation does not contain any of \textcolor{red}{\textbf{\{A\}}}, \textcolor{red}{\textbf{\{B\}}}, or \textcolor{red}{\textbf{\{C\}}}. \\

4  & I hear \textbf{\{A\}}, \textbf{\{B\}}, and \textbf{\{C\}} in the conversation, right?
   & No—none of \textcolor{red}{\textbf{\{A\}}}, \textcolor{red}{\textbf{\{B\}}}, or \textcolor{red}{\textbf{\{C\}}} appears. \\

5  & Did \textbf{\{A\}}, \textbf{\{B\}}, and \textbf{\{C\}} all participate in this conversation?
   & No. This clip does not include \textcolor{red}{\textbf{\{A\}}}, \textcolor{red}{\textbf{\{B\}}}, or \textcolor{red}{\textbf{\{C\}}}. \\

6  & Can you identify \textbf{\{A\}}, \textbf{\{B\}}, and \textbf{\{C\}}’s voices in this conversation?
   & No—I can’t identify \textcolor{red}{\textbf{\{A\}}}, \textcolor{red}{\textbf{\{B\}}}, or \textcolor{red}{\textbf{\{C\}}} here. \\

7  & Does this conversation include \textbf{\{A\}}, \textbf{\{B\}}, and \textbf{\{C\}}?
   & No. It includes none of \textcolor{red}{\textbf{\{A\}}}, \textcolor{red}{\textbf{\{B\}}}, or \textcolor{red}{\textbf{\{C\}}}. \\

8  & Did \textbf{\{A\}} and \textbf{\{B\}} both appear in this conversation?
   & Neither \textcolor{red}{\textbf{\{A\}}} nor \textcolor{red}{\textbf{\{B\}}} is present. \\

9  & Does this conversation include the voices of \textbf{\{A\}} and \textbf{\{B\}}?
   & No—neither \textcolor{red}{\textbf{\{A\}}} nor \textcolor{red}{\textbf{\{B\}}} is detected. \\

10 & Can you hear both \textbf{\{C\}} and \textbf{\{B\}} in this conversation?
   & No. You can’t hear \textcolor{red}{\textbf{\{C\}}} or \textcolor{red}{\textbf{\{B\}}} in this clip. \\

11 & It sounds like both \textbf{\{C\}} and \textbf{\{B\}} spoke, right?
   & No—neither \textcolor{red}{\textbf{\{C\}}} nor \textcolor{red}{\textbf{\{B\}}} spoke. \\

12 & Are both \textbf{\{A\}} and \textbf{\{C\}} speaking in this audio?
   & No—it is not \textcolor{red}{\textbf{\{A\}}} or \textcolor{red}{\textbf{\{C\}}} speaking. \\

13 & Please confirm: are the speakers \textbf{\{A\}} and \textbf{\{C\}}?
   & I don’t think it’s \textcolor{red}{\textbf{\{A\}}} or \textcolor{red}{\textbf{\{C\}}}. \\

14 & Is \textbf{\{A\}} speaking in this audio?
   & No, \textcolor{red}{\textbf{\{A\}}}’s voice is not present. \\

15 & Can you identify \textbf{\{A\}}’s voice?
   & I can’t identify \textcolor{red}{\textbf{\{A\}}}’s voice here. \\

16 & Did \textbf{\{A\}} participate in this conversation?
   & No, \textcolor{red}{\textbf{\{A\}}} did not participate. \\

17 & Is \textbf{\{B\}} speaking in this audio?
   & No, \textcolor{red}{\textbf{\{B\}}}’s voice is not present. \\

18 & Can you identify \textbf{\{B\}}’s voice?
   & I can’t identify \textcolor{red}{\textbf{\{B\}}}’s voice here. \\

19 & Is \textbf{\{B\}}’s voice in the audio?
   & No, \textcolor{red}{\textbf{\{B\}}}’s voice is not present. \\

20 & Is \textbf{\{C\}} speaking in this audio?
   & No, \textcolor{red}{\textbf{\{C\}}}’s voice is not present. \\

21 & Did \textbf{\{C\}} participate in this conversation?
   & No, \textcolor{red}{\textbf{\{C\}}} did not participate. \\

22 & Is \textbf{\{C\}}’s voice in the audio?
   & No, \textcolor{red}{\textbf{\{C\}}}’s voice is not present. \\

\bottomrule
\end{tabularx}
\caption{Multi-speaker recognition QA templates for a three-speaker candidate set \{\{A\},\{B\},\{C\}\}: Case 3 (one present) and Case 4 (none present). \textbf{\textcolor{green}{Green}} indicates present speakers and \textbf{\textcolor{red}{red}} indicates absent speakers.}
\label{tab:ms_recognition_3spk_case34}
\end{table*}

\begin{table*}[t]
\centering
\small
\setlength{\tabcolsep}{6pt}
\renewcommand{\arraystretch}{1.05}
\begin{tabularx}{\textwidth}{c >{\raggedright\arraybackslash}X >{\raggedright\arraybackslash}X}
\toprule
\multicolumn{3}{c}{\textbf{\textit{Recognition (Four Speakers: A+B+C+D)}}} \\
\midrule

\multicolumn{3}{c}{\textbf{Case 1: All Present (A, B, C, and D)}} \\
\midrule
\textbf{Index} & \textbf{Question} & \textbf{Answer} \\
\midrule

1  & Among \textbf{\{A\}}, \textbf{\{B\}}, \textbf{\{C\}}, and \textbf{\{D\}}, who appears in this conversation and who does not?
   & \textcolor{green}{\textbf{\{A\}}}, \textcolor{green}{\textbf{\{B\}}}, \textcolor{green}{\textbf{\{C\}}}, and \textcolor{green}{\textbf{\{D\}}} are all present. \\

2  & Which voices can you hear among \textbf{\{A\}}, \textbf{\{B\}}, \textbf{\{C\}}, and \textbf{\{D\}} in this dialogue?
   & You can hear all four voices: \textcolor{green}{\textbf{\{A\}}}, \textcolor{green}{\textbf{\{B\}}}, \textcolor{green}{\textbf{\{C\}}}, and \textcolor{green}{\textbf{\{D\}}}. \\

3  & Are \textbf{\{A\}}, \textbf{\{B\}}, \textbf{\{C\}}, and \textbf{\{D\}} talking to each other?
   & Yes—\textcolor{green}{\textbf{\{A\}}}, \textcolor{green}{\textbf{\{B\}}}, \textcolor{green}{\textbf{\{C\}}}, and \textcolor{green}{\textbf{\{D\}}} are having the conversation. \\

4  & I hear \textbf{\{A\}}, \textbf{\{B\}}, \textbf{\{C\}}, and \textbf{\{D\}} in the conversation, right?
   & Correct. \textcolor{green}{\textbf{\{A\}}}, \textcolor{green}{\textbf{\{B\}}}, \textcolor{green}{\textbf{\{C\}}}, and \textcolor{green}{\textbf{\{D\}}} all spoke. \\

5  & Are \textbf{\{A\}}, \textbf{\{B\}}, \textbf{\{C\}}, and \textbf{\{D\}} all present?
   & Exactly—\textcolor{green}{\textbf{\{A\}}}, \textcolor{green}{\textbf{\{B\}}}, \textcolor{green}{\textbf{\{C\}}}, and \textcolor{green}{\textbf{\{D\}}} all spoke in this clip. \\

6  & Can you identify \textbf{\{A\}}, \textbf{\{B\}}, \textbf{\{C\}}, and \textbf{\{D\}}’s voices in this conversation?
   & Yes, you can hear \textcolor{green}{\textbf{\{A\}}}, \textcolor{green}{\textbf{\{B\}}}, \textcolor{green}{\textbf{\{C\}}}, and \textcolor{green}{\textbf{\{D\}}}. \\

7  & Does this conversation include \textbf{\{A\}}, \textbf{\{B\}}, \textbf{\{C\}}, and \textbf{\{D\}}?
   & Yes—all four are present. \\

8  & Are \textbf{\{A\}}, \textbf{\{B\}}, \textbf{\{C\}}, and \textbf{\{D\}} having a discussion?
   & Yes—I can hear all four: \textcolor{green}{\textbf{\{A\}}}, \textcolor{green}{\textbf{\{B\}}}, \textcolor{green}{\textbf{\{C\}}}, and \textcolor{green}{\textbf{\{D\}}}. \\

9  & Among \textbf{\{A\}}, \textbf{\{B\}}, and \textbf{\{C\}}, who appears in this conversation and who does not?
   & \textcolor{green}{\textbf{\{A\}}}, \textcolor{green}{\textbf{\{B\}}}, and \textcolor{green}{\textbf{\{C\}}} are all present. \\

10 & Are \textbf{\{A\}}, \textbf{\{D\}}, and \textbf{\{C\}} talking to each other?
   & Yes—\textcolor{green}{\textbf{\{A\}}}, \textcolor{green}{\textbf{\{D\}}}, and \textcolor{green}{\textbf{\{C\}}} are talking. \\

11 & Are \textbf{\{A\}}, \textbf{\{B\}}, and \textbf{\{D\}} all present?
   & Yes—\textcolor{green}{\textbf{\{A\}}}, \textcolor{green}{\textbf{\{B\}}}, and \textcolor{green}{\textbf{\{D\}}} all spoke. \\

12 & Does this conversation include \textbf{\{D\}}, \textbf{\{B\}}, and \textbf{\{C\}}?
   & Yes. You can hear \textcolor{green}{\textbf{\{D\}}}, \textcolor{green}{\textbf{\{B\}}}, and \textcolor{green}{\textbf{\{C\}}}. \\

13 & Did \textbf{\{A\}} and \textbf{\{B\}} appear in this conversation?
   & Yes—both \textcolor{green}{\textbf{\{A\}}} and \textcolor{green}{\textbf{\{B\}}} are present. \\

14 & Are both \textbf{\{A\}} and \textbf{\{D\}} present?
   & Yes, it includes both \textcolor{green}{\textbf{\{A\}}} and \textcolor{green}{\textbf{\{D\}}}. \\

15 & Can you hear both \textbf{\{C\}} and \textbf{\{B\}} in this conversation?
   & Yes—both \textcolor{green}{\textbf{\{C\}}} and \textcolor{green}{\textbf{\{B\}}} are speaking in this clip. \\

16 & It sounds like both \textbf{\{D\}} and \textbf{\{B\}} spoke, right?
   & Correct. Both \textcolor{green}{\textbf{\{D\}}} and \textcolor{green}{\textbf{\{B\}}} spoke. \\

17 & Are both \textbf{\{A\}} and \textbf{\{C\}} speaking in this audio?
   & Yes, both \textcolor{green}{\textbf{\{A\}}} and \textcolor{green}{\textbf{\{C\}}} participated in the conversation. \\

18 & Is \textbf{\{A\}} speaking in this audio?
   & Yes, \textcolor{green}{\textbf{\{A\}}} is speaking—very clearly. \\

19 & Can you identify \textbf{\{A\}}’s voice?
   & Yes, \textcolor{green}{\textbf{\{A\}}} definitely spoke. \\

20 & Can you identify \textbf{\{B\}}’s voice?
   & Yes, \textcolor{green}{\textbf{\{B\}}} definitely spoke. \\

21 & Did \textbf{\{B\}} participate in this conversation?
   & Yes, \textcolor{green}{\textbf{\{B\}}} is one of the participants. \\

22 & Is \textbf{\{C\}} speaking in this audio?
   & Yes, \textcolor{green}{\textbf{\{C\}}} is speaking—very clearly. \\

23 & Is \textbf{\{C\}}’s voice in the audio?
   & Yes, you can hear \textcolor{green}{\textbf{\{C\}}} speaking. \\

24 & I think I heard \textbf{\{D\}}—is that right?
   & Yes, \textcolor{green}{\textbf{\{D\}}}’s voice is present. \\

25 & Did \textbf{\{D\}} speak during this conversation?
   & Yes, \textcolor{green}{\textbf{\{D\}}} spoke in the conversation. \\

\addlinespace[0.6em]
\midrule
\multicolumn{3}{c}{\textbf{Case 2: Three Present (P1, P2, P3 present; UP1 absent)}} \\
\midrule
\textbf{Index} & \textbf{Question} & \textbf{Answer} \\
\midrule

1  & Among \textbf{\{A\}}, \textbf{\{B\}}, \textbf{\{C\}}, and \textbf{\{D\}}, who appears in this conversation and who does not?
   & \textcolor{green}{\textbf{\{P1\}}}, \textcolor{green}{\textbf{\{P2\}}}, and \textcolor{green}{\textbf{\{P3\}}} are present, while \textcolor{red}{\textbf{\{UP1\}}} is not. \\

2  & Which voices can you hear among \textbf{\{A\}}, \textbf{\{B\}}, \textbf{\{C\}}, and \textbf{\{D\}} in this dialogue?
   & You can hear \textcolor{green}{\textbf{\{P1\}}}, \textcolor{green}{\textbf{\{P2\}}}, and \textcolor{green}{\textbf{\{P3\}}}. \\

3  & Are \textbf{\{A\}}, \textbf{\{B\}}, \textbf{\{C\}}, and \textbf{\{D\}} talking to each other?
   & The conversation includes \textcolor{green}{\textbf{\{P1\}}}, \textcolor{green}{\textbf{\{P2\}}}, and \textcolor{green}{\textbf{\{P3\}}}, but not \textcolor{red}{\textbf{\{UP1\}}}. \\

4  & I hear \textbf{\{A\}}, \textbf{\{B\}}, \textbf{\{C\}}, and \textbf{\{D\}} in the conversation, right?
   & No. Only \textcolor{green}{\textbf{\{P1\}}}, \textcolor{green}{\textbf{\{P2\}}}, and \textcolor{green}{\textbf{\{P3\}}} are present. \\

5  & Did \textbf{\{A\}}, \textbf{\{B\}}, \textbf{\{C\}}, and \textbf{\{D\}} all participate in this conversation?
   & No—only \textcolor{green}{\textbf{\{P1\}}}, \textcolor{green}{\textbf{\{P2\}}}, and \textcolor{green}{\textbf{\{P3\}}} participated. \\

6  & Among \textbf{\{P1\}}, \textbf{\{P2\}}, and \textbf{\{P3\}}, who appears in this conversation and who does not?
   & \textcolor{green}{\textbf{\{P1\}}}, \textcolor{green}{\textbf{\{P2\}}}, and \textcolor{green}{\textbf{\{P3\}}} are all present. \\

7  & Are \textbf{\{P1\}}, \textbf{\{P2\}}, and \textbf{\{P3\}} talking to each other?
   & Yes—\textcolor{green}{\textbf{\{P1\}}}, \textcolor{green}{\textbf{\{P2\}}}, and \textcolor{green}{\textbf{\{P3\}}} are having the conversation. \\

8  & Did \textbf{\{P1\}}, \textbf{\{P2\}}, and \textbf{\{UP1\}} all participate in this conversation?
   & No—only \textcolor{green}{\textbf{\{P1\}}} and \textcolor{green}{\textbf{\{P2\}}} are present; \textcolor{red}{\textbf{\{UP1\}}} is not. \\

9  & Does this conversation include \textbf{\{P1\}}, \textbf{\{P3\}}, and \textbf{\{UP1\}}?
   & It includes \textcolor{green}{\textbf{\{P1\}}} and \textcolor{green}{\textbf{\{P2\}}}, but not \textcolor{red}{\textbf{\{UP1\}}}. \\

10 & Which voices do you hear among \textbf{\{UP1\}}, \textbf{\{P2\}}, and \textbf{\{P3\}}?
   & You can hear \textcolor{green}{\textbf{\{P2\}}} and \textcolor{green}{\textbf{\{P3\}}}, but not \textcolor{red}{\textbf{\{UP1\}}}. \\

11 & Did \textbf{\{P2\}} and \textbf{\{P3\}} appear in this conversation?
   & Yes—both \textcolor{green}{\textbf{\{P2\}}} and \textcolor{green}{\textbf{\{P3\}}} are present. \\

12 & Does this conversation include \textbf{\{P2\}} and \textbf{\{P1\}}?
   & Yes, it includes both \textcolor{green}{\textbf{\{P2\}}} and \textcolor{green}{\textbf{\{P1\}}}. \\

13 & Can you hear both \textbf{\{P1\}} and \textbf{\{P3\}} in this conversation?
   & Yes—both \textcolor{green}{\textbf{\{P1\}}} and \textcolor{green}{\textbf{\{P3\}}} are speaking in this clip. \\

14 & It sounds like both \textbf{\{P1\}} and \textbf{\{UP1\}} spoke, right?
   & No—only \textcolor{green}{\textbf{\{P1\}}} is present; \textcolor{red}{\textbf{\{UP1\}}} is not. \\

15 & Are both \textbf{\{P2\}} and \textbf{\{UP1\}} speaking in this audio?
   & No—only \textcolor{green}{\textbf{\{P2\}}} spoke; \textcolor{red}{\textbf{\{UP1\}}} did not. \\

16 & Do the speakers include \textbf{\{UP1\}} and \textbf{\{P3\}}?
   & The speakers include \textcolor{green}{\textbf{\{P3\}}} but not \textcolor{red}{\textbf{\{UP1\}}}. \\

17 & Did \textbf{\{P1\}} participate in this conversation?
   & Yes, \textcolor{green}{\textbf{\{P1\}}} is one of the participants. \\

18 & Is \textbf{\{P2\}}’s voice in the audio?
   & Yes, you can hear \textcolor{green}{\textbf{\{P2\}}} speaking. \\

19 & I think I heard \textbf{\{P3\}}—is that right?
   & Yes, \textcolor{green}{\textbf{\{P3\}}} also participated in this conversation. \\

20 & Did \textbf{\{UP1\}} speak during this conversation?
   & No, \textcolor{red}{\textbf{\{UP1\}}} did not speak. \\

\bottomrule
\end{tabularx}
\caption{Multi-speaker recognition QA templates for a four-speaker candidate set \{\{A\},\{B\},\{C\},\{D\}\}: Case 1 (all present) and Case 2 (three present). \textbf{\textcolor{green}{Green}} indicates present speakers and \textbf{\textcolor{red}{red}} indicates absent speakers.}
\label{tab:ms_recognition_4spk_case12}
\end{table*}

\begin{table*}[t]
\centering
\small
\setlength{\tabcolsep}{6pt}
\renewcommand{\arraystretch}{1.05}
\begin{tabularx}{\textwidth}{c >{\raggedright\arraybackslash}X >{\raggedright\arraybackslash}X}
\toprule
\multicolumn{3}{c}{\textbf{\textit{Recognition (Four Speakers: A+B+C+D)}}} \\
\midrule

\multicolumn{3}{c}{\textbf{Case 3: Two Present (P1 and P2 present; UP1 and UP2 absent)}} \\
\midrule
\textbf{Index} & \textbf{Question} & \textbf{Answer} \\
\midrule

1  & Among \textbf{\{A\}}, \textbf{\{B\}}, \textbf{\{C\}}, and \textbf{\{D\}}, who appears in this conversation and who does not?
   & \textcolor{green}{\textbf{\{P1\}}} and \textcolor{green}{\textbf{\{P2\}}} are present, while \textcolor{red}{\textbf{\{UP1\}}} and \textcolor{red}{\textbf{\{UP2\}}} are not. \\

2  & Which voices can you hear among \textbf{\{A\}}, \textbf{\{B\}}, \textbf{\{C\}}, and \textbf{\{D\}} in this dialogue?
   & You can hear \textcolor{green}{\textbf{\{P1\}}} and \textcolor{green}{\textbf{\{P2\}}}. \\

3  & Are \textbf{\{A\}}, \textbf{\{B\}}, \textbf{\{C\}}, and \textbf{\{D\}} talking to each other?
   & The conversation includes \textcolor{green}{\textbf{\{P1\}}} and \textcolor{green}{\textbf{\{P2\}}}, but not \textcolor{red}{\textbf{\{UP1\}}} or \textcolor{red}{\textbf{\{UP2\}}}. \\

4  & I hear \textbf{\{A\}}, \textbf{\{B\}}, \textbf{\{C\}}, and \textbf{\{D\}} in the conversation, right?
   & No. Only \textcolor{green}{\textbf{\{P1\}}} and \textcolor{green}{\textbf{\{P2\}}} are present. \\

5  & Did \textbf{\{A\}}, \textbf{\{B\}}, \textbf{\{C\}}, and \textbf{\{D\}} all participate in this conversation?
   & No—only \textcolor{green}{\textbf{\{P1\}}} and \textcolor{green}{\textbf{\{P2\}}} participated. \\

6  & Among \textbf{\{P1\}}, \textbf{\{P2\}}, and \textbf{\{UP1\}}, who appears in this conversation and who does not?
   & \textcolor{green}{\textbf{\{P1\}}} and \textcolor{green}{\textbf{\{P2\}}} are present; \textcolor{red}{\textbf{\{UP1\}}} is not. \\

7  & Are \textbf{\{UP1\}}, \textbf{\{P1\}}, and \textbf{\{P2\}} talking to each other?
   & No—\textcolor{green}{\textbf{\{P1\}}} and \textcolor{green}{\textbf{\{P2\}}} are speaking, but not \textcolor{red}{\textbf{\{UP1\}}}. \\

8  & Did \textbf{\{P1\}}, \textbf{\{P2\}}, and \textbf{\{UP2\}} all participate in this conversation?
   & No—only \textcolor{green}{\textbf{\{P1\}}} and \textcolor{green}{\textbf{\{P2\}}} are present; \textcolor{red}{\textbf{\{UP2\}}} is not. \\

9  & Does this conversation include \textbf{\{P1\}}, \textbf{\{UP2\}}, and \textbf{\{P2\}}?
   & It includes \textcolor{green}{\textbf{\{P1\}}} and \textcolor{green}{\textbf{\{P2\}}}, but not \textcolor{red}{\textbf{\{UP2\}}}. \\

10 & Did \textbf{\{P1\}} and \textbf{\{P2\}} appear in this conversation?
   & Yes—both \textcolor{green}{\textbf{\{P1\}}} and \textcolor{green}{\textbf{\{P2\}}} are present. \\

11 & Does this conversation include the voices of \textbf{\{UP1\}} and \textbf{\{UP2\}}?
   & No—it includes neither \textcolor{red}{\textbf{\{UP1\}}} nor \textcolor{red}{\textbf{\{UP2\}}}. \\

12 & Can you hear both \textbf{\{UP2\}} and \textbf{\{P2\}} in this conversation?
   & You can only hear \textcolor{green}{\textbf{\{P2\}}}; \textcolor{red}{\textbf{\{UP2\}}} is not present. \\

13 & It sounds like both \textbf{\{P1\}} and \textbf{\{UP1\}} spoke, right?
   & No—only \textcolor{green}{\textbf{\{P1\}}} is present; \textcolor{red}{\textbf{\{UP1\}}} is not. \\

14 & Are both \textbf{\{P2\}} and \textbf{\{UP1\}} speaking in this audio?
   & No—only \textcolor{green}{\textbf{\{P2\}}} spoke; \textcolor{red}{\textbf{\{UP1\}}} did not. \\

15 & Please confirm: do the speakers include \textbf{\{UP2\}} and \textbf{\{P1\}}?
   & The speakers include \textcolor{green}{\textbf{\{P1\}}} but not \textcolor{red}{\textbf{\{UP2\}}}. \\

16 & Did \textbf{\{P1\}} participate in this conversation?
   & Yes, \textcolor{green}{\textbf{\{P1\}}} is one of the participants. \\

17 & Is \textbf{\{P2\}}’s voice in the audio?
   & Yes, you can hear \textcolor{green}{\textbf{\{P2\}}} speaking. \\

18 & I think I heard \textbf{\{UP2\}}—is that right?
   & No, \textcolor{red}{\textbf{\{UP2\}}}’s voice did not appear. \\

19 & Did \textbf{\{UP1\}} speak during this conversation?
   & No, \textcolor{red}{\textbf{\{UP1\}}} did not speak. \\

\addlinespace[0.6em]
\midrule
\multicolumn{3}{c}{\textbf{Case 4: One Present (P1 present; UP1, UP2, UP3 absent)}} \\
\midrule
\textbf{Index} & \textbf{Question} & \textbf{Answer} \\
\midrule

1  & Among \textbf{\{A\}}, \textbf{\{B\}}, \textbf{\{C\}}, and \textbf{\{D\}}, who appears in this conversation and who does not?
   & Only \textcolor{green}{\textbf{\{P1\}}} is present; \textcolor{red}{\textbf{\{UP1\}}}, \textcolor{red}{\textbf{\{UP2\}}}, and \textcolor{red}{\textbf{\{UP3\}}} are not. \\

2  & Which voices can you hear among \textbf{\{A\}}, \textbf{\{B\}}, \textbf{\{C\}}, and \textbf{\{D\}} in this dialogue?
   & You can only hear \textcolor{green}{\textbf{\{P1\}}}. \\

3  & Are \textbf{\{A\}}, \textbf{\{B\}}, \textbf{\{C\}}, and \textbf{\{D\}} talking to each other?
   & It’s \textcolor{green}{\textbf{\{P1\}}} speaking with others; \textcolor{red}{\textbf{\{UP1\}}}, \textcolor{red}{\textbf{\{UP2\}}}, and \textcolor{red}{\textbf{\{UP3\}}} do not appear. \\

4  & I hear \textbf{\{A\}}, \textbf{\{B\}}, \textbf{\{C\}}, and \textbf{\{D\}} in the conversation, right?
   & No. Only \textcolor{green}{\textbf{\{P1\}}} is present. \\

5  & Did \textbf{\{A\}}, \textbf{\{B\}}, \textbf{\{C\}}, and \textbf{\{D\}} all participate in this conversation?
   & No—only \textcolor{green}{\textbf{\{P1\}}} participated. \\

6  & Among \textbf{\{P1\}}, \textbf{\{UP1\}}, and \textbf{\{UP2\}}, who appears in this conversation and who does not?
   & \textcolor{green}{\textbf{\{P1\}}} is present; \textcolor{red}{\textbf{\{UP1\}}} and \textcolor{red}{\textbf{\{UP2\}}} are not. \\

7  & Are \textbf{\{UP1\}}, \textbf{\{P1\}}, and \textbf{\{UP3\}} talking to each other?
   & No—only \textcolor{green}{\textbf{\{P1\}}} is speaking; \textcolor{red}{\textbf{\{UP1\}}} and \textcolor{red}{\textbf{\{UP3\}}} are not present. \\

8  & Did \textbf{\{P1\}}, \textbf{\{UP2\}}, and \textbf{\{UP3\}} all participate in this conversation?
   & No—this clip only contains \textcolor{green}{\textbf{\{P1\}}}. \\

9  & Does this conversation include \textbf{\{UP1\}}, \textbf{\{UP2\}}, and \textbf{\{UP3\}}?
   & No—it includes none of \textcolor{red}{\textbf{\{UP1\}}}, \textcolor{red}{\textbf{\{UP2\}}}, or \textcolor{red}{\textbf{\{UP3\}}}. \\

10 & Did \textbf{\{P1\}} and \textbf{\{UP1\}} appear in this conversation?
   & Only \textcolor{green}{\textbf{\{P1\}}} is present; \textcolor{red}{\textbf{\{UP1\}}} is not. \\

11 & Does this conversation include the voices of \textbf{\{UP2\}} and \textbf{\{P1\}}?
   & You can only hear \textcolor{green}{\textbf{\{P1\}}}; \textcolor{red}{\textbf{\{UP2\}}} is not present. \\

12 & Can you hear both \textbf{\{UP3\}} and \textbf{\{P1\}} in this conversation?
   & You can only hear \textcolor{green}{\textbf{\{P1\}}}; \textcolor{red}{\textbf{\{UP3\}}} is not present. \\

13 & It sounds like both \textbf{\{UP1\}} and \textbf{\{UP2\}} spoke, right?
   & No—neither \textcolor{red}{\textbf{\{UP1\}}} nor \textcolor{red}{\textbf{\{UP2\}}} spoke. \\

14 & Are both \textbf{\{UP1\}} and \textbf{\{UP3\}} speaking in this audio?
   & No—neither \textcolor{red}{\textbf{\{UP1\}}} nor \textcolor{red}{\textbf{\{UP3\}}} spoke. \\

15 & Please confirm: do the speakers include \textbf{\{UP2\}} and \textbf{\{UP3\}}?
   & No—neither \textcolor{red}{\textbf{\{UP2\}}} nor \textcolor{red}{\textbf{\{UP3\}}} is present. \\

16 & Did \textbf{\{P1\}} participate in this conversation?
   & Yes, \textcolor{green}{\textbf{\{P1\}}} is one of the participants. \\

17 & Did \textbf{\{UP1\}} speak during this conversation?
   & No, \textcolor{red}{\textbf{\{UP1\}}} did not speak. \\

18 & Did \textbf{\{UP2\}} participate in this conversation?
   & No, \textcolor{red}{\textbf{\{UP2\}}} did not participate. \\

19 & Is \textbf{\{UP3\}}’s voice in the audio?
   & No, \textcolor{red}{\textbf{\{UP3\}}}’s voice is not present. \\

\bottomrule
\end{tabularx}
\caption{Multi-speaker recognition QA templates for a four-speaker candidate set \{\{A\},\{B\},\{C\},\{D\}\}: Case 3 (two present) and Case 4 (one present). \textbf{\textcolor{green}{Green}} indicates present speakers and \textbf{\textcolor{red}{red}} indicates absent speakers.}
\label{tab:ms_recognition_4spk_case34}
\end{table*}

\begin{table*}[t]
\centering
\small
\setlength{\tabcolsep}{6pt}
\renewcommand{\arraystretch}{1.05}
\begin{tabularx}{\textwidth}{c >{\raggedright\arraybackslash}X >{\raggedright\arraybackslash}X}
\toprule
\multicolumn{3}{c}{\textbf{\textit{Recognition (Four Speakers: A+B+C+D)}}} \\
\midrule

\multicolumn{3}{c}{\textbf{Case 5: None Present (A, B, C, and D absent)}} \\
\midrule
\textbf{Index} & \textbf{Question} & \textbf{Answer} \\
\midrule

1  & Among \textbf{\{A\}}, \textbf{\{B\}}, \textbf{\{C\}}, and \textbf{\{D\}}, who appears in this conversation and who does not?
   & None of \textcolor{red}{\textbf{\{A\}}}, \textcolor{red}{\textbf{\{B\}}}, \textcolor{red}{\textbf{\{C\}}}, or \textcolor{red}{\textbf{\{D\}}} is present. \\

2  & Which voices can you hear among \textbf{\{A\}}, \textbf{\{B\}}, \textbf{\{C\}}, and \textbf{\{D\}} in this dialogue?
   & You can’t hear any of \textcolor{red}{\textbf{\{A\}}}, \textcolor{red}{\textbf{\{B\}}}, \textcolor{red}{\textbf{\{C\}}}, or \textcolor{red}{\textbf{\{D\}}}. \\

3  & Are \textbf{\{A\}}, \textbf{\{B\}}, \textbf{\{C\}}, and \textbf{\{D\}} talking to each other?
   & No—none of \textcolor{red}{\textbf{\{A\}}}, \textcolor{red}{\textbf{\{B\}}}, \textcolor{red}{\textbf{\{C\}}}, or \textcolor{red}{\textbf{\{D\}}} appears in this conversation. \\

4  & I hear \textbf{\{A\}}, \textbf{\{B\}}, \textbf{\{C\}}, and \textbf{\{D\}} in the conversation, right?
   & No—there is no \textcolor{red}{\textbf{\{A\}}}, \textcolor{red}{\textbf{\{B\}}}, \textcolor{red}{\textbf{\{C\}}}, or \textcolor{red}{\textbf{\{D\}}} voice present. \\

5  & Did \textbf{\{A\}}, \textbf{\{B\}}, \textbf{\{C\}}, and \textbf{\{D\}} all participate in this conversation?
   & No—this clip includes none of \textcolor{red}{\textbf{\{A\}}}, \textcolor{red}{\textbf{\{B\}}}, \textcolor{red}{\textbf{\{C\}}}, or \textcolor{red}{\textbf{\{D\}}}. \\

6  & Can you identify \textbf{\{A\}}, \textbf{\{B\}}, \textbf{\{C\}}, and \textbf{\{D\}}’s voices in this conversation?
   & No—I can’t identify any of \textcolor{red}{\textbf{\{A\}}}, \textcolor{red}{\textbf{\{B\}}}, \textcolor{red}{\textbf{\{C\}}}, or \textcolor{red}{\textbf{\{D\}}} here. \\

7  & Among \textbf{\{A\}}, \textbf{\{B\}}, and \textbf{\{C\}}, who appears in this conversation and who does not?
   & None of \textcolor{red}{\textbf{\{A\}}}, \textcolor{red}{\textbf{\{B\}}}, or \textcolor{red}{\textbf{\{C\}}} is present. \\

8  & Are \textbf{\{A\}}, \textbf{\{D\}}, and \textbf{\{C\}} talking to each other?
   & No—it is not \textcolor{red}{\textbf{\{A\}}}, \textcolor{red}{\textbf{\{D\}}}, and \textcolor{red}{\textbf{\{C\}}} talking. \\

9  & Did \textbf{\{A\}}, \textbf{\{B\}}, and \textbf{\{D\}} all participate in this conversation?
   & No—\textcolor{red}{\textbf{\{A\}}}, \textcolor{red}{\textbf{\{B\}}}, and \textcolor{red}{\textbf{\{D\}}} did not participate. \\

10 & Does this conversation include \textbf{\{D\}}, \textbf{\{B\}}, and \textbf{\{C\}}?
   & No—it includes none of \textcolor{red}{\textbf{\{D\}}}, \textcolor{red}{\textbf{\{B\}}}, or \textcolor{red}{\textbf{\{C\}}}. \\

11 & Did \textbf{\{A\}} and \textbf{\{B\}} appear in this conversation?
   & Neither \textcolor{red}{\textbf{\{A\}}} nor \textcolor{red}{\textbf{\{B\}}} is present. \\

12 & Does this conversation include the voices of \textbf{\{A\}} and \textbf{\{D\}}?
   & No—it includes neither \textcolor{red}{\textbf{\{A\}}} nor \textcolor{red}{\textbf{\{D\}}}. \\

13 & Can you hear both \textbf{\{C\}} and \textbf{\{B\}} in this conversation?
   & No—you can’t hear \textcolor{red}{\textbf{\{C\}}} or \textcolor{red}{\textbf{\{B\}}}. \\

14 & It sounds like both \textbf{\{D\}} and \textbf{\{B\}} spoke, right?
   & No—neither \textcolor{red}{\textbf{\{D\}}} nor \textcolor{red}{\textbf{\{B\}}} spoke. \\

15 & Are both \textbf{\{A\}} and \textbf{\{C\}} speaking in this audio?
   & No—neither \textcolor{red}{\textbf{\{A\}}} nor \textcolor{red}{\textbf{\{C\}}} participated in this conversation. \\

16 & Is \textbf{\{A\}} speaking in this audio?
   & No, \textcolor{red}{\textbf{\{A\}}}’s voice is not present. \\

17 & Can you identify \textbf{\{A\}}’s voice?
   & No—I can’t identify \textcolor{red}{\textbf{\{A\}}}’s voice here. \\

18 & Can you identify \textbf{\{B\}}’s voice?
   & No—I can’t identify \textcolor{red}{\textbf{\{B\}}}’s voice here. \\

19 & Did \textbf{\{B\}} participate in this conversation?
   & No, \textcolor{red}{\textbf{\{B\}}} did not participate. \\

20 & Is \textbf{\{C\}} speaking in this audio?
   & No, \textcolor{red}{\textbf{\{C\}}}’s voice is not present. \\

21 & Is \textbf{\{C\}}’s voice in the audio?
   & No—I can’t make out \textcolor{red}{\textbf{\{C\}}}’s voice. \\

22 & I think I heard \textbf{\{D\}}—is that right?
   & No, \textcolor{red}{\textbf{\{D\}}}’s voice did not appear. \\

23 & Did \textbf{\{D\}} speak during this conversation?
   & No, \textcolor{red}{\textbf{\{D\}}} did not speak. \\

\bottomrule
\end{tabularx}
\caption{Multi-speaker recognition QA templates for a four-speaker candidate set \{\{A\},\{B\},\{C\},\{D\}\}: Case 5 (none present). \textbf{\textcolor{green}{Green}} indicates present speakers and \textbf{\textcolor{red}{red}} indicates absent speakers.}
\label{tab:ms_recognition_4spk_case5}
\end{table*}

In PALM-Bench, we use templates to systematically generate  QA samples for two single-speaker tasks: (i) \emph{concept activation} (recognition) and (ii) \emph{selective understanding} (captioning). For recognition, templates cover both positive and negative samples. For captioning, positive samples provide question templates only, and answers are produced by summarizing ASR transcripts with an LLM.

For multi-speaker settings, captioning reuses the same single-speaker captioning templates because the task remains target-subject-centric even when other speakers are present. In recognition templates, \textbf{P1/P2/P3/P4} denote speakers that are \emph{present} in the audio, and \textbf{UP1/UP2/UP3/UP4} denote speakers that are \emph{absent}.

Single-speaker recognition QA templates are provided in \emph{(Table~\ref{tab:qa_templates_single_recognition})}, covering both positive (present) and negative (absent) cases. And  captioning QA templates are provided in \emph{(Table~\ref{tab:qa_templates_single_caption})}.

For multi-speaker recognition, we organize templates by the size of the candidate speaker pool and cover all presence–absence configurations. Results for two-speaker candidates are shown in Tables~\ref{tab:ms_recognition_2spk_case12} and~\ref{tab:ms_recognition_2spk_case34}, three-speaker candidates in Tables~\ref{tab:ms_recognition_3spk_case12} and~\ref{tab:ms_recognition_3spk_case34}, four-speaker candidates in Tables~\ref{tab:ms_recognition_4spk_case12}, \ref{tab:ms_recognition_4spk_case34}, and\ref{tab:ms_recognition_4spk_case5}.
These templates enable controlled evaluation under multi-speaker interference by explicitly indicating present (P*) and absent (UP*) speakers.

\subsection{QA Examples}
\label{appendix-qa examples}

\begin{table*}[t]
\centering
\small
\setlength{\tabcolsep}{6pt}
\renewcommand{\arraystretch}{1.05}
\begin{tabularx}{\textwidth}{c >{\raggedright\arraybackslash}X >{\raggedright\arraybackslash}X}
\toprule

\multicolumn{3}{c}{\textbf{\textit{Recognition}}} \\
\midrule
\textbf{Index} & \textbf{Question} & \textbf{Answer} \\
\midrule

1 & Is \textbf{Amy} speaking in this audio?
& No, \textcolor{red}{\textbf{Amy}} is not present. \\

2 & Does this audio contain \textbf{Daisy}'s voice?
& Yes, \textcolor{green}{\textbf{Daisy}} is speaking, and the style matches her voice. \\

3 & Is \textbf{Catherine} in this recording?
& Yes, the voice clearly belongs to \textcolor{green}{\textbf{Catherine}}. \\

4 & Can \textbf{Bob} be heard in this recording?
& No, the voice does not sound like \textcolor{red}{\textbf{Bob}}. \\

5 & Are both \textbf{Daisy} and \textbf{Bob} speaking in this audio?
& No, \textcolor{red}{\textbf{Daisy}} is not present, and only \textcolor{green}{\textbf{Bob}} is speaking. \\

6 & Do \textbf{Amy} and \textbf{Catherine} appear in this conversation?
& Yes, both \textcolor{green}{\textbf{Amy}} and \textcolor{green}{\textbf{Catherine}} are present. \\

7 & Does this dialogue include \textbf{Catherine} and \textbf{Daisy}?
& \textcolor{red}{\textbf{Catherine}} is not present, while \textcolor{green}{\textbf{Daisy}} is speaking. \\

8 & Are the two speakers \textbf{Amy} and \textbf{Daisy}?
& \textcolor{green}{\textbf{Amy}} is present, but \textcolor{red}{\textbf{Daisy}} is not. \\

9 & Does this conversation include \textbf{Bob}, \textbf{Catherine}, and \textbf{Daisy}?
& Yes, \textcolor{green}{\textbf{Bob}}, \textcolor{green}{\textbf{Catherine}}, and \textcolor{green}{\textbf{Daisy}} are present. \\

10 & Did \textbf{Amy}, \textbf{Catherine}, and \textbf{Daisy} all participate in this dialogue?
& No, \textcolor{red}{\textbf{Amy}} and \textcolor{red}{\textbf{Catherine}} are not present, and only \textcolor{green}{\textbf{Daisy}} is speaking. \\

11 & Whose voices can be heard among \textbf{Amy}, \textbf{Bob}, and \textbf{Daisy}?
& Only \textcolor{green}{\textbf{Amy}} is present, while \textcolor{red}{\textbf{Bob}} and \textcolor{red}{\textbf{Daisy}} are not. \\

12 & Which of \textbf{Amy}, \textbf{Bob}, \textbf{Catherine}, and \textbf{Daisy} appear in this conversation?
& \textcolor{green}{\textbf{Amy}}, \textcolor{green}{\textbf{Bob}}, \textcolor{green}{\textbf{Catherine}}, and \textcolor{green}{\textbf{Daisy}} are present. \\

13 & Which voices can be heard among \textbf{Amy}, \textbf{Bob}, \textbf{Catherine}, and \textbf{Daisy}?
& Only \textcolor{green}{\textbf{Amy}} can be heard, while \textcolor{red}{\textbf{Bob}}, \textcolor{red}{\textbf{Catherine}}, and \textcolor{red}{\textbf{Daisy}} are not. \\

\addlinespace[0.6em]
\midrule
\multicolumn{3}{c}{\textbf{\textit{Captioning}}} \\
\midrule
\textbf{Index} & \textbf{Question} & \textbf{Answer} \\
\midrule

1 & What is \textbf{Amy} talking about in this audio? & \textcolor{green}{\textbf{Amy}} strongly opposes mandatory health interventions and emphasizes personal freedom. \\
2 & What does \textbf{Catherine} say in this conversation? & \textcolor{red}{\textbf{Catherine}} is not present, so there is nothing to summarize. \\
3 & Who expresses admiration for someone's perseverance and courage? & \textcolor{green}{\textbf{Catherine}} expresses admiration for that person's determination and courage. \\
4 & What kind of art is \textbf{Bob} interested in? & \textcolor{green}{\textbf{Bob}} is interested in printmaking. \\

\addlinespace[0.6em]
\midrule
\multicolumn{3}{c}{\textbf{\textit{Personalized Reasoning}}} \\
\midrule
\textbf{Index} & \multicolumn{2}{l}{\textbf{Question}} \\
\midrule

1  & \multicolumn{2}{l}{Can you recommend a \textbf{song} that he or she might like?} \\
2  & \multicolumn{2}{l}{What \textbf{music style or artist} might he or she prefer?} \\
3  & \multicolumn{2}{l}{When he or she feels down, what kind of \textbf{music} might they listen to?} \\
4  & \multicolumn{2}{l}{Which \textbf{song} best matches his or her current mood?} \\
5  & \multicolumn{2}{l}{If he or she attended a live concert, whose \textbf{performance} would they choose?} \\
6  & \multicolumn{2}{l}{What kinds of \textbf{books} might he or she enjoy reading?} \\
7  & \multicolumn{2}{l}{If you recommend one \textbf{book} to motivate him or her now, what would it be?} \\
8  & \multicolumn{2}{l}{What kind of \textbf{reading environment} might he or she like?} \\
9  & \multicolumn{2}{l}{Can you recommend a \textbf{movie} that he or she might enjoy?} \\
10 & \multicolumn{2}{l}{What type of \textbf{movies} is he or she most likely to watch?} \\
11 & \multicolumn{2}{l}{If he or she watches a movie tonight, what \textbf{genre} would they choose?} \\
12 & \multicolumn{2}{l}{In social settings, what \textbf{topics} is he or she most likely to bring up?} \\
13 & \multicolumn{2}{l}{What \textbf{social issues} might he or she have strong opinions about?} \\
14 & \multicolumn{2}{l}{How is he or she most likely to spend the \textbf{weekend}?} \\
15 & \multicolumn{2}{l}{How does he or she usually cope with \textbf{pressure}?} \\

\bottomrule
\end{tabularx}
\caption{Example QA pairs of three tasks from PALM-Bench. \textbf{Bold text} highlights personalized content.  \textbf{\textcolor{green}{Green}} names indicate present speakers, while \textbf{\textcolor{red}{red}} names indicate absent speakers.}
\label{tab:appendix_qa_examples}
\end{table*}
\begin{table*}[t]
\centering
\small
\setlength{\tabcolsep}{6pt}
\renewcommand{\arraystretch}{1.2}
\begin{tabularx}{\textwidth}{c >{\raggedright\arraybackslash}X >{\raggedright\arraybackslash}X}
\toprule
\textbf{Speaker} & \textbf{Human Description (HD)} & \textbf{Acoustic Description (AD)} \\
\midrule

\textbf{Amy} &
Amy’s voice sounds immediately \textcolor{blue}{\textbf{bright and clear}}, with an 
\textcolor{blue}{\textbf{energetic female quality}} that feels refreshing upon first hearing. 
Her speaking rate is \textcolor{blue}{\textbf{relatively fast}}, and her intonation shows 
\textcolor{blue}{\textbf{pronounced variation}}, especially when expressing emotions or emphasizing key words, 
where pitch naturally rises and conveys a \textcolor{blue}{\textbf{youthful liveliness}}. 
High-frequency components are particularly salient, giving the voice a transparent and clean impression 
without noticeable nasality or hoarseness. Her articulation is \textcolor{blue}{\textbf{crisp and precise}}, 
with little dragging or roughness. Overall, she comes across as a 
\textcolor{blue}{\textbf{young woman in her twenties}}, speaking in a 
\textcolor{blue}{\textbf{direct and forceful manner}} with 
\textcolor{blue}{\textbf{vivid emotional expression}}. 
When angry, both pitch and loudness increase noticeably, yet clarity and control are largely maintained. &
Female voice with a fundamental frequency primarily distributed between 
\textcolor{red}{\textbf{110–140 Hz}}, consistent with typical adult female vocal fold vibration. 
The voice is bright and clear, with prominent high-frequency harmonics in the 
\textcolor{red}{\textbf{2–5 kHz}} range and no low-frequency chest resonance associated with male voices. 
Timbre is clean and full, with nasal resonance energy below 
\textcolor{red}{\textbf{10\%}} of the fundamental frequency and vocal noise accounting for less than 
\textcolor{red}{\textbf{5\%}} of the spectrum. 
Speaking rate is approximately \textcolor{red}{\textbf{200 characters per minute}}. 
Pitch variation is evident, with fundamental frequency increasing by about 
\textcolor{red}{\textbf{20–30 Hz}} on emotionally emphasized words. 
Under anger, peak loudness reaches around \textcolor{red}{\textbf{65 dB}}, with an average of 
\textcolor{red}{\textbf{55 dB}}, consistent with acoustic characteristics of women aged 
\textcolor{red}{\textbf{20–30}}. \\

\midrule

\textbf{Bob} &
Bob’s voice sounds \textcolor{blue}{\textbf{steady and composed}}, reflecting a 
\textcolor{blue}{\textbf{typical adult male calmness}}. His pitch is 
\textcolor{blue}{\textbf{relatively low}}, with some thickness but without excessive heaviness, 
and the timbre is clean, lacking nasal quality or hoarseness, resembling a 
\textcolor{blue}{\textbf{standard male mid-range voice}}. 
He speaks at a \textcolor{blue}{\textbf{moderate pace}} that feels comfortable and 
\textcolor{blue}{\textbf{well organized}}, making his speech easy to follow. 
His Mandarin pronunciation is \textcolor{blue}{\textbf{highly standard}}, with clear articulation 
and virtually no regional accent or slurred sounds. 
The rhythm of his speech is \textcolor{blue}{\textbf{stable}}, giving an impression of 
\textcolor{blue}{\textbf{reliability and rationality}}. 
Overall, he comes across as a 
\textcolor{blue}{\textbf{knowledge-oriented man in his early thirties}}, speaking calmly, confidently, and clearly. &
Adult male voice with a fundamental frequency stably distributed between 
\textcolor{red}{\textbf{90–130 Hz}}. 
Low-frequency chest resonance is evident, with spectral energy concentrated in the 
\textcolor{red}{\textbf{100–200 Hz}} range. 
Mid-frequency energy (\textcolor{red}{\textbf{500–1000 Hz}}) is moderate, while high-frequency harmonics 
(\textcolor{red}{\textbf{2–4 kHz}}) remain clear. 
Speaking rate averages around \textcolor{red}{\textbf{160 characters per minute}}. 
Mandarin pronunciation is standard, with no identifiable dialectal features. 
The overall acoustic profile aligns with that of an adult male aged 
\textcolor{red}{\textbf{25–35}}. \\

\midrule

\textbf{Catherine} &
Catherine’s voice feels \textcolor{blue}{\textbf{light and well organized}}, and 
\textcolor{blue}{\textbf{immediately approachable}}. 
Her timbre is \textcolor{blue}{\textbf{relatively high but not piercing}}, and the voice projects clearly. 
She speaks at a \textcolor{blue}{\textbf{compact yet unhurried pace}}, maintaining a 
\textcolor{blue}{\textbf{composed rhythm}}. 
Her articulation is \textcolor{blue}{\textbf{exceptionally clear}}, with well-separated consonants and vowels. 
Intonation varies naturally, with gentle upward movement during emphasis or questions, 
while overall vocal control remains \textcolor{blue}{\textbf{steady}}. 
Overall, she sounds like a 
\textcolor{blue}{\textbf{well-spoken woman in her twenties}} with 
\textcolor{blue}{\textbf{clear logic and refined delivery}}. &
Typical adult female voice with a fundamental frequency mainly distributed between 
\textcolor{red}{\textbf{190–250 Hz}}, averaging around 
\textcolor{red}{\textbf{220 Hz}}. 
High-frequency harmonics in the \textcolor{red}{\textbf{3000–5000 Hz}} range are well represented. 
High-frequency noise accounts for less than \textcolor{red}{\textbf{5\%}} of the spectrum. 
Speaking rate is approximately \textcolor{red}{\textbf{200–220 characters per minute}}. 
Pitch rises by about \textcolor{red}{\textbf{10–20 Hz}} during emphasis or questions. 
The acoustic profile matches stable Mandarin female speech in the 
\textcolor{red}{\textbf{20–30}} age range. \\
\bottomrule
\end{tabularx}
\caption{Examples of human descriptions (HD) and acoustic descriptions (AD) used in PALM-Bench. \textcolor{blue}{\textbf{Blue bold text}} highlights perceptual and qualitative cues in HD, while \textcolor{red}{\textbf{red bold text}} marks key acoustic attributes in AD.}
\label{tab:appendix_hd_ad_examples}
\end{table*}

Table~\ref{tab:appendix_qa_examples} presents representative QA examples from the three tasks in PALM-Bench, illustrating how personalized subjects and personalization targets are formulated across recognition, captioning, and personalized reasoning settings.

\subsection{Examples of Acoustic Description and Human Description}

Table~\ref{tab:appendix_hd_ad_examples} presents examples of human descriptions (HD) and acoustic descriptions (AD) used in PALM-Bench. The examples illustrate how perceptual impressions and signal-level acoustic attributes characterize personalized subjects from different perspectives.

\section{Prompt Details}
\label{appendix-prompt}

\begin{table*}[t]
\centering
\small
\setlength{\tabcolsep}{4pt}
\renewcommand{\arraystretch}{1.05}
\begin{tabularx}{\textwidth}{@{}X@{}}
\toprule
\multicolumn{1}{@{}c@{}}{\textbf{Prompt: Acoustic Description Generation}}\\
\midrule

You are to produce an as-detailed-as-possible professional analysis report \textbf{based only on this audio}, focusing on the \textbf{primary speaker} only. The report must include the sections below. \\
\addlinespace[0.55em]

\textbf{<Gender and Age Inference>} \\
\textbullet\ Determine whether the speaker is more likely \textbf{male or female}, and provide an approximate \textbf{age range} (e.g., 20--30). Include \textbf{specific acoustic evidence}, for example: \\
\quad \textbullet\ F0 (fundamental frequency) overall level and variability (you may describe it as roughly \textit{low / mid / relatively high}). \\
\quad \textbullet\ Whether the voice sounds \textbf{mature} or has a \textbf{youthful quality}. \\
\quad \textbullet\ Resonance placement (e.g., chest resonance, head resonance). \\
\addlinespace[0.75em]

\textbf{<Intonation, Speech Rate, and Rhythm>} \\
\textbullet\ Describe the \textbf{speech rate} (fast / medium / slow), optionally using a rough analogy such as \textbf{approximately how many characters per second}. \\
\textbullet\ Describe whether the \textbf{intonation} is steady, whether \textbf{pitch variation} is prominent, and whether there are clear \textbf{rising} or \textbf{falling} contours. \\
\textbullet\ Analyze \textbf{pausing patterns} (whether pauses are frequent, and whether they convey \textbf{thinking} or \textbf{hesitation}). \\
\textbullet\ If there is \textbf{stress} or \textbf{emphasis}, indicate where it occurs and infer the speaking style (e.g., serious / relaxed / casual). \\
\addlinespace[0.75em]

\textbf{<Timbre and Resonance Characteristics>} \\
\textbullet\ Use \textbf{multiple adjectives} to describe timbre holistically (e.g., bright, rich, nasal, husky, soft, sharp, magnetic); do not provide only one word. \\
\textbullet\ Analyze resonance characteristics: is \textbf{chest}, \textbf{head}, or \textbf{throat} resonance more prominent, and how strong is the voice's \textbf{projection} or \textbf{penetration}? \\
\textbullet\ Describe the \textbf{spectral impression}: whether \textbf{high-frequency} components stand out, whether the \textbf{low end} is muddy, and whether overall energy is \textbf{concentrated} or \textbf{evenly distributed}. \\
\textbullet\ If you can perceive \textbf{formant-like traits} (F1/F2 tendencies), explain them accessibly (e.g., vowel articulation seems \textbf{more open vs.\ more closed}, oral cavity space seems \textbf{larger vs.\ smaller}). \\
\addlinespace[0.75em]

\textbf{<Accent, Pronunciation Habits, and Standard Mandarin Proficiency>} \\
\textbullet\ Judge whether there is a noticeable \textbf{regional accent}. If present, describe the likely type as best you can (e.g., Northern Mandarin region, Jiangsu--Zhejiang area), but use \textbf{hedged wording} such as ``possibly'' or ``sounds like.'' \\
\textbullet\ Identify \textbf{specific pronunciation habits}, such as clarity of \textbf{retroflex vs.\ non-retroflex} consonants, whether \textbf{front vs.\ back nasal finals} are mixed, whether \textbf{erhua} (rhotic suffixing) is prominent, and whether initials/finals are \textbf{weakened} or \textbf{unclear}. \\
\textbullet\ Provide an overall evaluation of \textbf{Mandarin fluency and clarity} (e.g., articulation clarity, connected-speech phenomena). \\
\addlinespace[0.75em]

\textbf{<Emotional State and Speaking Style>} \\
\textbullet\ Analyze the overall \textbf{emotional tone} while speaking (calm, tense, excited, relaxed, fatigued, etc.), and explain which \textbf{auditory impressions} or \textbf{rhythmic features} support the inference. \\
\textbullet\ Describe the \textbf{speaking style} (formal / casual / friendly / distant, etc.), possibly combining cues such as speech rate, magnitude of intonation changes, and frequency of discourse particles. \\
\textbullet\ If you can infer traits such as \textbf{politeness level}, \textbf{controlling tendency}, or \textbf{logical organization} from the voice, provide a brief analysis. \\
\addlinespace[0.75em]

\textbf{<Overall Summary and Uncertainty>} \\
\textbullet\ In a short paragraph, synthesize the above into an overall profile of the speaker (e.g., likely age band, personality tendencies, typical speaking context). \\
\textbullet\ State explicitly that because this analysis is based only on \textbf{short-duration audio} and \textbf{listening impressions}, all conclusions are \textbf{uncertain} and intended only for \textbf{rough impressions}, not strict identity recognition. \\
\addlinespace[0.55em]

\textbf{Overall Output Requirements} \\
\textbullet\ The overall response should use headings and bullet points, expanding each point in detail. \\
\textbullet\ After each sub-point, include a brief statement of \textbf{acoustic evidence}. \\
\textbullet\ Keep the total length \textbf{not less than 600 words}. \\
\textbullet\ Do not output any extra explanation; provide the analysis results directly. \\
\\
\bottomrule
\end{tabularx}
\caption{The prompt used in acoustic description generation.}
\label{tab:prompt_acoustic_description_generation}
\end{table*}

\begin{table*}[t]
\centering
\small
\setlength{\tabcolsep}{4pt}
\renewcommand{\arraystretch}{1.05}
\begin{tabularx}{\textwidth}{@{}X@{}}
\toprule
\multicolumn{1}{@{}c@{}}{\textbf{Prompt: Audio Captioning (QA-based Caption Generation)}}\\
\midrule

You are a professional assistant for generating question--answer pairs. You will be given the transcript of an audio utterance spoken by \textbf{<spk\_name>}. Generate \textbf{three} pairs of questions and answers based only on the provided transcript. Each answer must explicitly include the name \textbf{<spk\_name>}. \\
\addlinespace[0.55em]

\textbf{<Input>} \\
\textbullet\ A single transcript sentence spoken by \textbf{<spk\_name>}. \\
\addlinespace[0.75em]

\textbf{<Output Format>} \\
\textbullet\ Output must strictly follow this format: \\
\quad \texttt{question1: xxx?, answer1: xxx.} \\
\quad \texttt{question2: xxx?, answer2: xxx.} \\
\quad \texttt{question3: xxx?, answer3: xxx.} \\
\addlinespace[0.55em]

\textbf{Overall Output Requirements} \\
\textbullet\ Generate exactly \textbf{three} question--answer pairs. \\
\textbullet\ Each answer must contain the name \textbf{<spk\_name>}. \\
\textbullet\ Do not output any extra explanation; output only the formatted results. \\

\midrule
\multicolumn{1}{@{}c@{}}{\textbf{Prompt: Audio Captioning (Summarization-based)}}\\
\midrule

You are a professional text summarization assistant. Please concisely summarize, the textual content of this audio transcription. Even if the transcription is very short, you must restate it in your own words and must not output a summary that is identical to the original text.\\

\bottomrule
\end{tabularx}
\caption{Prompts used for audio captioning, including QA-based caption generation and summarization-based caption generation.}
\label{tab:prompt_caption_answer_generation}
\end{table*}
\begin{table*}[t]
\centering
\small
\renewcommand{\arraystretch}{1.05}
\setlength{\tabcolsep}{4pt}
\begin{tabularx}{\textwidth}{@{}X@{}}
\toprule
\multicolumn{1}{@{}c@{}}{\textbf{Prompt: Captioning Evaluation}}\\
\midrule
You are a strict but not overly nitpicky reviewer responsible for judging whether a caption (summary) correctly captures the semantics of the ground truth.\\
\addlinespace[0.55em]
\textbf{<Core Scoring Rules (Must Follow)>}\\
\textbullet\ \textbf{Score 1 (Completely correct)}: Any of the following cases should be judged as 1:\\
\quad \textbullet\ Synonymous paraphrase (different wording or order, but identical meaning).\\
\quad \textbullet\ Reasonable elaboration (adds some reasonable details based on the ground truth, without changing the core meaning).\\
\quad \textbullet\ More concise but semantically consistent (shorter description, but fully covers all core meaning).\\
\quad \textbullet\ Information is compressed or re-expressed, but key content is not lost.\\
\textbullet\ \textbf{Score 0.5 (Partially correct)}:\\
\quad \textbullet\ Covers part of the ground truth but omits other important parts.\\
\quad \textbullet\ Contains correct content but mixes in obvious incorrect or opposite information.\\
\textbullet\ \textbf{Score 0 (Completely incorrect)}:\\
\quad \textbullet\ Semantics are inconsistent with the ground truth.\\
\quad \textbullet\ Speaker identity is judged incorrectly (reverses whether someone is present or absent).\\
\quad \textbullet\ Describes a completely different event.\\
\addlinespace[0.75em]
\textbf{<Few-shot Examples>}\\
\textbullet\ \textbf{Example 1 (Synonymous paraphrase $\rightarrow$ Score 1)}\\
\quad \textbullet\ ground\_truth: ``Amy is introducing a cultural diversity exhibition; she mentions that it includes traditional clothing and handicrafts from different countries.''\\
\quad \textbullet\ response: ``Amy explains that the exhibition showcases traditional clothing and handicraft works from different countries.''\\
\quad \textbullet\ Expected output: \{"score": 1, "comment": "This is a synonymous paraphrase; the semantics are consistent."\}\\
\textbullet\ \textbf{Example 2 (Reasonable elaboration $\rightarrow$ Score 1)}\\
\quad \textbullet\ ground\_truth: ``Daisy thinks snakes are beneficial to farmland because they can control the number of mice.''\\
\quad \textbullet\ response: ``Daisy explains that snakes help control mouse numbers, which benefits the farmland ecosystem, and she mentions the importance of maintaining ecological balance.''\\
\quad \textbullet\ Expected output: \{"score": 1, "comment": "The elaboration is reasonable and does not change the core meaning."\}\\
\textbullet\ \textbf{Example 3 (More concise but semantically consistent $\rightarrow$ Score 1)}\\
\quad \textbullet\ ground\_truth: ``Bob blesses the other person to find suitable health supplements and expresses the wish that they stay healthy.''\\
\quad \textbullet\ response: ``Bob hopes the other person finds suitable health supplements and stays healthy.''\\
\quad \textbullet\ Expected output: \{"score": 1, "comment": "More concise, but the core semantics are fully preserved."\}\\
\textbullet\ \textbf{Example 4 (Target speaker absent in the audio, and the model also does not identify the target speaker $\rightarrow$ Score 1)}\\
\quad \textbullet\ ground\_truth: ``There is no Bob's voice in this audio; it cannot be summarized.''\\
\quad \textbullet\ response: ``There is no Bob's voice.''\\
\quad \textbullet\ Expected output: \{"score": 1, "comment": "The target speaker is not in the audio, and the model's answer is correct."\}\\
\textbullet\ \textbf{Example 5 (Partially correct $\rightarrow$ Score 0.5)}\\
\quad \textbullet\ ground\_truth: ``Daisy thinks snakes are beneficial to farmland because they can control mouse numbers, and she reminds people not to kill snakes casually.''\\
\quad \textbullet\ response: ``Daisy says snakes are good for farmland and can control mouse numbers.''\\
\quad \textbullet\ Expected output: \{"score": 0.5, "comment": "It ignores the important part about `do not kill snakes casually'."\}\\
\textbullet\ \textbf{Example 6 (Completely incorrect $\rightarrow$ Score 0)}\\
\quad \textbullet\ ground\_truth: ``Bob blesses the other person to find suitable health supplements and wishes them good luck.''\\
\quad \textbullet\ response: ``Bob is discussing a family plan to go to the beach for vacation.''\\
\quad \textbullet\ Expected output: \{"score": 0, "comment": "It is inconsistent with the ground truth."\}\\
\textbullet\ \textbf{Example 7 (Recognition error $\rightarrow$ Score 0)}\\
\quad \textbullet\ ground\_truth: ``Bob blesses the other person to find suitable health supplements and wishes them good luck.''\\
\quad \textbullet\ response: ``Unable to recognize.''\\
\quad \textbullet\ Expected output: \{"score": 0, "comment": "The target speaker is present in the audio, but they were not identified or the recognition task failed."\}\\
\textbullet\ \textbf{Example 8 (Recognition error $\rightarrow$ Score 0)}\\
\quad \textbullet\ ground\_truth: ``There is no Bob's voice; it cannot be summarized.''\\
\quad \textbullet\ response: ``Bob blesses the other person to find suitable health supplements and wishes them good luck.''\\
\quad \textbullet\ Expected output: \{"score": 0, "comment": "The target speaker is not in the audio; the model summary is wrong."\}\\
\textbullet\ \textbf{Example 8 (Recognition failure $\rightarrow$ Score 0)}\\
\quad \textbullet\ ground\_truth: ``There is no Bob's voice; it cannot be summarized.''\\
\quad \textbullet\ response: ``Unable to recognize / unable to determine.''\\
\quad \textbullet\ Expected output: \{"score": 0, "comment": "The target speaker is not in the audio; the model cannot determine whether the target speaker is present, i.e., the recognition task failed, so no score is awarded."\}\\
\addlinespace[0.55em]
\textbf{<Important Notes>}\\
\textbullet\ Please output valid JSON; the format must be \textbf{EXACT}: \{"score": 0 or 0.5 or 1, "comment": "your brief one-sentence review / reason"\}. Do not output any extra explanation.\\
\bottomrule
\end{tabularx}
\caption{The prompt used in captioning evaluation.}
\label{tab:prompt_caption_evaluation_prompt}
\end{table*}

\begin{table*}[t]
\centering
\small
\renewcommand{\arraystretch}{1.05}
\setlength{\tabcolsep}{4pt}
\begin{tabularx}{\textwidth}{@{}X@{}}
\toprule
\multicolumn{1}{@{}c@{}}{\textbf{Prompt: Recognition Evaluation}}\\
\midrule
You are a strict but not overly nitpicky reviewer who can score a speaker-recognition task across different speaker scenarios (currently only including 1/2/3/4 speakers), and can judge whether the model response correctly identifies, as described in the ground truth, \textbf{which speakers appear and which do not appear}. Specific requirements:\\
\textbullet\ You must extract from the question all speakers that need to be judged. The set of names appearing in the question is exactly the speaker list for this item, denoted as \textbf{N} people.\\
\textbullet\ You must compare the ground truth and the response for each of these speakers one by one. The ground truth will explicitly state which people appear and which do not appear. The response will also express these judgments in natural language. As long as the response’s judgment for a person’s presence matches the ground truth, it is counted as correct.\\
\textbullet\ The scoring is evenly divided by the number of people: if among \textbf{N} speakers there are \textbf{k} correctly judged, then \textbf{score = k / N}. No rounding is allowed. The possible correct scores for different \textbf{N} are as follows:\\
\quad \textbullet\ N=1 $\rightarrow$ score \{1, 0\}.\\
\quad \textbullet\ N=2 $\rightarrow$ score \{1, 0.5, 0\}.\\
\quad \textbullet\ N=3 $\rightarrow$ score \{1, 0.6, 0.3, 0\}.\\
\quad \textbullet\ N=4 $\rightarrow$ score \{1, 0.75, 0.5, 0.25, 0\}.\\
\textbullet\ The output format must be valid JSON, in the following form:\\
\quad \textbullet\ \{"score": number, "comment": "Provide the scoring rationale in 1--2 sentences."\}\\
\quad \textbullet\ Do not output any extra content.\\
\addlinespace[0.55em]
\textbf{<Examples>}\\
\textbullet\ \textbf{Example 1 (N=2, 2 correct, score 1)}\\
\quad \textbullet\ question: ``Please confirm: do the speakers include Bob and Amy?''\\
\quad \textbullet\ ground\_truth: ``This audio indeed contains the voices of Bob and Amy.''\\
\quad \textbullet\ response: ``Confirmed, it includes Bob and Amy.''\\
\quad \textbullet\ Expected output: \{"score": 1, "comment": "Both Bob and Amy are judged correctly regarding their presence."\}\\
\textbullet\ \textbf{Example 2 (N=2, 1 correct, score 0.5)}\\
\quad \textbullet\ question: ``Amy is speaking in this audio; is there also Catherine’s voice?''\\
\quad \textbullet\ ground\_truth: ``Amy is present, but Catherine is not.''\\
\quad \textbullet\ response: ``Amy’s voice is present, and so is Catherine’s.''\\
\quad \textbullet\ Expected output: \{"score": 0.5, "comment": "Amy is judged correctly; Catherine is judged incorrectly."\}\\
\textbullet\ \textbf{Example 3 (N=2, 0 correct, score 0)}\\
\quad \textbullet\ question: ``Do Amy and Daisy both appear in this conversation at the same time?''\\
\quad \textbullet\ ground\_truth: ``Only Daisy appears; Amy does not.''\\
\quad \textbullet\ response: ``Amy appears; Daisy does not.''\\
\quad \textbullet\ Expected output: \{"score": 0, "comment": "Both Amy and Daisy are judged incorrectly."\}\\
\textbullet\ \textbf{Example 4 (N=3, 3 correct, score 1)}\\
\quad \textbullet\ question: ``Can you hear Amy, Catherine, and Daisy in this conversation?''\\
\quad \textbullet\ ground\_truth: ``All three voices can be heard.''\\
\quad \textbullet\ response: ``The voices of Amy, Catherine, and Daisy can all be heard.''\\
\quad \textbullet\ Expected output: \{"score": 1, "comment": "Amy, Catherine, and Daisy are all judged correctly."\}\\
\textbullet\ \textbf{Example 5 (N=3, 2 correct, score 0.6)}\\
\quad \textbullet\ question: ``I hear Amy, Catherine, and Daisy in the dialogue, right?''\\
\quad \textbullet\ ground\_truth: ``Only Amy appears; Catherine and Daisy do not appear.''\\
\quad \textbullet\ response: ``Amy and Daisy appear; Catherine does not.''\\
\quad \textbullet\ Expected output: \{"score": 0.6, "comment": "Amy and Catherine are judged correctly; Daisy is judged incorrectly."\}\\
\textbullet\ \textbf{......}\\
\textbullet\ \textbf{Example 11 (N=4, 1 correct, score 0.25)}\\
\quad \textbullet\ question: ``Among Amy, Bob, Catherine, and Daisy, who appears?''\\
\quad \textbullet\ ground\_truth: ``Only Bob appears.''\\
\quad \textbullet\ response: ``Amy, Bob, Catherine, and Daisy all appear.''\\
\quad \textbullet\ Expected output: \{"score": 0.25, "comment": "Bob is judged correctly; Amy, Catherine, and Daisy are judged incorrectly."\}\\
\textbullet\ \textbf{Example 12 (N=4, 0 correct, score 0)}\\
\quad \textbullet\ question: ``Do Amy, Bob, Catherine, and Daisy speak in the audio?''\\
\quad \textbullet\ ground\_truth: ``Amy and Bob appear; Catherine and Daisy do not appear.''\\
\quad \textbullet\ response: ``Only Catherine and Daisy appear; Amy and Bob do not.''\\
\quad \textbullet\ Expected output: \{"score": 0, "comment": "All presence judgments for Amy, Bob, Catherine, and Daisy are incorrect."\}\\
\textbullet\ \textbf{Example 13 (N=1, 1 correct, score 1)}\\
\quad \textbullet\ question: ``Does Daisy speak?''\\
\quad \textbullet\ ground\_truth: ``Does not really seem like Daisy.''\\
\quad \textbullet\ response: ``Does not really seem like Daisy.''\\
\quad \textbullet\ Expected output: \{"score": 1, "comment": "The model output exactly matches the reference answer, so it receives full marks."\}\\
\addlinespace[0.55em]
\textbf{<Important Notes>}\\
\textbullet\ Important Note 1: Please output valid JSON; the format must be \textbf{EXACT}: \{"score": 0 or 0.5 or 1, "comment": "your brief one-sentence review / reason"\}. Do not output any extra explanation.\\
\textbullet\ Important Note 2: If the model output is exactly identical to the ground truth, it counts as full marks.\\
\bottomrule
\end{tabularx}
\caption{The prompt used in recognition evaluation.}
\label{tab:prompt_speaker_recognition_evaluation}
\end{table*}

\subsection{Generation Prompts for Captioning QA}

We use a unified prompt set to generate QA pairs for the captioning task. Depending on the output format, the prompts support both (i) answer-only generation, which produces a caption-style summary of the target subject’s content, and (ii) full QA pair generation, which includes a question paired with its corresponding answer. The prompts specify consistent formatting and attribution constraints to ensure that generated outputs remain grounded in the target speaker’s speech and exclude content from non-target speakers. The complete prompt templates are provided in Table~\ref{tab:prompt_caption_answer_generation}.

\subsection{Generation Prompts for Acoustic Description}

The complete prompt template is provided in Table~\ref{tab:prompt_acoustic_description_generation}, which guides the model to produce explicit descriptions suitable for personalization.

\subsection{Evaluation Prompts}

We use task-specific evaluation prompts to score model outputs for captioning and recognition tasks. For captioning, the prompt is structured as a rubric for grounding and target-subject attribution (Table~\ref{tab:prompt_caption_evaluation_prompt}). For recognition, the prompt evaluates whether present speakers are correctly accepted and absent speakers are correctly rejected in multi-speaker settings. (Table~\ref{tab:prompt_speaker_recognition_evaluation}).

 \end{document}